\documentclass[Afour,sageh,times]{sagej}

\usepackage{moreverb,url}
\usepackage[utf8]{inputenc}
\usepackage[T1]{fontenc}
\usepackage{multirow}
\usepackage{graphicx}
\usepackage{svg}
\usepackage{amsmath}
\usepackage{amsfonts,amssymb}
\usepackage{amsmath,amssymb,amsthm}
\usepackage{euscript}
\usepackage{newtxmath}

\DeclareMathAlphabet{\mathpzc}{T1}{pzc}{m}{it}
\DeclareMathOperator{\sgn}{sgn}
\DeclareMathOperator{\diag}{diag}

\usepackage{algorithm} 
\usepackage{algpseudocode}

\usepackage{amsfonts}

\usepackage[colorlinks,bookmarksopen,bookmarksnumbered,citecolor=red,urlcolor=red]{hyperref}

\newcommand\BibTeX{{\rmfamily B\kern-.05em \textsc{i\kern-.025em b}\kern-.08em
T\kern-.1667em\lower.7ex\hbox{E}\kern-.125emX}}

\usepackage{soul,color}
\soulregister\cite7
\soulregister\ref7
\soulregister\pageref7

\usepackage{algorithm} 
\usepackage{algpseudocode}
\usepackage{amsfonts}

\usepackage{lineno}

\begin{document}

\sethlcolor{yellow}

\runninghead{Modal-Graph Shape Servoing with Raw Point Clouds}

\title{Modal-Graph 3D Shape Servoing of Deformable Objects with Raw Point Clouds}

\author{Bohan Yang\affilnum{1},
Congying Sui\affilnum{1},
Fangxun Zhong\affilnum{1},
and Yun-Hui Liu\affilnum{1}
}

\affiliation{\affilnum{1} T Stone Robotics Institute, the Department of Mechanical and Automation Engineering, The Chinese University of Hong Kong, HKSAR, China}

\corrauth{Yun-Hui Liu, 
Room 208, 
William M.W. Mong Engineering Building, The Chinese University of Hong Kong, Shatin, N.T., HKSAR, China
}
\email{{\tt\small yhliu[at]cuhk.edu.hk}}

\begin{abstract}
Deformable object manipulation (DOM) with point clouds has great potential as non-rigid 3D shapes can be measured without detecting and tracking image features.
However,
robotic shape control of deformable objects with point clouds is challenging due to:
the unknown point correspondences 
and the noisy partial observability of raw point clouds;
the modeling difficulties of the relationship between point clouds and robot motions.
To tackle these challenges,
this paper introduces a novel modal-graph framework for the model-free shape servoing of deformable objects
with raw point clouds.
Unlike the existing works studying the object's geometry structure, 
we propose a modal graph to describe the low-frequency deformation structure of the DOM system,
which is robust to the measurement irregularities. 
The modal graph enables
us to directly extract low-dimensional deformation features from raw point clouds without extra processing of registrations, refinements, and occlusion removal.
It also preserves the spatial structure of the DOM system to inverse the feature changes into robot motions.
Moreover,
as the framework is built with unknown physical and geometric object models, 
we design an adaptive robust controller to deform the object toward the desired shape 
while tackling the modeling uncertainties, noises, and disturbances online.
The system is proved to be input-to-state stable (ISS) using Lyapunov-based methods.
Extensive experiments are conducted to validate our method using linear, planar, tubular, and volumetric objects under different settings.
\end{abstract}


\maketitle

\section{1. Introduction}
Automatic deformable object manipulation (DOM) can facilitate surgical robots (\cite{shademan2016supervised}), service robots (\cite{tsurumine2019deep}), and manufacturing industries (\cite{li2018vision}).
As a fundamental task in robotic DOM, 
shape servoing of deformable objects (i.e., deforming objects to desired shapes) is still an open problem.
High-dimensional 3D shapes undergoing complex deformation are difficult to be perceived and model.
In recent years, thanks to the development of 3D visual sensing techniques,
DOM with point cloud measurements has attracted increasing attention as
non-rigid shapes can be measured without detecting and tracking image features.

Although efforts have been devoted to the point-cloud-based deformation perception (\cite{tang2018track}, \cite{sundaresan2022diffcloud}) and learning (\cite{shen2022acid}),
robotic shape servoing with point clouds presents many technical challenges.
The reasons are twofold:
first, it is hard to model the relationship between high-dimensional point clouds and robot manipulation;
second, 
shape perceptions from raw point clouds suffer from
their unknown point-wise correspondences and noisy partial observability.
Existing works require extra processing (such as re-samplings (\cite{lagneau2020active}), registrations (\cite{jin2019robust}), and reconstructions (\cite{shi2022robocraft}, \cite{hu20193})) to refine or recover the shape geometry from raw point clouds.
However,
from the perspective of modal analysis (\cite{pentland1991closed}),
irregularities and noises of local measurements mainly influence the high-frequency deformation modes, leaving the low-frequency modes relatively unchanged.
Thus, we investigate the low-frequency modal representation of raw point clouds such that low-dimensional deformation features can be extracted for robot manipulation with finite degrees of freedom.
In addition, by properly setting the feature dimension, over-constrained solutions
can still be found to represent shape deformation even with partial or occluded observations.

This paper studies the model-free shape servoing of deformable objects with raw point clouds.
We introduce a novel graph-based framework called a modal graph that is generated by embedding modal analysis in a graph structure.
In the modal graph, nonrigid shapes can be modeled with 3D deformation organized on the graph nodes, which is further parameterized into a combination of low-frequency modes associated with the nodes.
The modal graph builds a low-frequency deformation structure for the DOM system.
It enables us to formulate low-frequency modal representations from raw point cloud measurements of the object, while also preserving the spatial structure to inverse changes in the modal representations into motions of the robot manipulation.
Based on the framework,
we develop the methods to extract low-dimensional deformation features from raw point clouds and to control the robot manipulation to deform the object to desired shapes.
Moreover, as the framework is constructed with unknown physical and geometric models of the object, we design adaptive robust control laws to tackle the modeling uncertainties, noises, and disturbances online.
The stability of the closed-loop system is analyzed using Lyapunov-based methods.
Extensive experiments are conducted to validate our method using linear, planar, tubular, and volumetric objects under different settings.
Compared with existing works, our main contributions are:\\
1) Unlike the existing works studying the object's geometry structure, 
we build a low-frequency deformation structure for the DOM system, which is robust to the measurement irregularities of raw point clouds;
\\
2) We can directly extract deformation features from raw point clouds without 
requiring extra processing of registrations, refinements, and occlusion removal;
\\
3) Our controller is input-to-state stable (ISS) without offline learning or identifying the object's physical and geometric models.

We also compare the proposed method with our previous modal-based deformation controller (\cite{10122176}) using stereo measurements of image features.
Our previous work is limited by
the requirements to track image features on objects and obtain samplings with known spatial orders and chronological correspondences.
In contrast, this paper proposes new representation and control methods to handle raw point clouds with random noises and unknown point-wise correspondences.
To be specific, first, unlike our previous method that computes deformation features using a linear basis, this paper develops a nonlinear feature extraction method to tackle raw point clouds;
Second, this paper designs new adaptive robust laws with dynamic state feedback to address the unmodeled effects in the nonlinear formulations, measurement noises, and disturbances;
In addition, by building the modal graph structure, this paper formulates features and control laws in a position-based manner, which improves the efficiency and generality of our previous element-dependent algorithms.

The rest of this paper is organized as follows:
Section 2 reviews the related works;
Section 3 formulates the problem and overviews the proposed method;
Section 4 presents the modal-graph framework;
Section 5 designs the adaptive robust controller;
Simulation analysis and experiment validations are presented in Section 6 and 7, respectively;
Section 8 states the discussions and conclusions.

\section{2. Related Works}
\subsection{2.1. Shape Control of Deformable Objects}
Through the efforts of many researchers,
varied methods have been proposed for the shape control of deformable objects.
Reviewing the vision-based approaches, a primitive stage is to use the feedback of the point (\cite{shin2019autonomous}, \cite{lagneau2020active}) and geometric features (\cite{navarro2016automatic}, \cite{hu2018three})
of some surface points.
Nevertheless, due to the insufficient description abilities of the point features, efforts have been devoted to designing 
more global shape descriptors.
Low-dimensional features, such as the spline curves (\cite{lagneau2020automatic}), the Fourier-based contour descriptors (\cite{navarro2018fourier}), 
and the modal-based deformation features (\cite{10122176}), are proposed 
to cope with the robot manipulation
with finite degrees of freedom.
For deformation controller design with unknown object models,
model-free strategies have been developed using online adapting (\cite{navarro2018fourier}, \cite{10122176}) and learning (\cite{yu2022global},
\cite{hu2018three}) techniques.
However, the requirements of tracking image features (i.e., corner points, curves, contours, etc.) affect the flexibility of these methods under practical applications.

For better application flexibility, deformation control under point cloud measurements has attracted increasing attention.
\cite{tang2022track} and \cite{jin2019robust}  developed methods to track key points from point cloud measurements to represent shapes and to estimate the deformation Jacobian matrix for cable manipulation.
\cite{ma2022shape} deformed cables based on data-driven geometric features fitted from RGB-D measurements.
They formulated control laws using the Jacobian matrix derived from geometric relationships.
However, these deformation controllers only deal with linear objects, and the influence of Jacobian estimation on the control performance is not further analyzed.
For planar and volumetric objects, 
\cite{hu20193} extracted the extended Fast Point Feature Histogram (FPFH) from point clouds, and learned manipulation laws with a Deep Neural Network (DNN).
However, the use of hard-coded features and the requirement of object-specific training limit the generalization of this work.

More recently, several learning-based methods have been proposed to work with raw point clouds.
\cite{thach2022learning} studied shape servoing with a neural network architecture called DeformerNet that operates on a partial-view point cloud of a deformable object.
They designed a sim-to-real method that both uses DNNs to extract features from point clouds and to learn manipulation laws to deform the object to a goal point cloud.
Instead of extracting features from point clouds, 
\cite{shi2022robocraft} proposed a model-based method to shape elasto-plastic objects, which transforms RGB-D data into particles and learns deformation behaviors using particle-based graph neural networks (GNNs).
The work (\cite{shen2022acid}) studying goal-conditioned DOM also involves deforming volumetric objects toward goal point clouds. 
The method learns the implicit neural representations to reconstruct full geometry and predict deformable dynamics from partial point cloud observations. 
The learned visual dynamics model is further applied to plan action sequences for DOM tasks.
By contrast, our method is
learning-free and requires no training process.
Moreover, we
compute deformation features and manipulation commands with transparent and interpretable mathematical formulations, which enables us to evaluate the stability of the system  analytically.

The aforementioned point-cloud-based works require additional processing of the noisy point clouds,
such as 
non-rigid registrations (\cite{jin2019robust}),
occlusion removal (\cite{hu20193}), 
re-samplings (\cite{lagneau2020active}, \cite{zhou2021lasesom}, \cite{thach2022learning}),
correspondence identification (\cite{shen2022acid}),
and surface reconstructions/refinements (\cite{shi2022robocraft}).
In comparison,
our method can directly extract deformation features from raw point clouds without extra point processing.

\subsection{2.2. Shape Representation of Deformable Objects}
Representation of non-rigid shapes is an essential problem when studying DOM.
The problem is technically challenging due to the complex physical properties and the infinite dimension of 3D deformation.
Early works of non-rigid shape modeling mainly use physically based deformable models (\cite{nealen2006physically}) to describe 3D shapes for computer simulations and animations.
Varied tools are proposed using 
energy functions (\cite{terzopoulos1987elastically}),
differential surface descriptors (\cite{botsch2007linear}),
and piece-wise quadratic model (\cite{fayad2010piecewise}).
For low-dimensional representations, methods using a reduced shape basis are also widely studied.
Orthogonal shape basis can be formulated using modal analysis (\cite{pentland1989good}), principal warps (\cite{bookstein1989principal}), and subspace deformation (\cite{barbivc2005real}).
Due to the complexities of physical models, other researchers develop geometrically based methods such as Fourier surfaces (\cite{kelemen1996segmentation}),
wavelet transform (\cite{davatzikos2003hierarchical}),
and deformation graph (\cite{davatzikos2003hierarchical}).
Nevertheless, 
directly transferring these shape modeling techniques to DOM tasks is challenging.
It can be quite difficult to obtain accurate physical models, and even harder to formulate the relationship between the geometrically based shape descriptors and robot motion.

Owing to the developments of computing and sensing techniques, real-time shape representation from point cloud measurements has drawn increasing attention.
Different shape reconstruction algorithms are proposed, including the template-based (\cite{zollhofer2014real}),
template-free (\cite{innmann2016volumedeform}), 
and correspondence-free (\cite{slavcheva2017killingfusion}) methods.
Learning-based methods are also proposed to further deal with partial observability (\cite{lin2022learning}), improve correspondence identification (\cite{shen2022acid}), and generate latent representations for point clouds (\cite{achlioptas2018learning}).
However, how to perform robotic manipulation from point cloud inputs
is still an open topic.
Several learning-based methods are proposed to establish the latent space roadmap (\cite{lippi2020latent}) and the latent feedback framework (\cite{zhou2021lasesom}) for DOM tasks.
Unlike these works using latent representations,
our method adopts modal analysis to construct a low-frequency deformation structure of the DOM system from raw point clouds.
This physically based technique allows us to analytically formulate control laws for robot manipulation and analyze modeling uncertainties in the low-dimensional modal space.

\subsection{2.3. Modeling Deformation for Robot Manipulation}
Another key component in robotic DOM is the modeling of 3D deformation.
Traditional approaches, such as the mass-spring system (\cite{liu2013fast}) and the finite element method (FEM) (\cite{zienkiewicz2005finite}), are established on high-dimensional discrete systems.
To cope with finite dimensions of robot motion under real-time performance requirements, research efforts have been devoted to reduced deformation models using subspace methods (\cite{an2008optimizing}) and data-driven methods (\cite{fulton2019latent}, \cite{tan2020realtime}).
Nevertheless, physical models are difficult to compute and identify because they are complex systems with highly coupled material parameters and high-dimensional geometric structures.
Other researchers explored the use of object geometries to approximate deformation behaviors under robot manipulation.
They proposed control laws using the diminishing rigidity approximation (\cite{mcconachie2020manipulating}), 
the As-Rigid-As-Possible (ARAP) model (\cite{shetab2022rigid}),
and the shape-template-based method (\cite{aranda2020monocular}).
Recently, increasing attention has been put to the deformation learning approaches that embed geometric graph structures into neural architectures.
Methods have been proposed using mesh-based (\cite{pfaff2020learning}) and particle-based (\cite{shi2022robocraft}, \cite{wang2022offline}) GNNs to learn deformation dynamics.
The learned deformation model 
can be used to 
predict deformation for planning manipulation actions.
By contrast, our modal graph is a physically based structure constructed by embedding modal analysis in a graph.
In addition, we do not use the graph to learn the deformation model of the object nor deform the graph to recover/predict shapes.
Instead, this paper uses the modal graph to formulate a low-frequency deformation representation.
The formulated representation is used in a feedback controller where online adaptations of low-dimensional modal parameters tackle unknown object models.
On the other hand, the modal-graph framework has the potential to be combined with the existing graph-based learning techniques
and extended to different DOM tasks.

\begin{figure*}[t]
    \centering
    \includegraphics[width=0.72\linewidth]{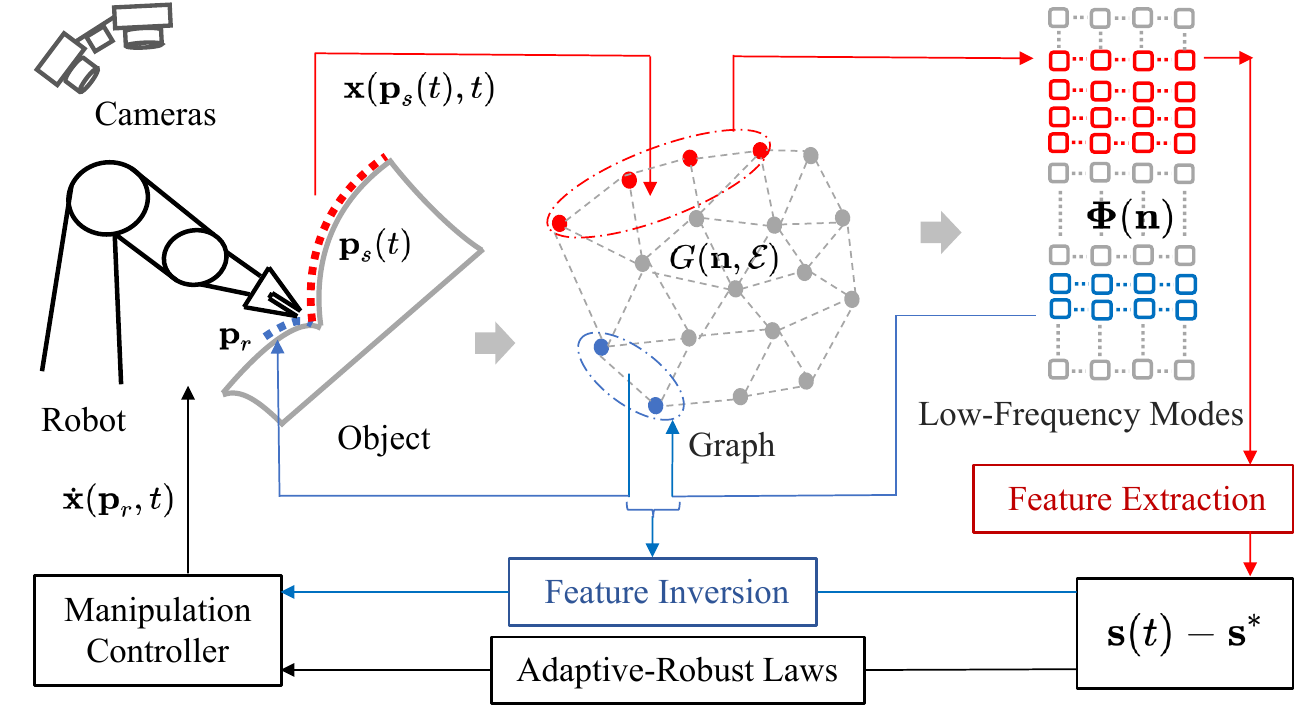}
    \caption{System configuration and the flowchart of the proposed method.
    Our modal-graph framework is developed on a graph $G(\mathbf{n},\mathcal{E})$ (with $n$ nodes) 
    embedded
    with the low-frequency deformation modes 
    $\mathbf{\Phi}(\mathbf{n})  \in \mathbb{R}^{3n \times m}$.
    Based on the framework, we develop an adaptive robust controller with feedback structures of the feature errors $\mathbf{s}(t) - \mathbf{s}^*$.
    The red/blue solid squares indicate the measured/manipulation points $\mathbf{p}(t)$/$\mathbf{r}$.
    The red/blue rounds in $G(\mathbf{n},\mathcal{E})$ indicate the nodes within the domain of influence of the measured/manipulation points (shown as the red/blue dashed circle), and their associated modes are indicated by the red/blue rows in $\mathbf{\Phi}(\mathbf{n})$.
     }
    \label{fig:system_configuration}
\end{figure*}

\section{3.Preliminaries}
\subsection{3.1 Notations}
In this paper, 
we denote scalar quantities by italic letters,
column vectors by lowercase bold letters,
and matrices by capital bold letters.
Leading superscripts are used to denote reference frames.
Variables without a leading superscript are defined under the default reference frame.
${}^a\mathbf{T}_b \in \mathbb{R}^{4 \times 4}$ denotes a homogeneous transformation matrix from the frame $\left \{ b \right \}$ to the frame $\left \{ a \right \}$.
$\left \{ {}^a\boldsymbol \epsilon_b^x, {}^a\boldsymbol \epsilon_b^y, {}^a\boldsymbol \epsilon_b^z \right \}$ denote the direction vectors of the Cartesian coordinate axes of frame $\left \{ b \right \}$ with respect to frame $\left \{ a \right \}$.
$(\mathbf{a} \cdot \mathbf{b})$ denotes the dot product of vector $\mathbf{a}$ and vector $\mathbf{b}$.
The symbol $\hat{}$ means estimated values; the symbol ${}^{*}$ means desired values.
The function $\max \left \{ a, b \right \}$ selects the largest values between the scalar $a$ and $b$.
The function $\min \left \{ a, b \right \}$ selects the smallest values between the scalar $a$ and $b$.
$\mathbf{I}_{k \times k} \in \mathbb{R}^{k \times k}$ denotes an identify matrix while  $\mathbf{0}_{k \times k} \in \mathbb{R}^{k \times k}$ denotes an zero matrix.
The operator $\diag(\mathbf{a})$ constructs a diagonal matrix from the vector $\mathbf{a}$.
The operator $\mathbf{\nabla} \times$ denotes taking the curl of a vector field.
$\dot{\mathbf{a}}(t)$ denotes the time derivative of the time-varying vector $\mathbf{a}(t)$.
A graph is denoted by $G(\mathbf{n},\mathcal{E})$, where $\mathbf{n}$ denotes the nodes and $\mathcal{E}$ denotes the edges.

\subsection{3.2. Problem Formulations}
This paper studies the shape control of a deformable object with unknown physical properties and undeformed geometry.
As shown in Figure~\ref{fig:system_configuration},
the object is manipulated by a robot via the manipulation points $\mathbf{r}$ of size $k$.
The shape of the object is measured with raw point clouds $\mathbf{p}(t)$.
Given sensing noises and occlusions, $\mathbf{p}(t)$ are of time-varying sizes $l(t)$, and their point-wise correspondences between times are unknown.
We propose a modal-graph framework to directly extract deformation features $\mathbf{s}(t) \in \mathbb{R}^{m}$ (with a low and fixed dimension $m$) from raw point cloud measurements.
Given the desired deformation features 
$\mathbf{s}^* \in \mathbb{R}^m$, 
we define the following shape servoing problem:
\newtheorem*{problem}{Problem}
\begin{problem}
Design a model-free controller with feedback of the deformation feature errors $\mathbf{e}_s(t) = \mathbf{s}(t) - \mathbf{s}^* \in \mathbb{R}^m$ to control the velocities $\dot{\mathbf{x}}(\mathbf{r},t) \in \mathbb{R}^{3k}$ of the manipulation points $\mathbf{r}$ such that as $\mathbf{e}_s(t)$ are minimized, the object is deformed to the desired shape.
\end{problem}

To clarify the derivation of our controller, we make the following assumptions:
\newtheorem{assumption}{Assumption}
\begin{assumption} 
The object is rigidly and firmly grasped by the robot via the manipulation points.
\label{as:point_manipulation}
\end{assumption}
\begin{assumption}
3D point cloud measurements of the object are segmented out of the background.
\end{assumption}
\begin{assumption}
The robot manipulating motion is sufficiently slow such that we can only consider the quasi-static elastic deformation of the object (\cite{navarro2016automatic}).
\label{as:qs_motion}
\end{assumption}

In the object manipulation system, we consider four coordinate frames: 
the camera frame $\left \{ c \right \}$ which is fixed in space;
the robot end-effector frame $\left \{ e \right \}$ which changes with the robot manipulation;
the point cloud frame $\left \{ p \right \}$ which changes with the object deformation;
the graph frame $\left \{ g \right \}$ which is fixed in the space.
In this paper, we select the graph frame $\left \{ g \right \}$ to be the default frame.

\subsection{3.3. Algorithm Overview}
Our modal graph is generated by embedding modal analysis in a graph structure.
Each graph node is associated with a set of low-frequency deformation modes.
We model the object shape as a deformed result of the modal graph.
Deformation applied on the graph boundary can propagate to the nearby graph nodes within a domain of influence
that is determined by a weight function computing rigidity.
In this way, the positions of object points can be modeled with weighted collections of displacements of graph nodes.
These nodal displacements are further transformed into a low-dimensional combination of the low-frequency modes associated with the nodes.
The modal coefficients are selected to be our deformation features.

The modal graph allows us to investigate the deformation of the manipulated object in two spaces:
the nodal space describing positions and displacements of the graph nodes;
the modal space spanned by the low-frequency modes of the graph.
In the nodal space, we can establish the relationships of the graph nodes both with the point cloud measurements and the manipulated points.
In the modal space, we describe the object deformation with the fixed- and low-dimensional deformation features $\mathbf{s}(t)$.
In this way, a low-frequency deformation structure for the DOM system is built with our modal-graph framework.

Based on the framework, we can extract the deformation features $\mathbf{s}(t)$ from raw point cloud measurements, while also inverting the feature changes to the motion of the manipulated points.
Then, using the feature inversion function, we design adaptive robust laws to control the robot manipulation while online dealing with unknown object models, noises, and disturbances.
The flowchart of the proposed method is shown in Figure~\ref{fig:system_configuration}.

\section{4. Modal-Graph Framework for DOM}
In this section, we present the construction and computation of the modal graph.
Then, based on the modal-graph framework, we derive the methods of feature extraction and inversion.

\subsection{4.1. Modal Graph Construction}
To begin with, we consider a graph $G(\mathbf{n},\mathcal{E})$ constructed from a discrete geometric model.
Note that we do not restrict the discrete geometric model to be element-based.
It can be a mesh or a particle system.
For a mesh, the graph nodes $\mathbf{n}$ and edges $\mathcal{E}$ are the vertices and the edges of the mesh elements.
For a partial system, the graph nodes $\mathbf{n}$ are the discrete particles, while the graph edges $\mathcal{E}$ store the nearest-neighbor relationships of the particles.

Afterward, without the need of identifying the object's physical properties, we assign the graph with arbitrary material parameters (including Young's modulus $E$, Poisson's ratio $v$, and the total mass $M$).
Using the finite element method (FEM) (\cite{zienkiewicz2005finite}) or an element-free modeling technique (such as \cite{belytschko1994element},
\cite{el1999new}), we can compute the stiffness matrix $\mathbf{K}(\mathbf{n}) \in \mathbb{R}^{3N \times 3N}$ and mass matrix $\mathbf{M}(\mathbf{n}) \in \mathbb{R}^{3N \times 3N}$ of the graph.
Then, we embed linear modal analysis in the graph.
The free vibration modes of $G(\mathbf{n},\mathcal{E})$ can  be computed by solving the following eigenproblem:
\begin{equation}
    \lambda \boldsymbol{\phi}(\mathbf{n}) = \mathbf{M}^{-1}(\mathbf{n}) \mathbf{K}(\mathbf{n}) \boldsymbol{\phi}(\mathbf{n})
\end{equation}
where 
$\boldsymbol{\phi}(\mathbf{n}) \in \mathbb{R}^{3N}$ is the eigenvector called the mode shape vector, which describes the deformation of all nodes $\mathbf{n}$ (with size $N$) in the mode; 
$\lambda$ is the eigenvalue whose square root corresponds to the natural frequency of the mode.
High-frequency modes are more susceptible to local measurement changes and noises but typically have little effect on the overall shape.
Thus, we can discard the high-frequency modes to find a global deformation description robust to the measurement irregularities.
Note that the modes are $\mathbf{M}$-orthonormal, but the mass matrix is computed with arbitrary mass parameters.
We assign the graph with the normalized modes.
These normalized modes are also orthogonal and frequency-ordered, which can form a unique and compact description for the deformation of the graph.

\begin{figure}[t]
    \centering
    \includegraphics[width=0.95\linewidth]{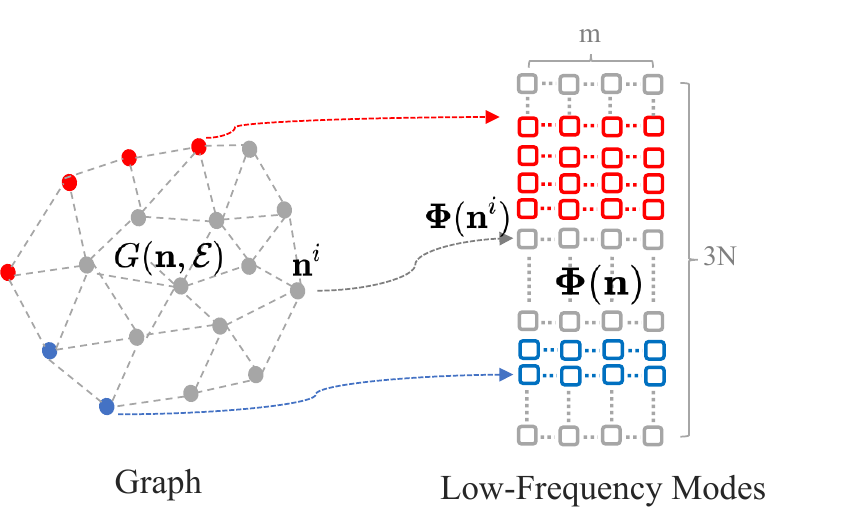}
    \caption{Illustrations of the node-wise modes.
    In our modal graph, each graph node $\mathbf{n}^i$ is associated with a set of low-frequency modes $\mathbf{\Phi}(\mathbf{n}^i)$.
    }
    \label{fig:node_wise}
\end{figure}

Assembling the normalized modes with the lowest $m$ frequencies, we obtain $\mathbf{\Phi}(\mathbf{n}) \in \mathbb{R}^{3N \times m}$, the low-frequency modes for $G(\mathbf{n},\mathcal{E})$.
Note that these modes are node-wise.
As shown in Figure~\ref{fig:node_wise}, 
each node $\mathbf{n}^i$ corresponds to a
sub-matrix $\mathbf{\Phi}(\mathbf{n}^i) \in \mathbb{R}^{3 \times m}$ (consisting of three rows in $\mathbf{\Phi}(\mathbf{n})$).
Finally, by assigning the node-wise modes to all of the graph nodes, we finish the construction of our modal graph.
We here give two examples of constructing a modal graph:

\subsubsection{From a geometric primitive:}
The first example is to construct a modal graph from a particle system discretized from a geometric primitive.
As superquadrics are widely used as primitives for 3D shape representation in computer graphics (\cite{solina1990recovery}) and vision (\cite{paschalidou2019superquadrics})
we consider a geometric primitive defined with the following superquadric surface:
\begin{equation}
    \boldsymbol{\gamma}(\mathbf{c}) = 
    \begin{bmatrix}
    a_x \cos^{\alpha_1}\zeta \cos^{\alpha_2}\sigma \\
    a_y \cos^{\alpha_1}\zeta \sin^{\alpha_2}\sigma \\
    a_z \sin^{\alpha_1}\zeta
\end{bmatrix}
\label{eq:global_elip}
\end{equation}
where
$\left \{ a_x, a_y, a_z \right \} $ are the size parameters in the x, y, and z dimension, respectively;
$\left \{ \alpha_1,\alpha_2 \right \}$ are the squareness parameters in the latitude and the longitude plane;
$\zeta \in [-\pi/2,\pi/2]$ and $\sigma \in [-\pi,\pi )$ correspond to the latitude and longitude in the ellipsoidal coordinates.
As discussed in \cite{solina1990recovery}, superquadrics define a flexible family of parametric shapes such as ellipsoids (with $\alpha_1 = \alpha_2 = 1$), cylinders (with $\alpha_1 \ll 1$ and $ \alpha_2 = 1$, and parallelepipeds (with $\alpha_1 \ll 1$ and $\alpha_2 \ll 1$).
After discretizing the primitive into a particle system, we construct a modal graph via the following procedures:
first, we assign the particle system with arbitrary mass and material properties;
second, we perform modal analysis on the particle system;
third, we assign the low-frequency modes to all of the particles.

In this paper, as we consider the setup where the geometric model of the object is totally unknown, we select this way to build our modal graph.
In addition, by using this parametric shape to construct a graph, we can online project object points to the graph boundary by simply computing the longitude and latitude coordinates of the point.

\subsubsection{From the object's rest shape measurements:}
The second example is to construct a modal graph from a mesh that is reconstructed using the object's rest shape measurements.
Given the point cloud measurements of the object at the rest configuration, we can roughly reconstruct a 3D shape using the method of \cite{solina1990recovery} or \cite{pentland1991closed}.
We discretize the shape into a mesh structure.
Using the moment-based method in \cite{pentland1991closed}, we can establish an ellipsoidal coordinate system for the mesh, and compute the latitude and longitude coordinates of the mesh vertices.
Then, we construct a modal graph via the following procedures:
first, we assign the mesh with arbitrary material properties;
second, we perform modal analysis on the meth;
third, we assign the low-frequency modes to all of the mesh vertices.
For a mesh-based graph, we online project object points to the graph boundary by searching the surface vertex with the closest latitude and longitude coordinates.

\begin{figure*}[t]
    \centering
    \includegraphics[width=0.95\linewidth]{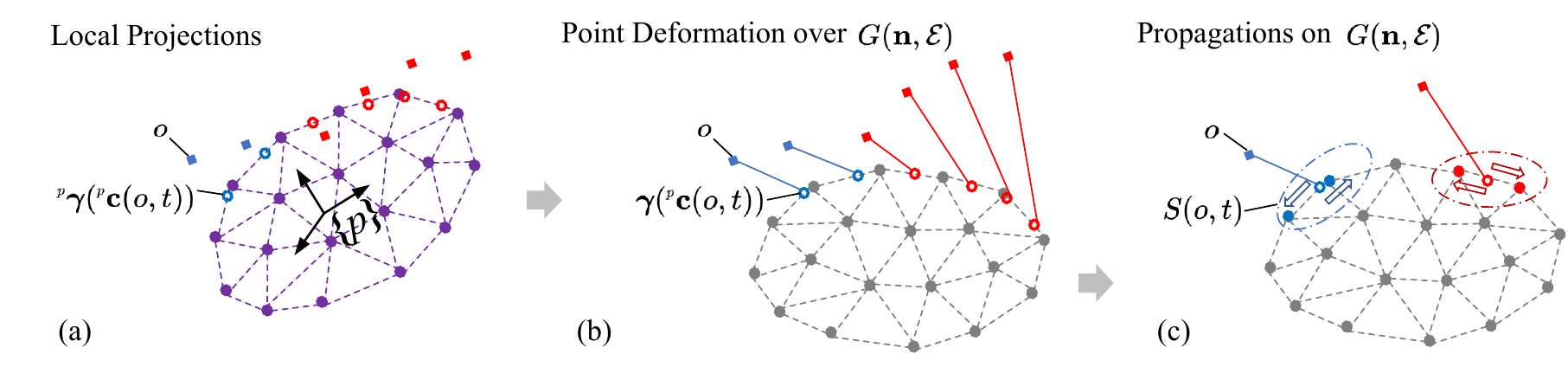}
    \caption{Illustrations of the point measurements organized and propagated on the graph.
    (a) the local projection ${}^{\scriptscriptstyle p}\boldsymbol{\gamma}({}^{\scriptscriptstyle p}\mathbf{c}(o,t))$) of points $o$ on a virtual graph located in the point cloud frame $\left \{ p \right \}$.
    (b) the point deformation between the point $o$ and its projection $\boldsymbol{\gamma}({}^{\scriptscriptstyle p}\mathbf{c}(o,t))$ on the graph $G(\mathbf{n},\mathcal{E})$.
    (c) the point deformation propagates to the nodes within $S(o,t)$, the point's domain of influence.}
    \label{fig:over_graph}
\end{figure*}

Finally, we conclude the structure of the modal graph.
In the modal graph, each node $\mathbf{n}^i$ is associated with:
a set of Cartesian coordinates $\mathbf{x}(\mathbf{n}^i) \in \mathbb{R}^{3}$ describing the absolute positions in the Euclidean space;
a set of parametric coordinates  $\mathbf{c} = \left \{ \zeta,\sigma \right \}^T \in \mathbb{R}^{2}$ (with $\zeta$ and $\sigma$ be the latitude and longitude of the 
ellipsoidal
coordinates), which describes the relative positions among the nodes;
a boundary indicator of whether the node is on the graph boundary;
and a set of mode shape vectors $\mathbf{\Phi}(\mathbf{n}^i)$ describing how the node deforms in low-frequency modes.
The Cartesian coordinates of the graph nodes and the spatial relationship stored in the graph edges $\mathcal{E}$ are used to compute the mode shape matrix $\mathbf{\Phi}(\mathbf{n})$ for the graph.
The parametric coordinates and the boundary indicators are used to compute the deformation propagation among the boundary nodes of the graph (the detailed derivation is in the subsequent subsection). 
The low-frequency modes of all the graph nodes (i.e., columns of $\mathbf{\Phi}(\mathbf{n})$) span a modal space where the propagated deformation can be described in a low- and fixed-dimensional manner.

\subsection{4.2. Computations on the Modal Graph}
We define the computations on the constructed modal graph such that the object shape can be represented with the low-frequency modes embedded in the graph.
Inspired by \cite{barr1987global} and \cite{pentland1987perceptual} that treat non-rigid shapes as deformed results from some reference shape,
we model the object shape using the displacement field of the object points over the graph.
For each object point $o$, we compute its 3D position $\mathbf{x}(o,t) \in \mathbb{R}^{3}$ over $G(\mathbf{n},\mathcal{E})$ with:
\begin{equation}
    \mathbf{x}(o,t) = \boldsymbol{\gamma}({}^{\scriptscriptstyle p}\mathbf{c}(o,t)) + \mathbf{u}_{\boldsymbol{\gamma}}(o,t)
\end{equation}
where
$\mathbf{u}_{\boldsymbol{\gamma}}(o,t) \in \mathbb{R}^{3}$ is the displacement vector between the point $o$ and $\boldsymbol{\gamma}({}^{\scriptscriptstyle p}\mathbf{c}(o,t)) \in \mathbb{R}^3$, its projection on the graph boundary.
We determine this projection using the point's local parametric coordinates ${}^{\scriptscriptstyle p}\mathbf{c}(o,t) \in \mathbb{R}^{2}$ 
(i.e., the latitude and longitude of the spherical coordinates computed under the point cloud frame $\left \{ p \right \}$).
As illustrated in Figure~\ref{fig:over_graph}(a,b), this pair of local parametric coordinates
corresponds to a local projection ${}^{\scriptscriptstyle p}\boldsymbol{\gamma}({}^{\scriptscriptstyle p}\mathbf{c}(o,t))$ on a virtual graph located in the point cloud frame $\left \{ p \right \}$, which reflects the spatial properties of $o$ among all the measured points.
Next, to establish the relationship between the point deformation $\mathbf{u}_{\boldsymbol{\gamma}}(o,t)$ and the deformation of graph nodes,
we define the propagation of $\mathbf{u}_{\boldsymbol{\gamma}}(o,t)$ among the graph nodes.

\subsubsection{4.2.1. Deformation Propagation among Graph Nodes}
As the point deformation applied on the graph boundary propagates to the nearby nodes, its effect will diminish with distance.
Thus, for each point, we consider a domain of influence that is determined by the support of an inverse-distance weight function formulated in the following exponential form (\cite{atluri2000new}):
\begin{equation}
    \omega(d_i(o,t))\!=\!
    \left\{
    \begin{aligned}
    & \frac{e^{-(d_i(o,t)/c^2)} - e^{-(d_s/c^2)} }{ 1 -e^{-(d_s/c^2)} }   &   , d_i \le d_s     \\
  & 0 & ,  d_i > d_s
   \end{aligned}
\right.
\label{eq:weight}
\end{equation}
where 
the constant $c$ controls the shape of the weight function;
$d_i(o,t)$ is defined as a parametric distance vector computed with:
\begin{equation}
    d_i(o,t) = ({}^{\scriptscriptstyle p}\mathbf{c}(o,t) - \mathbf{c}(\mathbf{n}^{i}))^T
    ({}^{\scriptscriptstyle p}\mathbf{c}(o,t) - \mathbf{c}(\mathbf{n}^{i}));
\end{equation}
the positive scalar $d_s$ is the support size of the weight function. 
$d_s$ can be computed via:
\begin{equation}
    d_s = r_s \overline{d}(\mathcal{E})
    \label{eq:support_size_parameter}
\end{equation}
where the support size parameter $r_s > 0$ 
and $\overline{d}(\mathcal{E})$ is the average connection distance of the graph.
Note that within the support, the weight function is positive and continuous.

\newtheorem{remark}{Remark}
\begin{remark}
The weight function can be used to compute rigidity because it satisfies the following properties (\cite{berenson2013manipulation}):
\begin{enumerate}
    \item $\omega(d_i(o,t)) \in \left [ 0, 1  \right ] $;
    \item $\omega(0) = 0$;
    \item $\omega(d_i(o,t)) > \omega(d_j(o,t))$, if $d_i < d_j$.
\end{enumerate}
As the support size parameter $r_s$ tends to $+\infty$, the weight function becomes:
\begin{equation}
    \omega(d_i(o,t)) = e^{-(d_i(o,t)/c^2)}
    \label{eq:weight_ifty}
\end{equation}
which means that the point deformation is propagated over (all of) the graph nodes in a \textit{diminishing rigidity} (\cite{berenson2013manipulation}) manner.
\end{remark}

Then, using the Shepard's method (\cite{shepard1968two}, \cite{nguyen2008meshless}), a well-known inverse-distance weighted formula for scattered data interpolation, we model the point deformation $\mathbf{u}_{\boldsymbol{\gamma}}(o,t)$ as a weighted collection of the displacements of the propagated nodes:
\begin{equation}
\begin{aligned}
    \mathbf{u}_{\boldsymbol{\gamma}}(o,t) 
    & = \sum\limits_{\scriptscriptstyle{i \in S(o,t)}}^{}
    \frac{\omega(d_i(o,t))}{\sum\limits_{\scriptscriptstyle{i \in S(o,t)}} \!\!\! \omega(d_i(o,t))}
    \mathbf{u}(\mathbf{n}^i, t)
\end{aligned}
    \label{eq:point_distribution}
\end{equation}
where 
$S(o)$ is the domain of influence of $o$ on $\partial G(\mathbf{n},\mathcal{E})$ (i.e., the graph boundary) where $\omega(d_i(o,t)) \ne 0$;
$\mathbf{u}(\mathbf{n}^i, t) \in \mathbb{R}^{3}$ is the displacement vector of $\mathbf{n}^i$.
The next step is to find the modal representation of these nodal displacements.

\begin{remark}
The above deformation propagation method provides a simple and efficient way to model point positions without requiring point-wise correspondences. 
It should be mentioned that the formulations are not physically accurate.
We will further address the modeling uncertainties and unmodeled effects by designing adaptive robust control laws (in the subsequent controller design section).
\end{remark}

\subsubsection{4.2.2. Modal Representation:}
For each node $\mathbf{n}^i$ in $S(o)$, 
as $det(\mathbf{\Phi}(\mathbf{n})) \ne 0$,
its associated modes $\mathbf{\Phi}(\mathbf{n}_i)$ specify a transformation from 
the modal space of $G(\mathbf{n},\mathcal{E})$ to the nodal displacement:
\begin{equation}
\begin{matrix} 
\mathbf{u}(\mathbf{n}^i,t) = \mathbf{\Phi}(\mathbf{n}^i)\mathbf{s}(t)
\end{matrix}
\end{equation}
where the modal coefficients $\mathbf{s}(t) \in \mathbb{R}^m$.
Using such node-wise modal transformation,
we can further represent the point measurements $\mathbf{x}(o,t)$ into the following modal combinations:
\begin{equation}
  \begin{aligned}
       \mathbf{x}(o,t) = 
       \boldsymbol{\gamma}({}^{\scriptscriptstyle p}\mathbf{c}(o,t)) 
       +
     \sum\limits_{\scriptscriptstyle{i \in S(o,t)}}^{} \!
    \frac{\omega(d_i(o,t))}{\sum\limits_{\scriptscriptstyle{i \in S(o,t)}} \!\!\! \omega(d_i(o,t))}
    \mathbf{\Phi}(\mathbf{n}_i)
    \mathbf{s}(t).
  \end{aligned}
  \label{eq:graph_computation}
\end{equation}

\subsection{4.3. Feature Extraction}
Based on the above discussion, the 3D shape of the object can be modeled with a combination of the orthonormal, low-frequency deformation modes.
We adopt the modal coefficients $\mathbf{s}(t)$ to be our deformation features.
In this subsection, we present the method to compute $\mathbf{s}(t)$ from raw point cloud measurements.

Note that during the object deformation, the measured raw point clouds $\mathbf{p}(t)$ have time-varying dimensions, and their point-wise correspondences are unknown.
In this way, at each moment $t$, we need to re-establish the relationships between the measured points and the graph.
For this purpose, we online perform the above graph computations (equation~\eqref{eq:graph_computation}) for the measured points $\mathbf{p}(t)$. 
We have:
\begin{equation}
   \begin{aligned}
      \mathbf{x}(\mathbf{p}(t),t) 
      = &
      \boldsymbol{\gamma}({}^{\scriptscriptstyle p}\mathbf{c}(\mathbf{p}(t),t)) \\
      & +
       \sum\limits_{\scriptscriptstyle{i \in S(\mathbf{p}(t),t)}}^{} \!
    \frac{\omega(d_i(\mathbf{p}(t),t))}{\sum\limits_{\scriptscriptstyle{i \in S(\mathbf{p}(t),t)}} \!\!\! \omega(d_i(\mathbf{p}(t),t))}
    \mathbf{\Phi}(\mathbf{n}_i)
    \mathbf{s}(t) \\
       = & \boldsymbol{\gamma}({}^{\scriptscriptstyle p}\mathbf{c}(\mathbf{p}(t),t)) + \mathbf{\Psi}(\mathbf{p}(t),\mathbf{n}_p(t))  \mathbf{\Phi}(\mathbf{n}_p(t))\mathbf{s}(t)
   \end{aligned}
    \label{eq:shape_modeling}
\end{equation}
where 
$\boldsymbol{\gamma}({}^{\scriptscriptstyle p}\mathbf{c}(\mathbf{p}(t),t)) \in \mathbb{R}^{3l(t)}$ are the projections of $\mathbf{p}(t)$ on the graph boundary;
$\mathbf{n}_p(t) = \left \{ \mathbf{n}^i \mid \mathbf{n}^i \in S(\mathbf{p}(t),t) \right \}  $ are the supporting nodes of $\mathbf{p}(t)$, whose displacements are denoted by $\mathbf{u}(\mathbf{n}_p(t),t) \in \mathbb{R}^{3n(t)}$;
the shape matrix $\mathbf{\Psi}(\mathbf{p}(t),\mathbf{n}_p(t))$ $\in \mathbb{R}^{3l(t) \times 3n(t)}$ establishes the deformation of $\mathbf{p}(t)$ structured on the graph nodes $\mathbf{n}_p(t)$.

Given the time-varying dimensions of $\mathbf{p}(t)$ and $\mathbf{n}_p(t)$, the direct inverse of the equation~\eqref{eq:shape_modeling} may be under-determined when $l(t) < n(t)$.
To estimate the modal states without dealing with the possible under-determined problems,
we consider an attractive potential function that attracts the graph to fit to the point cloud measurements.
The potential function is constructed with the following paraboloidal form:
\begin{equation}
\begin{aligned}
    U(\boldsymbol{\gamma}({}^{\scriptscriptstyle p}\mathbf{c}(\mathbf{p}(t),t))) = & \frac{1}{2}(\mathbf{x}(\mathbf{p}(t),t) - \boldsymbol{\gamma}({}^{\scriptscriptstyle p}\mathbf{c}(\mathbf{p}(t),t))^T  \\
   & \cdot (\mathbf{x}(\mathbf{p}(t),t) - \boldsymbol{\gamma}({}^{\scriptscriptstyle p}\mathbf{c}(\mathbf{p}(t),t)).
\end{aligned}
\end{equation}
Afterward, we propagate the resulting attractive force field $ -\nabla U(\boldsymbol{\gamma}({}^{\scriptscriptstyle p}\mathbf{c}(\mathbf{p}(t),t)))$ applied on $\boldsymbol{\gamma}({}^{\scriptscriptstyle p}\mathbf{c}(\mathbf{p}(t),t))$ to the nearby graph nodes via:
\begin{equation}
    \mathbf{f}(\mathbf{n}_p(t),t) = -\mathbf{\Psi}^T(\mathbf{p}(t),\mathbf{n}_p(t))\nabla U(\boldsymbol{\gamma}({}^{\scriptscriptstyle p}\mathbf{c}(\mathbf{p}(t),t)))
\end{equation}
where $\mathbf{f}(\mathbf{n}_p(t),t) \in \mathbb{R}^{3n(t)}$ are the propagated forces of $\mathbf{n}_p(t)$;
Then, by further adopting the modal techniques proposed by \cite{pentland1991closed}, we compute the modal displacements introduced by $\mathbf{f}(\mathbf{n}_p(t),t)$ with:
\begin{equation}
    \begin{aligned}
        \mathbf{s}(t)  = & 
        (\widetilde{\mathbf{K}}+\mathbf{I}_6)^{-1}\mathbf{\Phi}^T(\mathbf{n}_p(t)) \mathbf{f}(\mathbf{n}_p(t),t)
        \\ = &(\widetilde{\mathbf{K}}+\mathbf{I}_6)^{-1}\mathbf{\Phi}^T(\mathbf{n}_p(t))\mathbf{\Psi}^T(\mathbf{p}(t),\mathbf{n}_p(t)) 
        \mathbf{u}_{\boldsymbol{\gamma}}(\mathbf{p}(t),t)
    \end{aligned}
    \label{eq:feature_computation}
\end{equation}
where 
$\mathbf{\Phi}(\mathbf{n}_p(t)) \in \mathbb{R}^{3n(t) \times m}$ consists of the deformation modes associated with $\mathbf{n}_p(t)$;
the modal stiffness matrix $\widetilde{\mathbf{K}}=\mathbf{\Phi}^T(\mathbf{n})\mathbf{K}(\mathbf{n})\mathbf{\Phi}(\mathbf{n}) \in \mathbb{R}^{m \times m}$;
$\mathbf{I}_{6} \in \mathbb{R}^{m \times m}$ is a diagonal matrix whose first six diagonal elements are ones, while the other elements are zeros.

Note that the nonlinear feature extraction function (equation~\eqref{eq:feature_computation}) is formulated in a position-based manner.
All the computations are online structured on the graph
using the position measurements of the points.
We do not need to know the geometric relationships (such as the spatial connections and orders) among the points, nor their one-to-one correspondences between times.
We also do not need to know the physical and geometric models of the object.
We emphasize that our deformation features are computed not to recover the object shape accurately but to obtain a compact shape description from raw point cloud measurements.
The rank of the nonlinear feature extraction function satisfies:
$$
rank((\widetilde{\mathbf{K}}+\mathbf{I}_6)^{-1}\mathbf{\Psi}^T(\mathbf{p}(t),\mathbf{n}_p(t))\mathbf{\Phi}_n^T(\mathbf{n}_p(t))) = m.
$$
Thus, unique and over-constrained solutions can be obtained even for partial or occluded observations as long as the feature dimension is properly selected such that $m \le \min\left \{3l(t),3n(t)\right \}$.

\begin{remark}
As the modal graph is in the discrete structure, changes of the node set $\mathbf{n}_p(t)$ will make the feature extraction function (equation~\eqref{eq:feature_computation}) discontinuous to the point measurements $\mathbf{x}(\mathbf{p}(t),t)$.
However, such discontinuity on the graph can be eliminated by setting the supporting size parameter $r_s$ to a finite value at $t_0$, and then increasing it to $+\infty$ after the object is deformed.
\end{remark}

\begin{figure*}[t]
    \centering
    \includegraphics[width=0.8 \linewidth]{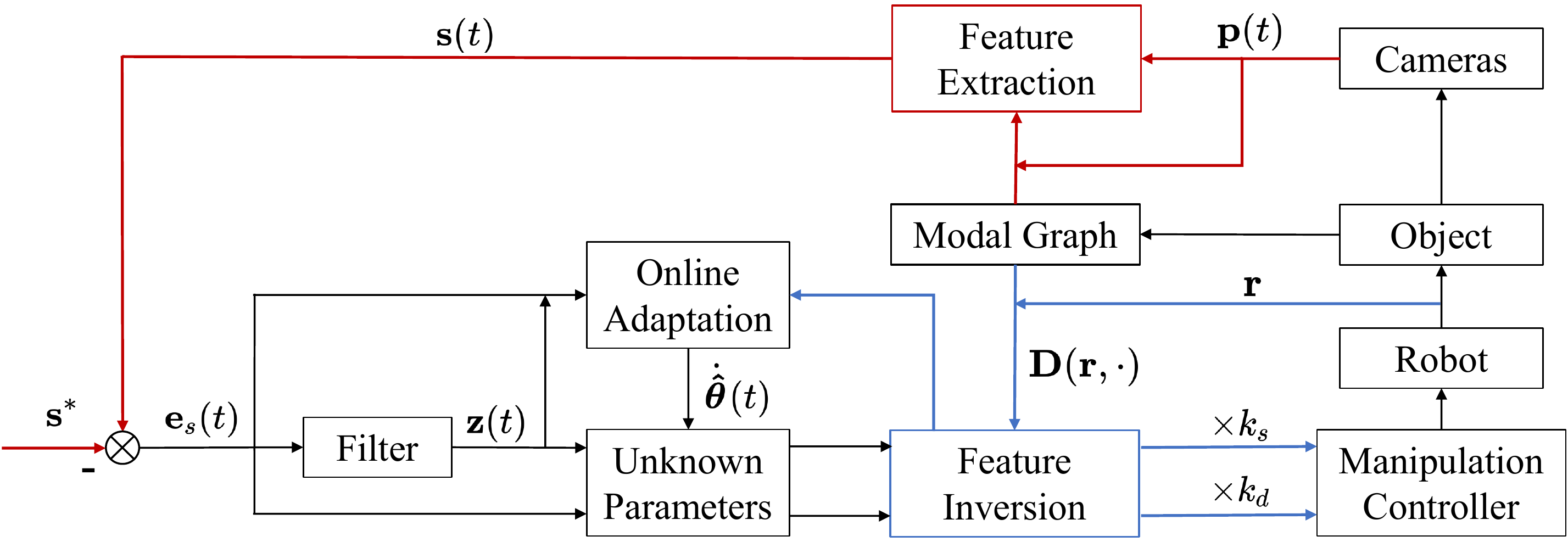}
    \caption{The block diagram of the proposed controller. The red parts are for the feature extraction, while the blue parts are for the feature inversion.}
    \label{fig:controller}
\end{figure*}

\subsection{4.4. Feature Inversion}
In our robotic DOM system,
the object is manipulated via the manipulation points $\mathbf{r}$.
Based on the modal-graph computations, we can model the point positions $\mathbf{x}(\mathbf{r},t) \in \mathbb{R}^{3k}$ into a weighted collection of the displacements of the graph nodes:
\begin{equation}
\begin{aligned}
    \mathbf{x}(\mathbf{r},t) 
      & = \boldsymbol{\gamma}({}^{\scriptscriptstyle p}\mathbf{c}(\mathbf{r},t)) 
      + 
    \sum\limits_{\scriptscriptstyle{i \in S(\mathbf{r})}}^{} \!
    \frac{\omega(d_i(\mathbf{r},t))}{\sum\limits_{\scriptscriptstyle{i \in S(\mathbf{r})}} \!\!\! \omega(d_i(\mathbf{r},t))}
    \mathbf{u}(\mathbf{n}^i,t) \\
    & = \boldsymbol{\gamma}({}^{\scriptscriptstyle p}\mathbf{c}(\mathbf{r},t)) + \mathbf{\Psi}(\mathbf{r},\mathbf{n}_r)\mathbf{u}(\mathbf{n}_r,t)
\end{aligned}
\label{eq:manipulation_modeling}
\end{equation}
where 
$\boldsymbol{\gamma}({}^{\scriptscriptstyle p}\mathbf{c}(\mathbf{r})) \in \mathbb{R}^{3k}$ are the projections of $\mathbf{r}$ on the graph boundary,
and $S(\mathbf{r})$ is the domain of influence of $\mathbf{r}$.
Unlike the raw point clouds, with respect to Assumption~\ref{as:point_manipulation}, $\mathbf{r}$ are fixed object points whose positions can be tracked using robot measurements.
Thus, we only compute them at $t_0$.
During the subsequent deformation, we compute the deformation of $\mathbf{r}$ over the graph with:
\begin{equation}
 \mathbf{u}_{\boldsymbol{\gamma}}(\mathbf{r},t) 
   = \mathbf{x}(\mathbf{r},t) - \boldsymbol{\gamma}({}^{\scriptscriptstyle p}\mathbf{c}(\mathbf{r},t_0)) 
  = \mathbf{\Psi}(\mathbf{r},\mathbf{n}_r)\mathbf{u}(\mathbf{n}_r,t)
\end{equation}
where
$\mathbf{n}_r = \left \{ \mathbf{n}^i \mid \mathbf{n}^i \in S(\mathbf{r}) \right \}$ is the supporting node set of $\mathbf{r}$, whose displacements are denoted by $\mathbf{u}(\mathbf{n}_r,t) \in \mathbb{R}^{3h}$;
the shape matrix $\mathbf{\Psi}(\mathbf{r},\mathbf{n}_r) \in \mathbb{R}^{3k \times 3h}$.
Given $\mathbf{n}_r$, we assemble the matrix $\mathbf{\Phi}(\mathbf{n}_r) \in \mathbb{R}^{3h \times m}$, which describes how the nodes $\mathbf{n}_r$ deforming in the low-frequency modes.
Using these matrices, we propose the following feature inversion function:
\begin{equation}
  \mathbf{D}(\mathbf{r},\Delta \mathbf{s}(t)) 
  = \mathbf{\Psi}(\mathbf{r},\mathbf{n}_r)\mathbf{\Phi}(\mathbf{n}_r) \Delta \mathbf{s}(t).
\end{equation}
The smooth function $\mathbf{D}(\mathbf{r},\cdot): \mathbb{R}^m \to \mathbb{R}^{3k}$ inverses the feature changes ($\Delta \mathbf{s}(t) \in \mathbb{R}^m$) to the manipulation motions $\Delta \mathbf{u}_{\boldsymbol{\gamma}}(\mathbf{r},t) = \Delta \mathbf{x}(\mathbf{r},t) \in \mathbb{R}^{3k}$.

\section{5. Adaptive Robust Controller}
Based on the modal-graph framework,
we design an adaptive robust controller with feedback structures of the deformation feature errors $\mathbf{e}_s (t)$.
As shown in the control block diagram (Figure~\ref{fig:controller}),
the modal graph is used to extract and invert deformation features, while the adaptive robust laws are used to deal with modeling uncertainties and disturbances online.

\subsection{5.1. System Modeling}
For slow robot manipulation under kinematic motion control, 
only the kinematic model of the system is investigated in our method.
We consider an intermediate variable:
the modal displacement vector ${\widetilde{\mathbf{u}}}_r(t) \in \mathbb{R}^{m}$ 
quantifying the deformation of the modal graph (rather than the object) under the robot manipulation.
The kinematic relations between the modal velocities ${\dot{\widetilde{\mathbf{u}}}}_r(t) \in \mathbb{R}^{m}$ and the manipulation velocities $\mathbf{v}(t) = \dot{\mathbf{x}}(\mathbf{r},t) \in \mathbb{R}^{3k}$
can be locally approximated by:
\begin{equation}
    {\dot{\widetilde{\mathbf{u}}}}_r(t)  =  (\widetilde{\mathbf{K}}+\mathbf{I}_6)^{-1}
\mathbf{\Phi}^T(\mathbf{n}_r)
\mathbf{\Psi}^T(\mathbf{r},\mathbf{n}_r) \mathbf{v}(t)
    \label{eq:modal_flow}
\end{equation}
where $\mathbf{\Psi}(\mathbf{r},\mathbf{n}_r)$ transforms the velocities of the manipulation points $\mathbf{r}$ to the velocities of the graph nodes $\mathbf{n}_r$, while 
$(\widetilde{\mathbf{K}}+\mathbf{I}_6)^{-1}
\mathbf{\Phi}^T(\mathbf{n}_r)$ transform the velocities of $\mathbf{n}_r$ to the modal velocities.
Note that the matrices $\widetilde{\mathbf{K}}$ and  $\mathbf{\Phi}(\mathbf{n}_r)$ are computed with the deformation model of the graph rather than the object.
Also, the formulation of $\mathbf{\Psi}(\mathbf{r},\mathbf{n}_r)$ (determined by the deformation propagation method presented in Section 4.2.1) is not physically accurate.
We must further consider these modeling uncertainties and unmodeled effects.
For this purpose, we introduce the following parametric and non-parametric uncertainty terms, and model the system kinematics with:
\begin{equation}
    \dot{\mathbf{s}}(t) = \mathbf{Q}(\boldsymbol{\theta})\dot{\widetilde{\mathbf{u}}}_r(t) + \boldsymbol{\eta}(t)
    \label{eq:raw_model}
\end{equation}
where the parametric term $\mathbf{Q}(\boldsymbol{\theta})  = \diag(\boldsymbol{\theta})  \in \mathbb{R}^{m \times m}$ is a diagonal matrix of the unknown modal parameters $\boldsymbol{\theta} \in \mathbb{R}^m$
(\cite{10122176}) reflecting the uncertainties of the deformation models between the graph and the object in the modal space,
and the non-parametric term $\boldsymbol{\eta}(t) \in \mathbb{R}^m$ reflects unmodeled effects and sensor noises.
Combining with equation~\eqref{eq:modal_flow}, we rewrite equation~\eqref{eq:raw_model} into the following form:
\begin{equation}
    \dot{\mathbf{s}}(t)
    = \mathbf{W}(\boldsymbol{\theta})
    \mathbf{H}^T(\mathbf{r},\mathbf{n}_r)
    \mathbf{v}(t) + \boldsymbol{\eta}(t)
    \label{eq:system_ma}
\end{equation}
where 
the notation $\mathbf{H}(\mathbf{r},\mathbf{n}_r) = 
\mathbf{\Psi}(\mathbf{r},\mathbf{n}_r)
\mathbf{\Phi}(\mathbf{n}_r)$ is formulated for the sake of simplicity in the rest derivations;
the diagonal matrix 
$\mathbf{W}(\boldsymbol{\theta}) = \mathbf{Q}(\boldsymbol{\theta})(\widetilde{\mathbf{K}}+\mathbf{I}_6)^{-1} \in \mathbb{R}^{m \times m}$.
In addition,
as discussed, the measurement irregularities and noises mainly influence the high-frequency deformation modes.
Therefore, for slow and quasi-static robot motion, the non-parametric uncertainties $\boldsymbol{\eta}(t)$ defined in the low-frequency modal space are bounded.

\subsection{5.2. Kinematic Control Law}
To cope with the parametric and non-parametric uncertainties in the system model (equation~\eqref{eq:system_ma}), we propose the following control law to generate the manipulation velocity commands:
\begin{equation}
    \mathbf{v}(t) 
    = \dot{\mathbf{x}}(\mathbf{r},t)
    = \mathbf{v}_{s}(t) + \mathbf{v}_{d}(t)
    \label{eq:v_cmd}
\end{equation}
which consists of two parts: the nominal dynamic state feedback term $\mathbf{v}_{s}(t) \in \mathbb{R}^{3k}$, and the robust feedback term $\mathbf{v}_{d}(t) \in \mathbb{R}^{3k}$ to dissipate the time-varying non-parametric uncertainties.
\subsubsection{5.2.1. Nominal Dynamic State Feedback Term:}
As changes of $\mathbf{p}(t)$ and $\mathbf{n}_p(t)$ will cause the changes of the feature value, we adopt a dynamic extension of $\mathbf{e}_s$ using the following filter-like structure to smooth the deformation feature errors $\mathbf{e}_s(t)$:
\begin{equation}
    \mathbf{z}(t) = -k \dot{\mathbf{z}}(t) + \mathbf{e}_s(t)
\end{equation}
where $\mathbf{z}(t) \in \mathbb{R}^m$ is the dynamic extension of $\mathbf{e}_s(t)$; 
the scalar $k$ is defined by:
$$k = \frac{1-q}{q} $$ 
with the smooth gain $q \in (0,1]$.
Using $\mathbf{z}(t)$, we propose the nominal dynamic state feedback term:
\begin{equation}
    \mathbf{v}_{s}(t) = - k_s \mathbf{D}(\mathbf{r},\mathbf{W}^T({\boldsymbol{\hat\theta}}(t))\mathbf{z}(t))
\end{equation}
where the feedback gain $k_s > 0$;
$\mathbf{W}({\boldsymbol{\hat\theta}}(t))$ is computed using the online adapted modal parameters ${\boldsymbol{\hat\theta}}(t)$.

\subsubsection{5.2.2. Robust Feedback Term:}
To dissipate the disturbances of the non-parametric uncertainties $\boldsymbol{\eta}(t)$, we propose the following robust feedback term:
\begin{equation}
    \mathbf{v}_{d}(t) = - k_d
    \mathbf{D}(\mathbf{r},\mathbf{W}^T({\boldsymbol{\hat\theta}}(t))\mathbf{e}_s(t))
\end{equation}
where the dissipating gain $k_d > 0$.

Based on the above derivations, the closed-loop error dynamics of the system can be formulated as:
\begin{equation}
  \begin{aligned}
      \dot{\mathbf{e}}_s(t)
       = & -\mathbf{W}(\boldsymbol{\theta})\mathbf{H}^T(\mathbf{r},\mathbf{n}_r) [k_s \mathbf{D}(\mathbf{r},\mathbf{W}^T({\boldsymbol{\hat\theta}}(t))\mathbf{z}(t)) +  
       \\ & k_d \mathbf{D}(\mathbf{r},\mathbf{W}^T({\boldsymbol{\hat\theta}}(t))\mathbf{e}_s(t))] + \boldsymbol{\eta}(t) 
  \end{aligned}
  \label{eq:closed_loop}
\end{equation}
\begin{equation}
    \dot{\mathbf{z}}(t) = \frac{1}{k}(-\mathbf{z}(t) + \mathbf{e}_s(t))
    \label{eq:closed_z}
\end{equation}

\subsection{5.3. Online Parameter Adaptation}
To cope with the parametric uncertainties,
we propose an adaptive law for the unknown modal parameters $\boldsymbol{\theta}$, such that the parameter estimation errors in the closed loop can be compensated online.
For this purpose, we first add a zero term to the closed-loop error dynamics in equation~\eqref{eq:closed_loop} and rewrite it into:
\begin{equation}
  \begin{aligned}
      \dot{\mathbf{e}}_s&(t) \!
       = 
      [-\mathbf{W}({\boldsymbol{\hat\theta}}(t)) + (\mathbf{W}({\boldsymbol{\hat\theta}}(t)) - \mathbf{W}(\boldsymbol{\theta}))]\mathbf{H}^T(\mathbf{r},\mathbf{n}_r)
        \\ \cdot
       & [k_s\mathbf{D}(\mathbf{r},\mathbf{W}^T({\boldsymbol{\hat\theta}}(t))\mathbf{z}(t))
      +  k_d\mathbf{D}(\mathbf{r},\mathbf{W}^T({\boldsymbol{\hat\theta}}(t))\mathbf{e}_s(t))] + \boldsymbol{\eta}(t)
  \end{aligned}
  \label{eq:closed_regression}
\end{equation}
Then, we define the following parameter errors $\Delta \boldsymbol{\theta}(t) \in \mathbb{R}^m$ with:
\begin{equation}
    \Delta \boldsymbol{\theta}(t) = \boldsymbol{\hat{\theta}}(t) - \boldsymbol{\theta}
\end{equation}
which can be used to linearize the parametric estimation errors in equation \eqref{eq:closed_regression}. 
We have:
\begin{equation}
   \begin{aligned}
      (\mathbf{W}({\boldsymbol{\hat\theta}}(t)) - \mathbf{W}(\boldsymbol{\theta}))\mathbf{H}^T(\mathbf{r},\mathbf{n}_r)
    [k_s\mathbf{D}(\mathbf{r},\mathbf{W}^T({\boldsymbol{\hat\theta}}(t))\mathbf{z}(t)) \\
      +  k_d\mathbf{D}(\mathbf{r},\mathbf{W}^T({\boldsymbol{\hat\theta}}(t))\mathbf{e}_s(t))] 
    =\mathbf{Y}(\mathbf{e}_s(t),
    \mathbf{z}(t),
    \boldsymbol{\hat{\theta}}(t) )\Delta\boldsymbol{\theta}(t) 
   \end{aligned}
   \label{eq:closed_linear}
\end{equation}
where the regression matrix $\mathbf{Y}(\mathbf{e}_s(t),
\mathbf{z}(t),
\hat{\boldsymbol{\theta}}(t)) \in \mathbb{R}^{m \times m}$ is independent of the unknown parameters $\boldsymbol{\theta}$.
Then, using this regression matrix, we propose the following parameter adaptation law:
\begin{equation}
    \boldsymbol{\dot{\hat{\theta}}}(t)=-\Gamma^{-1}
    \mathbf Y^T(\mathbf{e}_s(t) ,
    \mathbf{z}(t),
    \boldsymbol{\hat{\theta}}(t) ){\mathbf e}_s(t)
    \label{eq:theta_dot}
\end{equation}
where $\Gamma$ is a positive definite gain for the parameter adaptations.

\subsection{5.4. Stability Analysis}
To analyze the stability of the closed-loop system (equation~\eqref{eq:closed_loop} and \eqref{eq:closed_z}) of $\mathbf{e_s}$ with disturbances,
we use the ISS-Lyapunov-based method (\cite{sontag2008input}) with the following Lyapunov candidate function:
\begin{equation}
    V(t) = \frac{1}{2}\mathbf{e}_s^T(t)\mathbf{e}_s(t) + \frac{\Gamma}{2}\Delta \boldsymbol{\theta}^T(t)\Delta \boldsymbol{\theta}(t).
\end{equation}
First, we consider:
\begin{equation}
    V_e(t) = \frac{1}{2}\mathbf{e}_s^T(t)\mathbf{e}_s(t).
    \label{eq:V_e}
\end{equation}
Differentiating it with respect to time, we have:
\begin{equation}
\begin{aligned}
    \dot{V}_e(t) 
     = & \mathbf{e}_s^T(t) \dot{\mathbf{e}}_s(t) \\
     = & - \mathbf{e}_s^T(t)\mathbf{W}(\boldsymbol{\theta})\mathbf{H}^T(\mathbf{r},\mathbf{n}_r) [k_s \mathbf{D}(\mathbf{r},\mathbf{W}^T({\boldsymbol{\hat\theta}}(t))\mathbf{z}(t))   
       \\ & + k_d \mathbf{D}(\mathbf{r},\mathbf{W}^T({\boldsymbol{\hat\theta}}(t))\mathbf{e}_s(t))] +  \mathbf{e}_s^T(t) \boldsymbol{\eta}(t). 
\end{aligned}
\label{eq:raw_ved}
\end{equation}
Extracting the parametric estimation errors, we rewrite equation~\eqref{eq:raw_ved} into:
\begin{equation}
\begin{aligned}
    \dot{V}_e(t) 
    = & - \mathbf{e}_s^T(t)\mathbf{W}({\boldsymbol{\hat\theta}}(t))\mathbf{H}^T(\mathbf{r},\mathbf{n}_r) [k_s \mathbf{D}(\mathbf{r},\mathbf{W}^T({\boldsymbol{\hat\theta}}(t))\mathbf{z}(t))  
       \\ & + k_d \mathbf{D}(\mathbf{r},\mathbf{W}^T({\boldsymbol{\hat\theta}}(t))\mathbf{e}_s(t))] +  \mathbf{e}_s^T(t) \boldsymbol{\eta}(t)  \\
      & + \mathbf{e}_s^T(t)(\mathbf{W}({\boldsymbol{\hat\theta}}(t)) - \mathbf{W}(\boldsymbol{\theta}))[k_s \mathbf{D}(\mathbf{r},\mathbf{W}^T({\boldsymbol{\hat\theta}}(t))\mathbf{z}(t))   
       \\ & + k_d \mathbf{D}(\mathbf{r},\mathbf{W}^T({\boldsymbol{\hat\theta}}(t))\mathbf{e}_s(t))].
\end{aligned}
\label{eq:dv_e}
\end{equation}
Second, we consider:
\begin{equation}
    V_\theta(t) = \frac{\Gamma}{2}\Delta \boldsymbol{\theta}^T(t)\Delta \boldsymbol{\theta}(t).
\end{equation}
Differentiating $V_\theta(t)$ with respect to time, and combining $\dot{V}_\theta(t)$ with equations~\eqref{eq:closed_linear} and \eqref{eq:theta_dot}, we have:
\begin{equation}
\begin{aligned}
    \dot{V}_\theta(t) 
     = & \Gamma \boldsymbol{\dot{\hat{\theta}}}^T\!\!(t)\Delta \boldsymbol{\theta}(t)
     = - \mathbf{e}_s^T(t)\mathbf{Y}(\mathbf{e}_s(t) ,
     \mathbf{z}(t),
     \boldsymbol{\hat{\theta}}(t) )\Delta\boldsymbol{\theta}(t) \\
     = & - \mathbf{e}_s^T(t)(\mathbf{W}({\boldsymbol{\hat\theta}}(t)) - \mathbf{W}(\boldsymbol{\theta}))[k_s \mathbf{D}(\mathbf{r},\mathbf{W}^T({\boldsymbol{\hat\theta}}(t))\mathbf{z}(t))   
       \\ &  + k_d \mathbf{D}(\mathbf{r},\mathbf{W}^T({\boldsymbol{\hat\theta}}(t))\mathbf{e}_s(t))].
\end{aligned}
\label{eq:dv_theta}
\end{equation}
Next, by adding equations~\eqref{eq:dv_e} and \eqref{eq:dv_theta}, we have:
\begin{equation}
\begin{aligned}
    \dot{V}(t)  = & \dot{V}_e(t) + \dot{V}_\theta(t) \\
                = & -k_s\mathbf{e}_s^T(t)\mathbf{W}({\boldsymbol{\hat\theta}}(t))\mathbf{H}^T(\mathbf{r},\mathbf{n}_r)\mathbf{H}(\mathbf{r},\mathbf{n}_r)\mathbf{W}^T({\boldsymbol{\hat\theta}}(t))\mathbf{z}(t) \\
               & - k_d\mathbf{e}_s^T(t)\mathbf{W}({\boldsymbol{\hat\theta}}(t))\mathbf{H}^T(\mathbf{r},\mathbf{n}_r)\mathbf{H}(\mathbf{r},\mathbf{n}_r)\mathbf{W}^T({\boldsymbol{\hat\theta}}(t))\mathbf{e}_s(t) \\
               & + \mathbf{e}_s^T(t) \boldsymbol{\eta}(t). 
\end{aligned}
\label{eq:dv_raw}
\end{equation}
Let $X_1(t) = \mathbf{e}_s^T(t)\mathbf{W}({\boldsymbol{\hat\theta}}(t))\mathbf{H}^T(\mathbf{r},\mathbf{n}_r)\mathbf{H}(\mathbf{r},\mathbf{n}_r)\mathbf{W}^T({\boldsymbol{\hat\theta}}(t))\mathbf{z}(t)$.
According to the closed-loop error dynamics in equation~\eqref{eq:closed_z}, $X_1(t)$ can be rewritten into the following form by completing the squares:
\begin{equation}
\begin{aligned}
    & X_1(t)  
     = \frac{1}{2}\mathbf{e}_s^T(t)\mathbf{W}({\boldsymbol{\hat\theta}}(t))\mathbf{H}^T(\mathbf{r},\mathbf{n}_r)\mathbf{H}(\mathbf{r},\mathbf{n}_r)\mathbf{W}^T({\boldsymbol{\hat\theta}}(t))\mathbf{e}_s(t) \\
     & +  \frac{1}{2}\mathbf{z}^T(t)\mathbf{W}({\boldsymbol{\hat\theta}}(t))\mathbf{H}^T(\mathbf{r},\mathbf{n}_r)\mathbf{H}(\mathbf{r},\mathbf{n}_r)\mathbf{W}^T({\boldsymbol{\hat\theta}}(t))\mathbf{z}(t) \\ 
     & - \frac{1}{2}(k\dot{\mathbf{z}}(t))^T\mathbf{W}({\boldsymbol{\hat\theta}}(t))\mathbf{H}^T(\mathbf{r},\mathbf{n}_r)\mathbf{H}(\mathbf{r},\mathbf{n}_r)\mathbf{W}^T({\boldsymbol{\hat\theta}}(t))(k\dot{\mathbf{z}}(t)).
\end{aligned}
\label{eq:h1}
\end{equation}
Note that $\mathbf{z}(t)$ is a smooth version of $\mathbf{e}_s(t)$.
We assume that for slow robot motion, and with a proper set of $r_s$, $c$, and $k$, 
$\left \| k\dot{\mathbf{z}}(t) \right \| = \left \| \mathbf{e}_s(t) - \mathbf{z}(t) \right \|$ is bounded. 
Then, there exists $\lambda_1 > 0$, such that:
\begin{equation}
    X_1(t) \ge \lambda_1 \mathbf{e}_s^T(t)\mathbf{W}({\boldsymbol{\hat\theta}}(t))\mathbf{H}^T(\mathbf{r},\mathbf{n}_r)\mathbf{H}(\mathbf{r},\mathbf{n}_r)\mathbf{W}^T({\boldsymbol{\hat\theta}}(t))\mathbf{e}_s(t).
    \label{eq:h1_eq}
\end{equation}
In addition,
we consider $\mathbf{d}(t) \in \mathbb{R}^{3k}$, the disturbances in the work space of the robot manipulation introduced by $\boldsymbol{\eta}(t)$.
Then, let $$X_2(t) = \mathbf{e}_s^T(t) \boldsymbol{\eta}(t)= \mathbf{e}_s^T(t)\mathbf{W}({\boldsymbol{\hat\theta}}(t))\mathbf{H}^T(\mathbf{r},\mathbf{n}_r)\mathbf{d}(t). $$
According to Young's inequality with $\varepsilon$, we have:
\begin{equation}
\begin{aligned}
    X_2(t) 
    \le &  \frac{\varepsilon}{2}  \mathbf{e}_s^T(t)\mathbf{W}({\boldsymbol{\hat\theta}}(t))\mathbf{H}^T(\mathbf{r},\mathbf{n}_r)\mathbf{H}(\mathbf{r},\mathbf{n}_r)\mathbf{W}^T({\boldsymbol{\hat\theta}}(t))\mathbf{e}_s(t) \\
    & + \frac{1}{2\varepsilon}\mathbf{d}^T(t)\mathbf{d}(t)
    \label{eq:h2}
\end{aligned}
\end{equation}
which is valid for $\forall \varepsilon > 0$.
Finally, based on the above derivations, we have:
\begin{equation}
\begin{aligned}
    \dot{V}(t) 
    = &
    - k_d\mathbf{e}_s^T(t)\mathbf{W}({\boldsymbol{\hat\theta}}(t))\mathbf{H}^T(\mathbf{r},\mathbf{n}_r)\mathbf{H}(\mathbf{r},\mathbf{n}_r)\mathbf{W}^T({\boldsymbol{\hat\theta}}(t))\mathbf{e}_s(t) \\
   & - X_1(t) + X_2(t) \\
    \le 
    & - (k_s\lambda_1 + k_d-\frac{\varepsilon}{2})  \mathbf{e}_s^T(t)\mathbf{W}({\boldsymbol{\hat\theta}}(t))\mathbf{H}^T(\mathbf{r},\mathbf{n}_r)  \\
    & \cdot \mathbf{H}(\mathbf{r},\mathbf{n}_r)\mathbf{W}^T({\boldsymbol{\hat\theta}}(t))\mathbf{e}_s(t) + \frac{1}{2\varepsilon}\mathbf{d}^T(t)\mathbf{d}(t)\\
    & =  -\alpha(\left \| \mathbf{H}(\mathbf{r},\mathbf{n}_r)\mathbf{W}^T({\boldsymbol{\hat\theta}}(t))\mathbf{e}_s(t) \right \|) +   \gamma(\left \| \mathbf{d}(t) \right \|).
\end{aligned}
\end{equation}
By properly selecting $k_s$ and $k_d$, the continuous, unbounded functions $\alpha( \cdot )$ and $\gamma( \cdot )$ can be strictly increasing.
They also satisfy: $\alpha(0)=0$, $\gamma(0)=0$. 
Thus, $\alpha( \cdot ) \in \mathcal{K}_\infty$, and $\gamma( \cdot ) \in \mathcal{K}_\infty$.
The closed-loop system is ISS (\cite{sontag2008input}) with respect to the disturbances $\mathbf{d}(t)$.
Therefore,
if $\lim_{t \to \infty} \mathbf{d}(t) = 0$, the deformation feature errors $\mathbf{e}_s(t)$ are convergent in the following way:
$$
\lim_{t \to \infty} \mathbf{H}(\mathbf{r},\mathbf{n}_r)\mathbf{W}^T({\boldsymbol{\hat\theta}}(t))\mathbf{e}_s(t) = \mathbf{0}
$$
which means that $\mathbf{e}_s(t)$ will be attracted to the null space of 
$\mathbf{H}(\mathbf{r},\mathbf{n}_r)\mathbf{W}^T({\boldsymbol{\hat\theta}}(t))$ as $t \rightarrow \infty$.
If $m = 3k$, $\mathbf{e}_s(t)$ will converge to zeros.
If $m > 3k$ (the case for most robotic tasks of DOM), the control problem is over-determined, thus the non-trivial null space will give rise to local minima configurations.
On the other hand, if $\mathbf{d}(t)$ is only bounded, the deformation feature errors $\mathbf{e}_s(t)$ will converge to small, bounded steady-state errors whose amplitude is influenced by the bounds of $\mathbf{d}(t)$ when $t \to \infty$.

\section{6. Simulation Analysis}
Simulations are conducted for comparative studies and analyzing different parameter settings.

\begin{figure}[!t]
    \centering
    \includegraphics[width=0.875\linewidth]{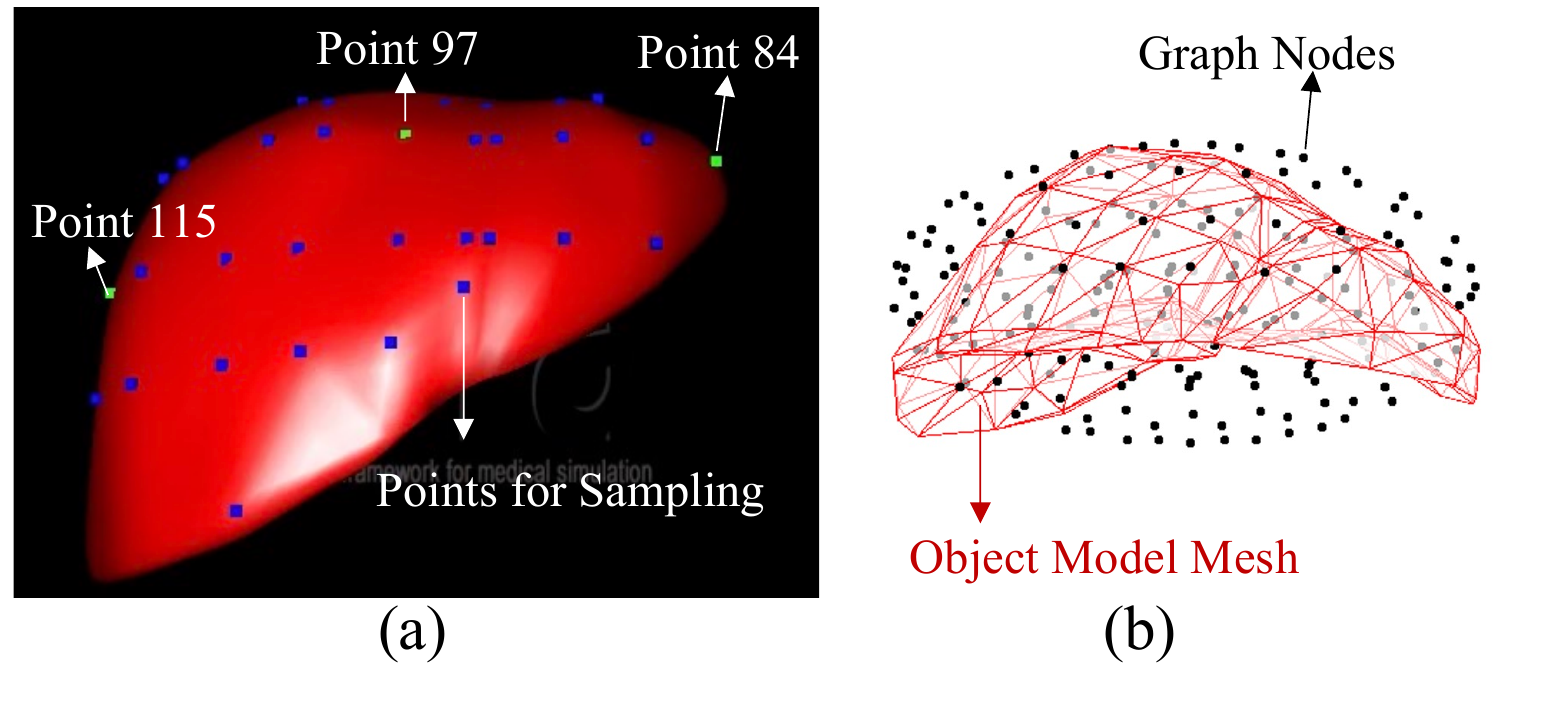}
    \caption{Simulation setup.
    (a). the manipulation points and the points for sampling. 
    (b). the comparison of the modal graph and the model mesh of the object.
    }
    \label{fig:simulation_setup}
\end{figure}
\begin{algorithm}[t]
	\caption{Controller($\mathbf{x}^*(\mathbf{p})$, 
	$G(\mathbf{n},\mathcal{E})$, $\widetilde{\mathbf{K}}$,
	$d_s$, $c$, $r_s$,
    $k_s$, $k$, $k_d$, $\Gamma$)}  
	\begin{algorithmic}[1]
	  \While {receiving vision and robot measurements} 
	  \State measure $\mathbf{x}(\mathbf{p}(t),t)$, $\mathbf{x}(\mathbf{r},t)$,
	  \If{ $t = t_0$ }
        \State compute $\boldsymbol{\gamma}({}^{\scriptscriptstyle p}\mathbf{c}^*(\mathbf{p}))$
        $\gets$ $\mathbf{x}^*(\mathbf{p})$
        \State compute 
        $\boldsymbol{\gamma}({}^{\scriptscriptstyle p}\mathbf{c}(\mathbf{r},t_0)$
        $\gets$  $\mathbf{x}(\mathbf{r},t_0)$
        \State find $\mathbf{n}_p^*$, $\mathbf{n}_r$ $\gets$ equation \eqref{eq:point_distribution}, $d_s$, $c$, $\mathbf{n}$
        \State compute $\mathbf{s}^*$ $\gets$ equation \eqref{eq:feature_computation}, $\mathbf{x}^*(\mathbf{p})$, $\widetilde{\mathbf{K}}$
	    \State initialize $\boldsymbol{\hat\theta}(t_0)$ with ones
	  \EndIf
	  
	  \If{ $t > t_0$ }
	    \State compute $\boldsymbol{\gamma}({}^{\scriptscriptstyle p}\mathbf{c}(\mathbf{p}(t),t))$
        $\gets$ $\mathbf{x}(\mathbf{p}(t),t)$
        \State find $\mathbf{n}_p(t)$
        $\gets$ equation \eqref{eq:point_distribution}, $d_s$, $c$, $\mathbf{n}$
	    \State compute $\mathbf{s}(t)$ $\gets$ equation \eqref{eq:feature_computation}, $\mathbf{x}(\mathbf{p}(t),t)$
	    
	    \While {robot is running}
	      \State compute $\mathbf{e}_s(t)$ $\gets$ $\mathbf{s}(t)-\mathbf{s}^*$
	      \State update $\boldsymbol{\hat\theta}(t)$ $\gets$ equation \eqref{eq:theta_dot}, $\Gamma$
	      \State compute $\mathbf{v}(t)$ $\gets$ equation $\eqref{eq:v_cmd}$, $k_s$, $k$, $k_d$
	      \State send $\mathbf{v}(t)$ to robot
	    \EndWhile
	  \EndIf
	  \EndWhile
	\end{algorithmic} 
\end{algorithm}

\subsection{6.1. Simulation Setup}
\begin{figure}[!t]
    \centering
    \includegraphics[width=0.875\linewidth]{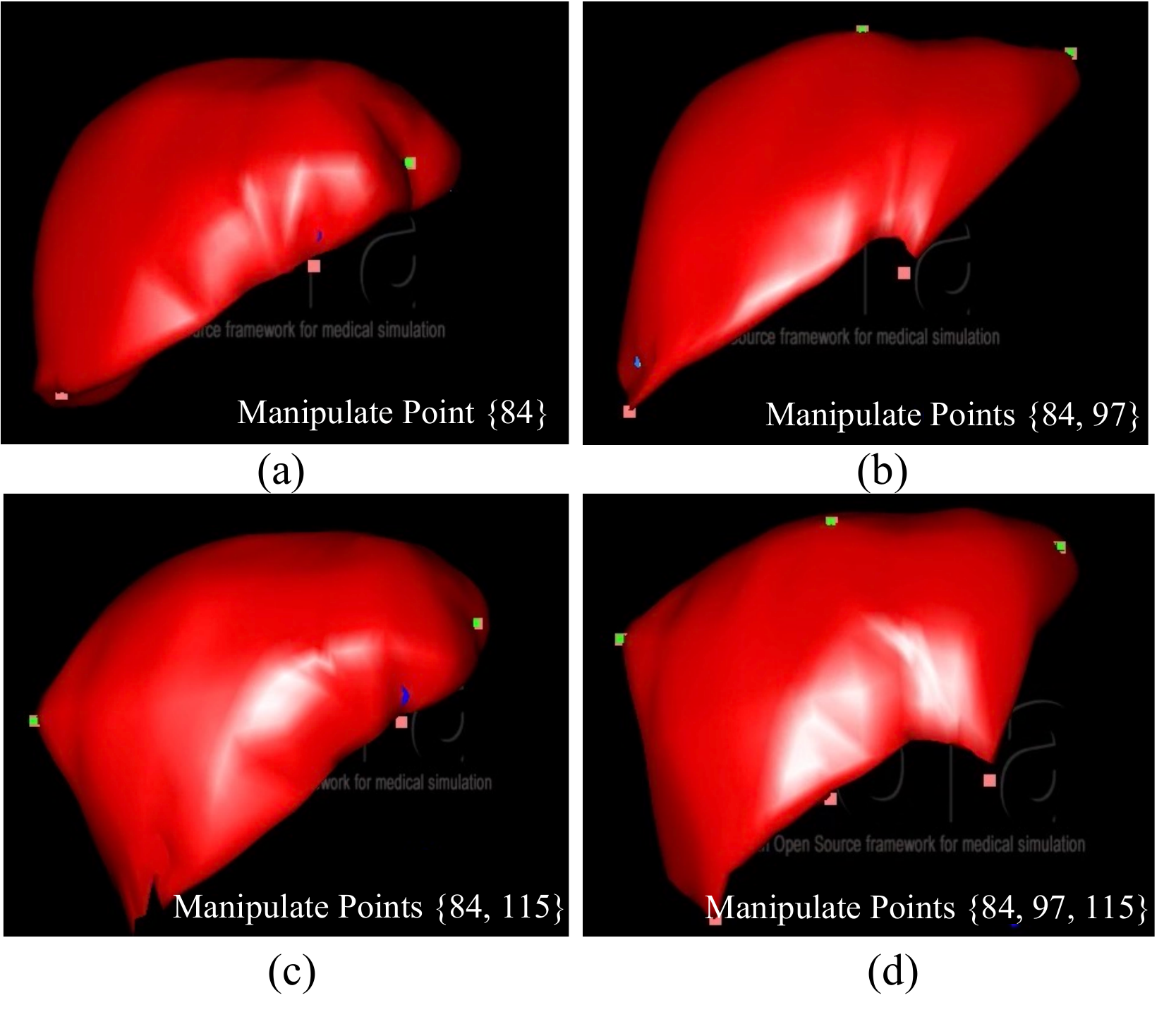}
    \caption{The desired deformation of different simulation cases.
    (a) with one manipulation point $\left \{ 84 \right \}$.
    (b) with two manipulation points $\left \{ 84, 97 \right \}$.
    (c) with two manipulation points $\left \{ 84, 115 \right \}$.
    (d) with three manipulation points $\left \{ 84, 97, 115 \right \}$.
    }
    \label{fig:simulation_cases}
\end{figure}

Our simulations are conducted on the Simulation Open Framework Architecture (SOFA) (\cite{allard2007sofa}) platform with a liver-shaped object (Figure~\ref{fig:simulation_setup}(a)).
To deform the object,
we set 1-3 manipulation points (selected from the three green points with the indices $\left \{ 84,97,115 \right \}$  in Figure~\ref{fig:simulation_setup}(a)) to validate our method.
The object is fixed in the simulation space by setting zero-displacement constraints on the pink points in Figure~\ref{fig:simulation_cases}.
To simulate noisy point measurements, at each loop, 
we obtain the measured points $\mathbf{p}(t)$ by randomly sampling $l$ points from the $30$ points (the blue points in Figure~\ref{fig:simulation_setup}(a)) on the front surface of the object (using the c++ functions "srand()" and "rand()").
Figure~\ref{fig:simulation_setup}(b) compares the object model used in the simulator and the modal graph we used for control.
The graph consists of $210$ nodes, whose size and pose are determined using the moment-based method used by \cite{pentland1991closed}.
Following the unit system in SOFA, the graph is set with the following parameters:
$\alpha_1 = \alpha_2 = 1$, $E=50$, $v=0.45$, $M=1$.
Physical properties of the object model are set to be: $E=5000,v=0.47,M=100$.
The control parameters are set to be: $k_s = 150$, $k=0.6$, $k_d=7.5$, $\Gamma=500$, $c=2$.
The frequency of the simulation is $50$ Hz.
Algorithm 1 shows the implementation of our shape-servoing controller.

During the simulations, we first computed the domain of influence $S(\mathbf{p}(t_0),t_0)$ for the measured points $\mathbf{p}(t_0)$ by setting $r_s = 2$ at $t_0$.
Then, after the object was deformed, we could change the settings of $r_s$ but only online searched the supporting nodes $\mathbf{n}_p(t)$ among the nodes in $S(\mathbf{p}(t_0),t_0)$.
For result analysis, we set the following evaluation metrics:
first, as our controller is formulated with the feedback feature errors,
we recorded the norm of the feature errors' dynamic extension $\left \| \mathbf{z}(t) \right \|
 = \sqrt{\mathbf{z}^T(t) \mathbf{z}(t)}$ to show the performance 
 of our feedback controller;
Second, to show the convergence of the global 3D shape, we recorded the following total mesh error sum:
\begin{equation}
    e_{x}(t) = (\mathbf{x}(\mathbf{p}_o,t) - \mathbf{x}^*(\mathbf{p}_o))^{T}(\mathbf{x}(\mathbf{p}_o,t) - \mathbf{x}^*(\mathbf{p}_o))
\end{equation}
where $\mathbf{x}(\mathbf{p}_o,t)$ is the position vector of all the points $\mathbf{p}_o$ on the object model mesh,
and $\mathbf{x}^*(\mathbf{p}_o)$ is their desired position vector.

\begin{figure}[t]
    \centering
    \includegraphics[width=\linewidth]{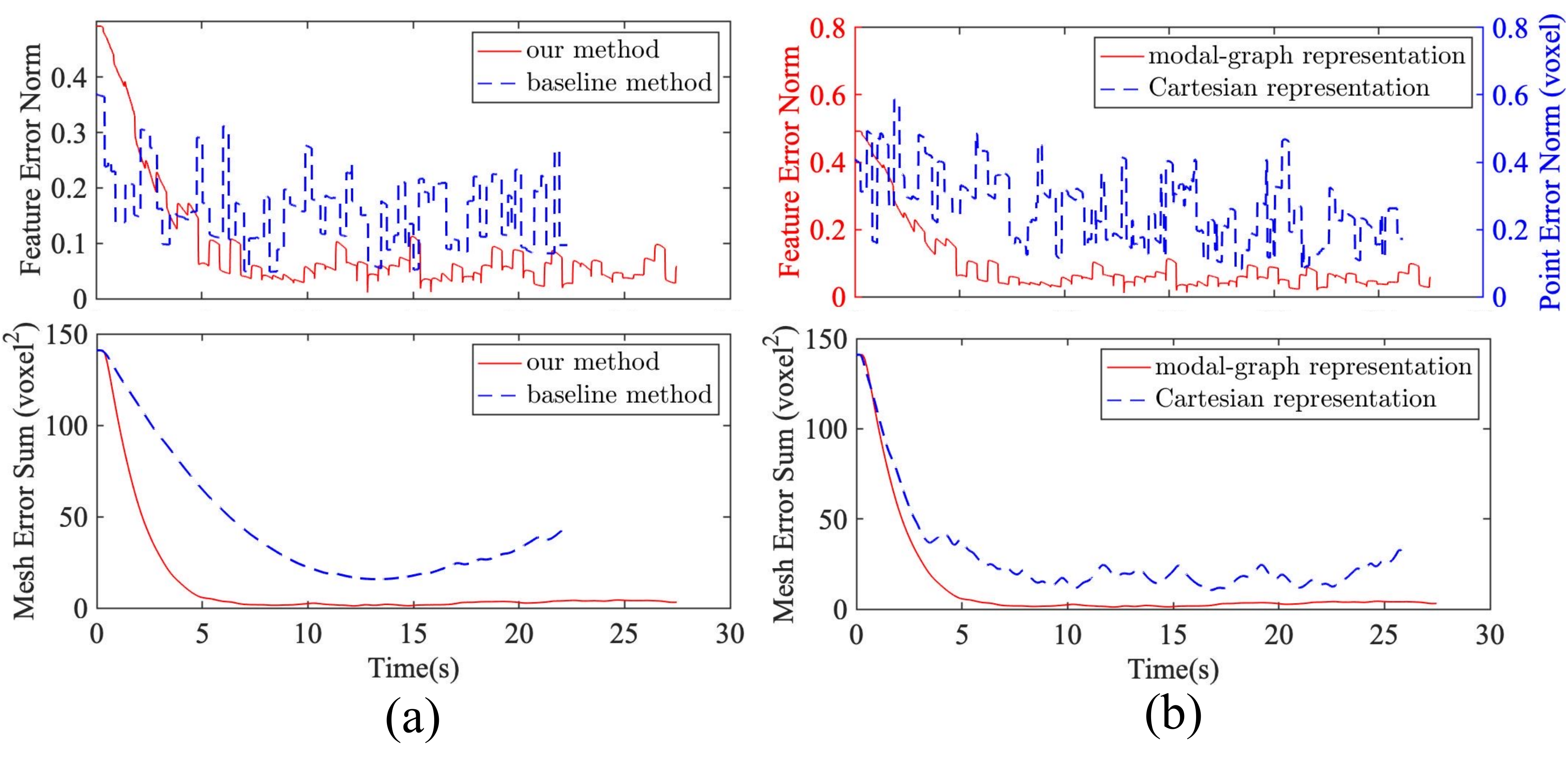}
    \caption{Simulation results of: (a) the cases using our method and the baseline method;
    (b) the cases using the modal representation and Cartesian representation.
    The mesh error curve of the case using Cartesian representation is not oscillating in strict synchronization with the point error curve because the measured points were randomly sampled.
    }
    \label{fig:1mp_base_modal}
\end{figure}

\subsection{6.2. Comparative Studies}
We conducted several comparative studies
to compare the proposed method 
with the baseline method (our previous modal-based deformation controller (\cite{10122176}) and to show the effects of different parts in the proposed controller.
In the comparative studies,
we set one manipulation point $\left \{ 84 \right \}$, and manually generated the desired deformation shown in Figure~\ref{fig:simulation_cases}(a).
We randomly sampled $l=20$ points and set the dimension of our deformation features to be: $m=20$.

\subsubsection{6.2.1. Comparison with the baseline method:}
The baseline method deals with points with known spatial orders
and chronological correspondences.
Due to the known point correspondence, the method can compute modal-based deformation features with an offline formulated linear deformation basis.
However, to deal with raw point clouds with unknown correspondence, we revised the method by online re-computation of its deformation basis given the real-time positions of the measured points.
We compared the performance of the (revised) baseline method with the proposed modal-graph-based method.

The result curves in Figure~\ref{fig:1mp_base_modal}(a) show that:
1) the proposed method achieves the minimization of both $\left \| \mathbf{z}(t) \right \|$ and $e_{x}(t)$ while the baseline method fails;
2) the feature errors of the baseline method change more drastically than the proposed method.
This is because in the baseline method:
online re-computation of the deformation basis requires projecting the randomly measured points to a mesh element, which brings enormous computational burdens and affects real-time performance;
the element-dependent computation of deformation features causes extra discontinuities as the projected element changes, which not only leads to significant changes in the feature value but also affects the control stability.

\subsubsection{6.2.2. Effect of the modal-graph representation:} 
Our method uses the modal-graph representation (i.e., the modal-based deformation features $\mathbf{s}(t)$) of the randomly measured points.
For comparison, we set a case that directly uses the Cartesian representation (i.e., $\mathbf{x}(\mathbf{p}(t),t)$) of the randomly measured points.
In that case, we generated the following control laws based on the geometric model (the red mesh in Figure~\ref{fig:simulation_setup}(b) of the object:
\begin{equation}
    \mathbf{v}(t) = \dot{\mathbf{x}}(\mathbf{r},t) = -k_b \mathbf{J}_b^+(\mathbf{x}(\mathbf{p}(t),t))(\mathbf{x}(\mathbf{p}(t),t) - \mathbf{x}^*(\mathbf{p}^*))
\end{equation}
where $k_b$ is a positive gain, $\mathbf{J}_b^+(\mathbf{x}(\mathbf{p}(t),t)) \in \mathbb{R}^{3 \times 60}$ is the pseudo-inverse of the deformation Jacobian matrix online computed using the diminishing-rigidity approximation (\cite{berenson2013manipulation}).

To compare the results, 
we also recorded the point error norm (i.e., $\left \| \mathbf{x}(\mathbf{p}(t),t) - \mathbf{x}^*(\mathbf{p}^*) \right \|$) to visualize the Cartesian representation.
The result curves in Figure~\ref{fig:1mp_base_modal}(b) imply that:
1) changes in the measured points introduce much larger disturbances to the Cartesian representation than the modal-graph representation;
2) although the Cartesian representation can reflect the deformation trend toward the target shape to some extent,
the method using this representation can only deform the object when the current shape is relatively far from the target.
As the deformation target becomes closer, such a trend produces a smaller influence than the disturbances of measurement changes,
the method cannot control the object to achieve the desired shape.
As shown in the mesh error curve of the case with Cartesian representation, after $e_x(t)$ decreases to a certain small value, it begins to oscillate around and fails to be minimized;
3) the controller using modal-graph representation achieves the minimization of both $\left \| \mathbf{z}(t) \right \|$ and $e_{x}(t)$.
To be brief, by representing 
$\mathbf{x}(\mathbf{p}(t),t) -\mathbf{x}^*(\mathbf{p}^*)$ 
through the modal graph,
we can extract the deformation features that are robust to measurement changes.
Moreover, using the modal-graph representation, we can formulate control laws without knowing both the physical and geometric models of the object.

\begin{figure}[t]
    \centering
    \includegraphics[width=\linewidth]{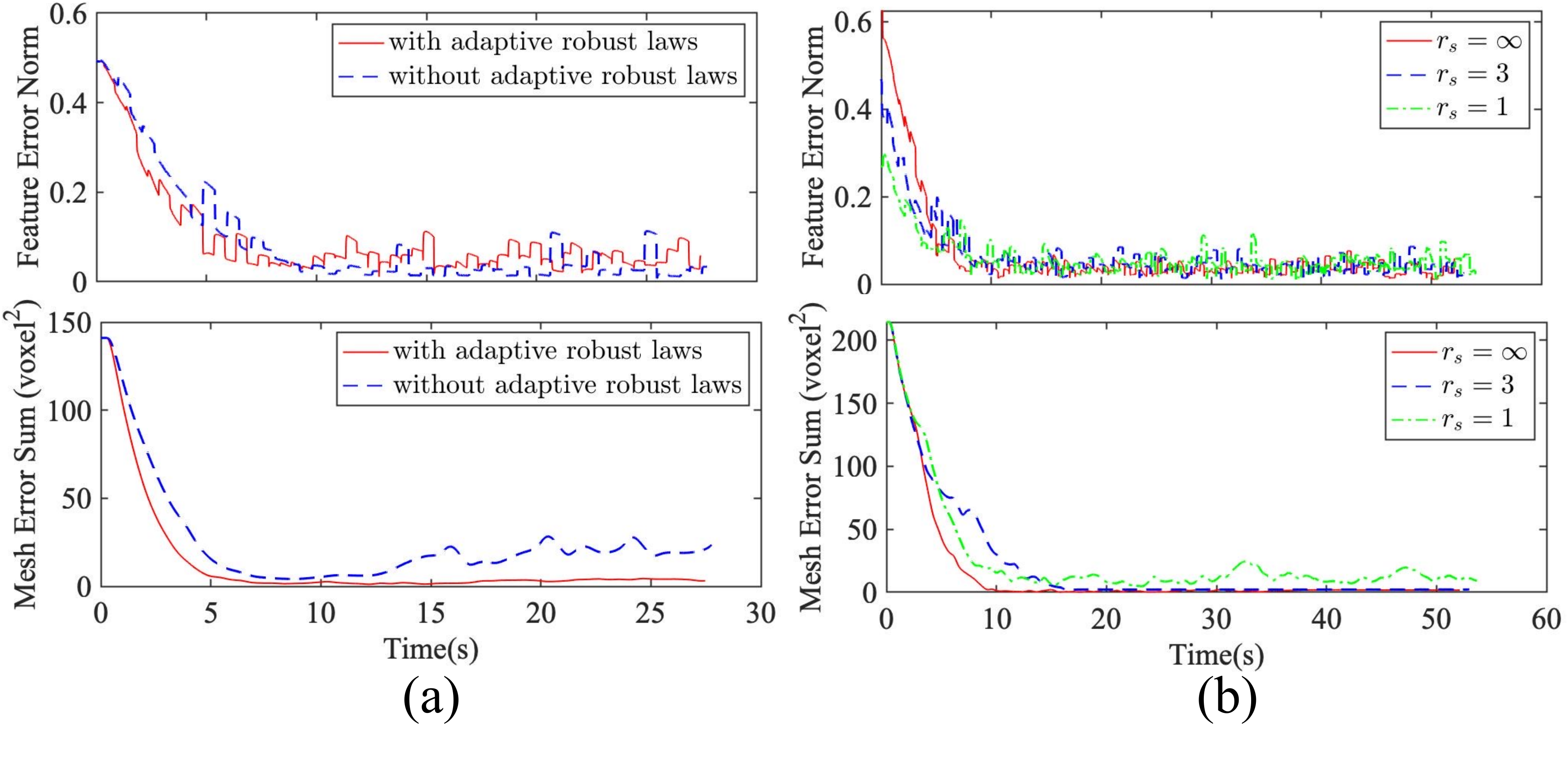}
    \caption{Simulation results of: (a) the cases with and without the adaptive robust laws;
    (b) the cases with different $r_s$.
    }
    \label{fig:1mp_control_2mp}
\end{figure}

\subsubsection{6.2.3. Effect of the adaptive robust control laws:}
Our method designs a new adaptive robust controller with dynamic state feedback to tackle unmodeled effects and noises.
To show the effect of the designed adaptive robust laws, we set a case where 
we remove the robust term and no longer use the dynamic extension of $\mathbf{e}_s(t)$.
Then, we compared the performances with and without the adaptive robust laws.
The result curves in Figure~\ref{fig:1mp_control_2mp}(a) show that the case without the adaptive robust laws cannot minimize the mesh errors $e_{x}(t)$ to a small and bounded region.
This is because we cannot ensure the controller is ISS without the adaptive robust laws.

\subsection{6.3. Parameter Analysis}
\subsubsection{6.3.1. With different settings of $r_s$:}
We conducted simulations to validate our method with different $r_s$.
In these cases, we set two manipulation points $\left \{ 84,97 \right \}$, and manually generated the desired deformation shown in Figure~\ref{fig:simulation_cases}(b).
We set the deformation feature dimension to be: $m=30$, and randomly sampled $20$ points.
We compared three conditions:
$r_s = 1$, $r_s = 3$, $r_s = +\infty$.
Simulation results in Figure~\ref{fig:1mp_control_2mp}(b) show that both $\left \| \mathbf{z}(t) \right \|$ and $e_{x}(t)$ are minimized to a small and bounded region.
The curves also imply that:
using larger $r_s$, changes in the sampled nodes introduce fewer changes in the deformation features, which results in smaller oscillations and steady-state errors.

\begin{figure}[t]
    \centering
    \includegraphics[width=\linewidth]{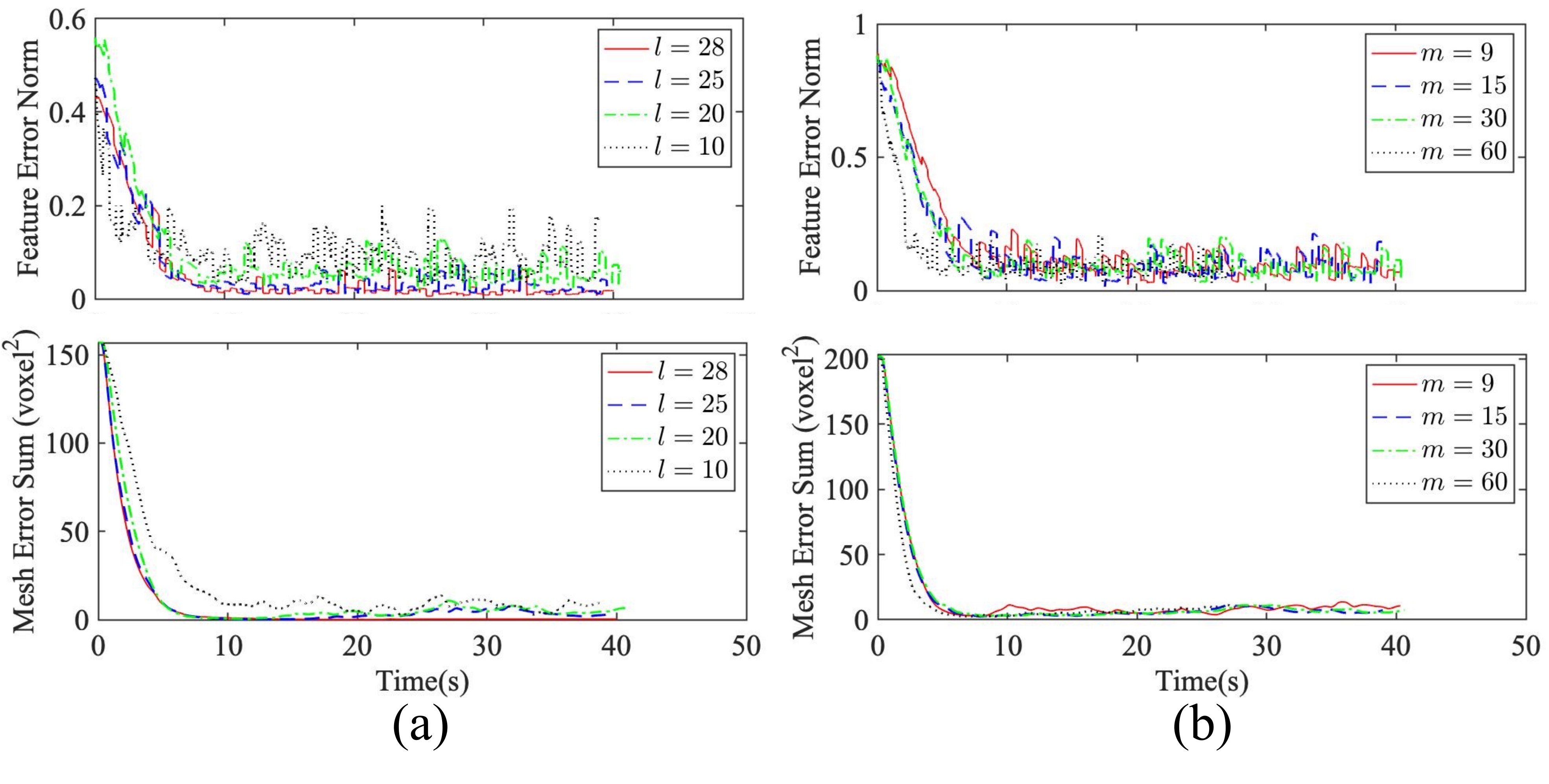}
    \caption{Simulation results of: (a) the cases with different $l$;
    (b) the cases with different $m$.
    }
    \label{fig:2mp_n_3mp}
\end{figure}

\subsubsection{6.3.2. With different settings of $l$:}
We conducted simulations with different numbers of the randomly-measured points.
In these cases, we set another pair of manipulation points $\left \{ 84,115 \right \}$, and manually generated the desired deformation shown in Figure~\ref{fig:simulation_cases}(c).
We set the deformation feature dimension to be: $m=30$, and compared four different conditions:
$l=28$, $l=25$, $l=20$, and $l=10$.
Simulation results in Figure~\ref{fig:2mp_n_3mp}(a) show that both $\left \| \mathbf{z}(t) \right \|$ and $e_{x}(t)$ are minimized to a small and bounded region.
As we sampled more points,
we obtained more global measurements of the object.
Therefore,
the deformation features change less, which results in smaller oscillations and steady-state errors.

\subsubsection{6.3.3. With different settings of $m$:}
We also conducted simulations to validate our method with different deformation feature dimensions.
In these cases, we set three manipulation points $\left \{ 84,97,115 \right \}$, and manually generated the desired deformation shown in Figure~\ref{fig:simulation_cases}(d).
We randomly sampled $20$ points, and compared four different conditions:
$m=9$, $m=15$, $m=30$, and $m=60$.
Simulation results in Figure~\ref{fig:2mp_n_3mp}(b) show that both $\left \| \mathbf{z}(t) \right \|$ and $e_{x}(t)$ are minimized to a small and bounded region.
The starting positions of the feature error curves are the same because we sampled the same points at $t_0$.
The curves also show that:
for the case $m=3k$, the deformation converges faster and to smaller steady-state errors than the other over-determined control cases.

\subsection{6.4. Discussions on Simulations}
Simulation results validate the effectiveness of our method with different sampling conditions, deformation feature dimensions, and supporting sizes.
Based on the results,
we also discuss the parameter settings for better control performance:
1) the optimal supporting size parameter during manipulation (when $t > t_0$) is $r_s = +\infty$;
2) the greater the number of the measured points (than the feature dimension $m$),
the smaller oscillations and steady-state errors are.

\section{7. Experiments}
We further validate and analyze our controller with a real robotic system.
Various validations are conducted with different kinds of objects and under different settings.
To show the simplicity and generality of our method,
we use the same modal graph (in a simple ellipsoidal shape and with fixed physical properties) to deal with objects of different shapes, sizes, and materials.

\begin{figure}[t]
    \centering
    \includegraphics[width=\linewidth]{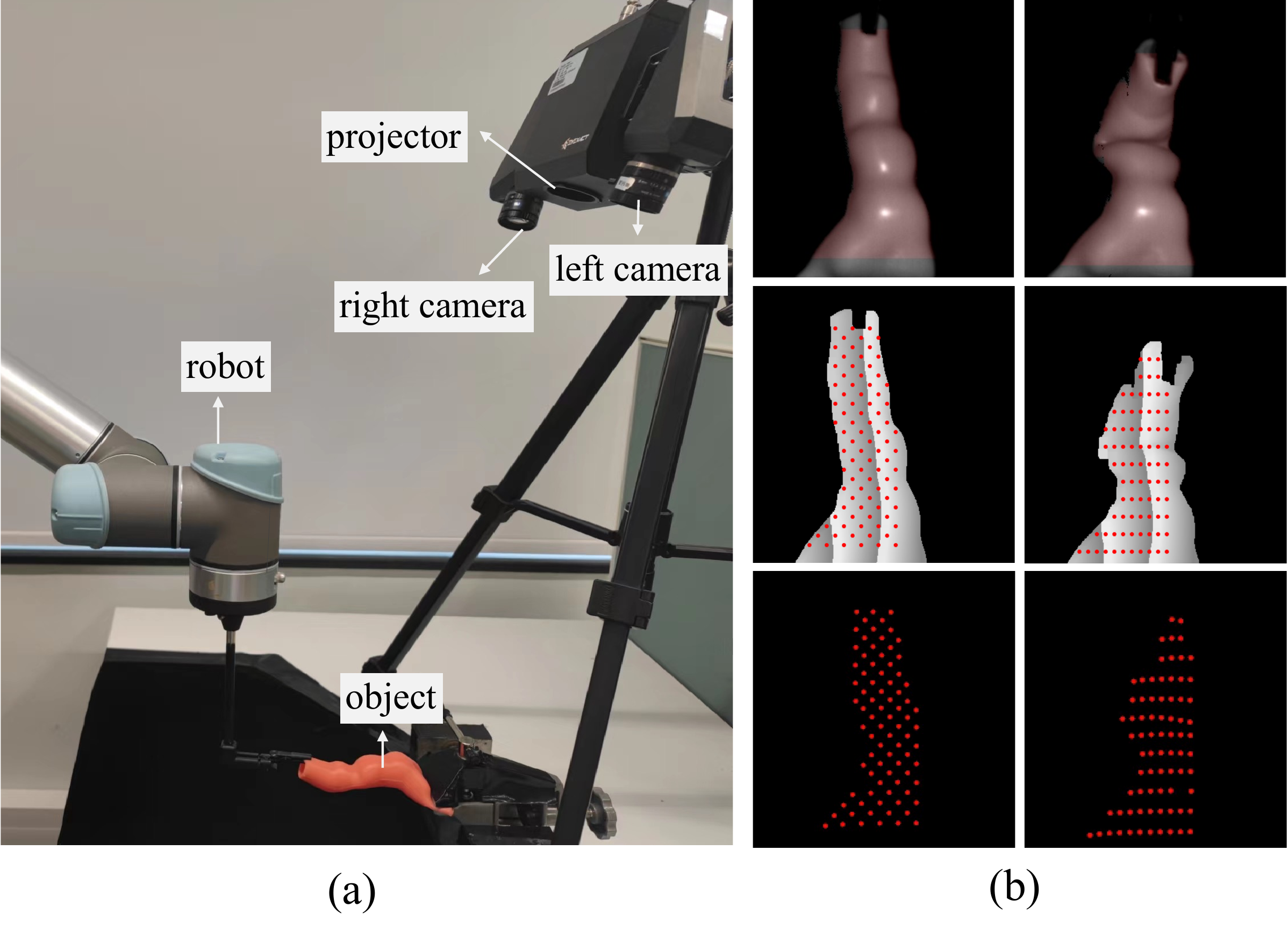}
    \caption{Experiment setup.
    (a) the robot-camera system.
    (b) illustrations of the point measuring:
    the first row shows the left view images, where the translucent red masks indicate the measured area;
    the second row shows the codeword images, where the red dots indicate the selected pixels for reconstruction;
    the third row shows the reconstructed points.}
    \label{fig:experiment_setup}
\end{figure}

\begin{figure}[t]
    \centering
    \includegraphics[width=0.9\linewidth]{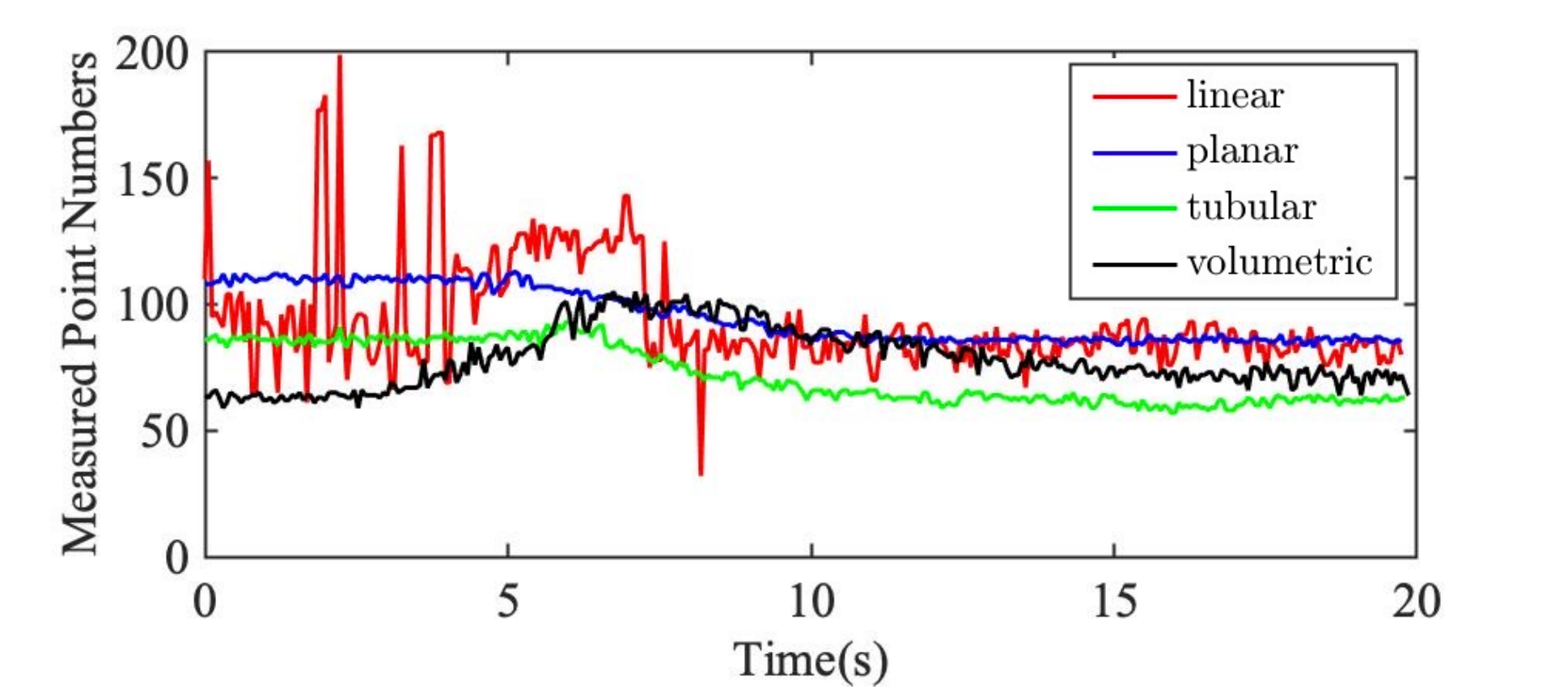}
    \caption{The numbers of the measured points during the robot manipulation within the first $20$ seconds for the validation cases with different objects.}
    \label{fig:fp_n}
\end{figure}

\subsection{7.1. Experiment Setup}
The experiment setup is given in Figure~\ref{fig:experiment_setup}(a).
In this setup, one end of the object is fixed in space (using a vise) while another end is rigidly and firmly grasped by the robot.
Deformation of the object is monitored by a structure-lighted vision system consisting of two calibrated cameras and one uncalibrated projector.
Hand-eye calibrations of this robot-camera system are performed using the method proposed by \cite{shiu1989calibration}.
The vision system measures point clouds using the method (\cite{sui20193d}) based on active pattern projections.
For real-time performance, 
only $100$ to $300$ points are measured.
Figure~\ref{fig:experiment_setup}(b) shows how we obtain the point measurements.
First, on the codeword image (the second row of Figure~\ref{fig:experiment_setup}(b)) of the left camera, we select some pixels within the segmented object area by pixel length;
Then, 3D reconstructions are only performed on the selected pixels (the third row of Figure~\ref{fig:experiment_setup}(b)).
The missing points between the second row and the third row of Figure~\ref{fig:experiment_setup}(b) are caused by the reconstruction noises of the 3D scanning system.
During manipulation, the obtained points change with the object deformation, lighting conditions,
occlusions, and measurement noises.
For example, Figure~\ref{fig:fp_n} illustrates how the numbers of the measured points change in the experiments with linear, planar, tubular, and volumetric objects (for Section 7.2.1-7.2.4).

The modal graph used in experiments consists of $418$ nodes, and is set with the following parameters:
$\left \{ a_x, a_y,a_z\right \} = \left \{ 0.03, 0.045,0.0025\right \}$ (unit: m), $\alpha_1 = \alpha_2 = 1$, $E=1 GPa$, $v=0.4$, $M=0.05 kg$.
Under our experiment setup,
the pose of the modal graph with respect to the camera frame is set to be:
\begin{equation}
    {}^c\mathbf{T}_m = 
    \begin{bmatrix}
    {}^{c}\mathbf{R}_e(t_0)  &  {}^c\overline{\mathbf{x}}(\mathbf{p}(t_0),t_0)-a_z \begin{bmatrix}
  0 & 0 & 1
\end{bmatrix}^T\\
\mathbf{0}_{3\times 1} & 1
\end{bmatrix}
\label{eq:base_mesh_R}
\end{equation}
where ${}^{c}\mathbf{R}_e(t_0)$ is the rotation matrix of the robot end-effector measured at $t_0$, ${}^c\overline{\mathbf{x}}(\mathbf{p}(t_0),t_0)$ is the position vector of the mass center of the measured points $\mathbf{p}(t_0)$.
Then, with respect to the modal graph frame (i.e., our default frame), the point cloud frame is defined with:
\begin{equation}
\mathbf{T}_p(t) = \begin{bmatrix}
 \mathbf{I}_{3\times 3} & \overline{\mathbf{x}}(\mathbf{p}(t),t) - a_z \begin{bmatrix}
  0 & 0 & 1
\end{bmatrix}^T   \\
  \mathbf{0}_{3\times 1} & 1
\end{bmatrix}
\end{equation}
where $\overline{\mathbf{x}}(\mathbf{p}(t),t) \in \mathbb{R}^3$ is the position vector of the mass center of $\mathbf{p}(t)$.
Under this frame, the local Cartesian coordinates of the measured points are denoted by:
\begin{equation}
{}^{\scriptscriptstyle p}\mathbf{x}(\mathbf{p}(t),t) = \left \{ {}^{\scriptscriptstyle p} x(\mathbf{p}(t),t), {}^{\scriptscriptstyle p} y(\mathbf{p}(t),t), {}^{\scriptscriptstyle p} z(\mathbf{p}(t),t) \right \}^T,
\end{equation}
using which their constrained local parametric coordinates:
\begin{equation}
{}^{\scriptscriptstyle p}\mathbf{c}(\mathbf{p}(t),t) = \left \{ {}^{\scriptscriptstyle p}\zeta(\mathbf{p}(t),t), {}^{\scriptscriptstyle p}\sigma(\mathbf{p}(t),t) \right \}^T
\end{equation}
can be computed with:
\begin{equation}
    {}^{\scriptscriptstyle p}\sigma(\mathbf{p}(t),t) = \tan^{\scriptscriptstyle{-1} } \frac{{}^{\scriptscriptstyle p} y(\mathbf{p}(t),t)}{{}^{\scriptscriptstyle p} x(\mathbf{p}(t),t)}  
\end{equation}
\begin{equation}
    {}^{\scriptscriptstyle p}\zeta(\mathbf{p}(t),t) \!= \!\frac{ \sgn( \boldsymbol \epsilon _{c}^z \! \cdot \! \boldsymbol \epsilon_{\scriptscriptstyle p}^z(t))\tan^{\scriptscriptstyle{-1} }(^{\scriptscriptstyle p} z(\mathbf{p}(t),t)) \cos({}^{\scriptscriptstyle p}\sigma(\mathbf{p}(t),t)) }
    { \max \left \{ ^{\scriptscriptstyle p} x(\mathbf{p}(t),t), {}^{\scriptscriptstyle p} y(\mathbf{p}(t),t)\right \}  }  
\end{equation}
where the sign function $\sgn( \boldsymbol \epsilon _{c}^z\!\cdot\!\boldsymbol \epsilon_{\scriptscriptstyle p}^z (t))$ is added to constrain the points on the upper surface of the graph towards the camera.

In our experiments, we set the deformation feature dimension to be: $m=20$.
The other parameters in our controller are set to be:
$k_s = 4$, $k=0.65$, $k_d=0.1$, $\Gamma=800$, $c=2$.
The frequencies of the 3D vision program and the control program are 15 Hz.
The desired deformation is generated by manually moving the robot. 
For each experiment, we first computed the domain of influence $S(\mathbf{p}(t_0),t_0)$ for the measured points $\mathbf{p}(t_0)$ by setting $r_s = 2$ at $t_0$.
Then, after the controller running, we set $r_s = +\infty$ but only online searched the supporting nodes $\mathbf{n}_p(t)$ among the nodes in $S(\mathbf{p}(t_0),t_0)$.
The controller computed $\mathbf{v}(t)$, the velocity commands for the manipulation point, which were further transformed to the joint velocities of the robot.

For result analysis, we set the following evaluation metrics:
First, as our controller is formulated and analyzed with the feedback feature errors,
we recorded the norm of the feature errors $\left \| \mathbf{e}_{s}(t) \right \|$ to show the performance and validate the stability of our feedback controller;
Second, to show the manipulation process toward the target configuration,
we recorded $\mathbf{e}_d(t) = \mathbf{x}(\mathbf{r},t)-\mathbf{x}^*(\mathbf{r})$, the manipulation errors between the measured positions $\mathbf{x}(\mathbf{r},t)$ and the target positions $\mathbf{x}^*(\mathbf{r})$ of the manipulation point;
In addition, with reference to the formula in \cite{liu2020morphing} and \cite{wu2021balanced},
we computed the following Chamfer Distance (CD):
\begin{equation}
d_{CD} = \frac{1}{\left | P_r \right  | } 
\sum_{x \in P_r} \min_{y \in P_d} \left \| x-y \right \| 
+  \frac{1}{\left | P_d \right  | } \sum_{y \in P_d} \min_{x \in P_r} \left \| y-x \right \|
\end{equation}
to measure the mean distance
between each point in the resulting points $P_r$ (with $\left | P_r \right  |$ points) to its nearest neighbor
in the desired points $P_d$ (with $\left | P_d \right  |$ points).

\begin{figure}[t]
	\centering
	\begin{minipage}{\linewidth}
 \centering
    \includegraphics[width=0.825\linewidth]{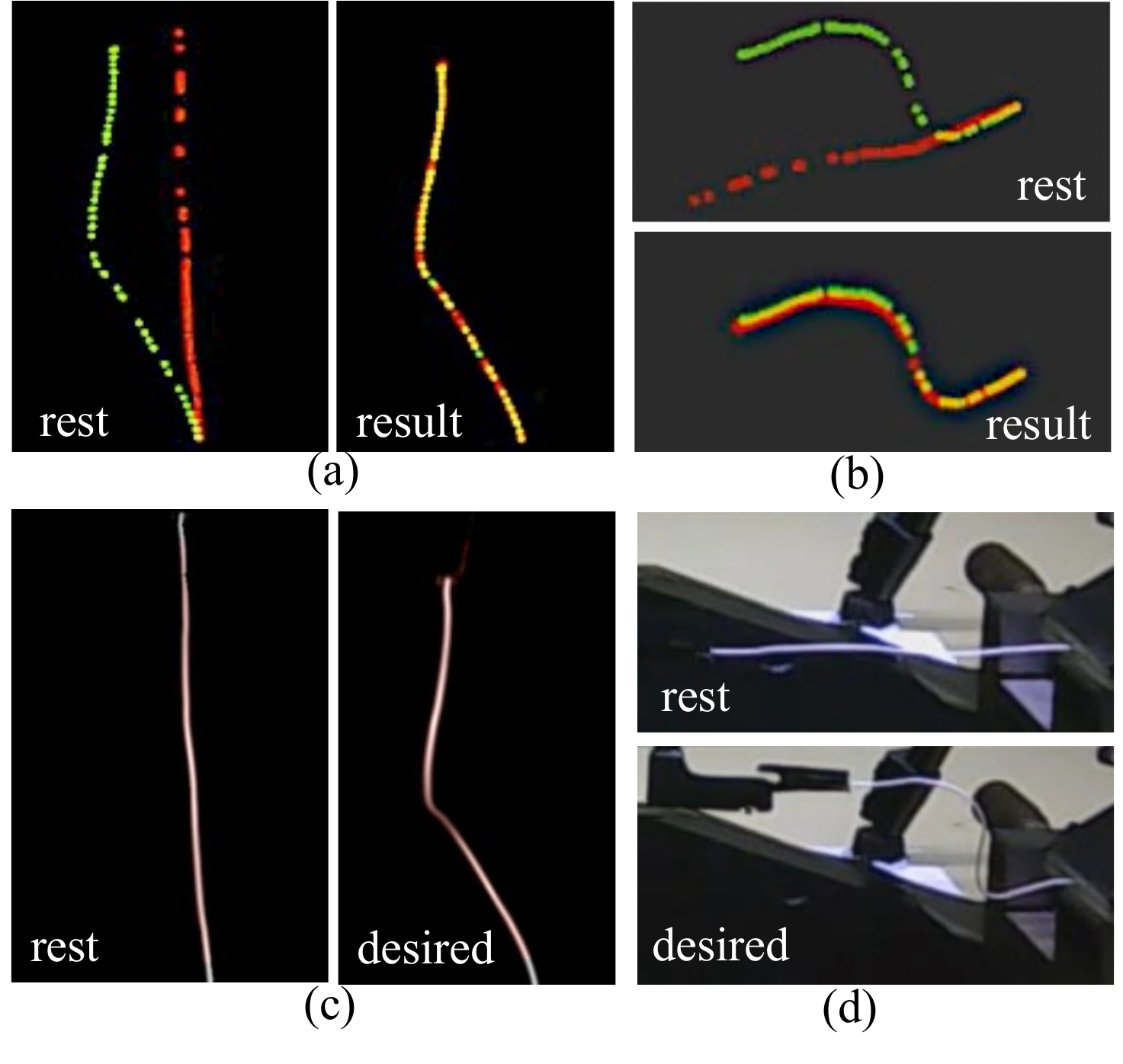}
    \vspace{-0.3cm}
    \caption{The linear object experiment.
    (a/b) the front/side views of the points, where the red/green points are for the measured/desired configuration. (c/d) the left/third camera views.
    }
    \label{fig:linear_setup}
	\end{minipage}
	\\
        \vspace{0.3cm}
	\begin{minipage}{\linewidth}
 \centering
    \includegraphics[width=0.85\linewidth]{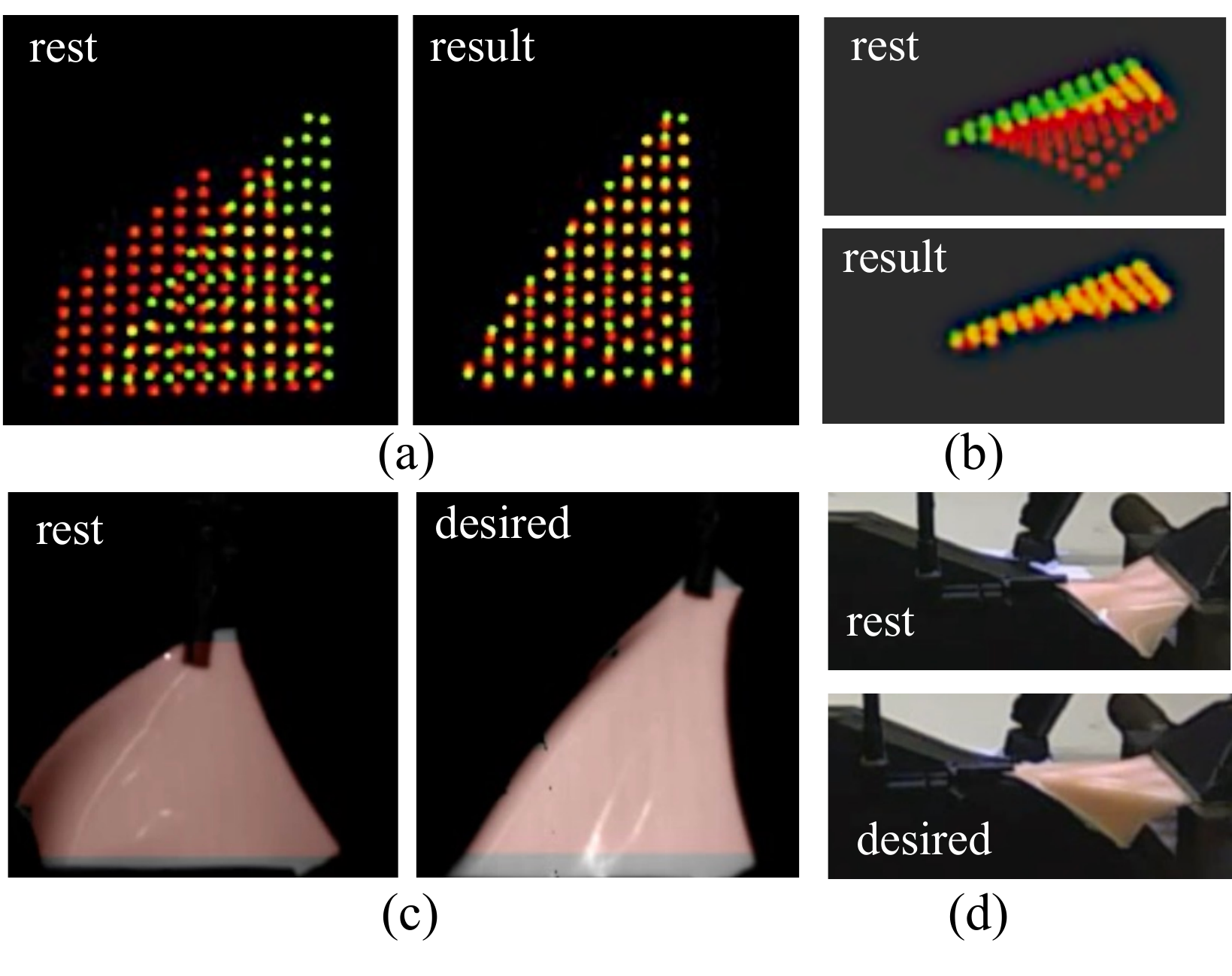}
    \vspace{-0.3cm}
    \caption{The planar object experiment.
(a/b) the front/side views of the points, where the red/green points are for the measured/desired configuration. (c/d) the left/third camera views.
    }
    \label{fig:planar_setup}
	\end{minipage}
        \\
        \vspace{0.3cm}
	\begin{minipage}{\linewidth}
 \centering
    \includegraphics[width=\linewidth]{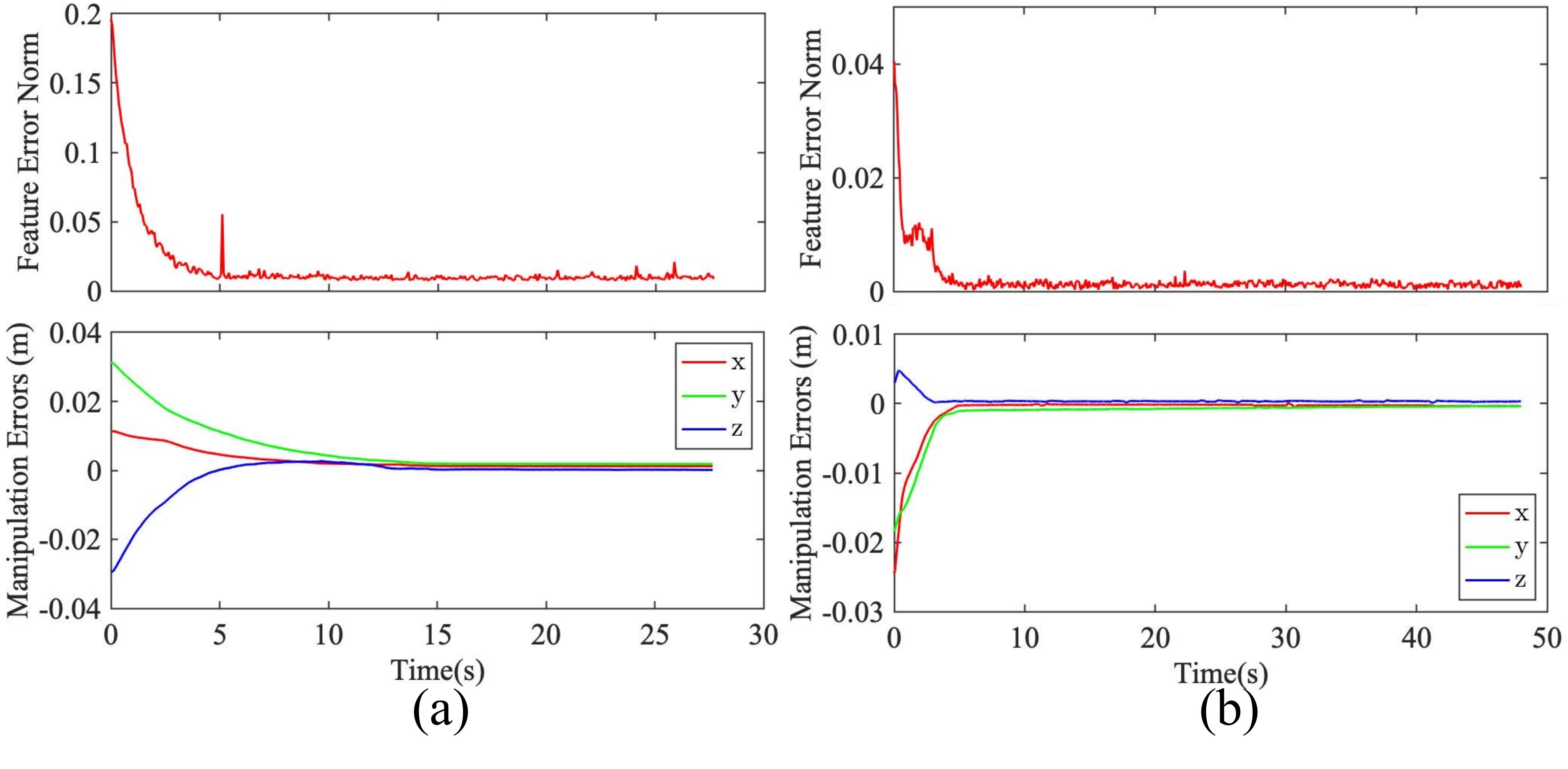}
    \vspace{-0.5cm}
    \caption{The result curves of the deformation feature errors and the manipulation errors of: (a) the linear object experiment, where the bulges in the feature error curves are caused by sensing noises;
    (b) the planar object experiment, where
    the turning points in the feature error curves are caused by the disturbances of the surface reflections during the manipulation. \\
    }
    \label{fig:linear_planar}
	\end{minipage}
	\vspace{-0.5cm}
\end{figure}

\subsection{7.2. Basic Validation Cases}
Experiments with different kinds of objects (i.e., linear, planar, tubular, and volumetric objects) are conducted to validate our controller with different desired deformation.
Then, we discuss the influence of different graph shapes and different selections of modes.

\subsubsection{7.2.1. Lifting a linear object forward:}
We selected the deformable linear object to be a piece of thin wire.
As shown in Figure~\ref{fig:linear_setup}(c,d), we set the desired deformation to be: lifting the wire forward.
The comparisons between the measured and the desired points of the object are given in Figure~\ref{fig:linear_setup}(a,b), from which we can see that the object reached its desired deformation in general.
The CD between the resulting and desired points is: $d_{CD} = 3.51(mm)$.
In addition, the result curves in Figure~\ref{fig:linear_planar}(a) show the minimization of both the deformation feature errors and the manipulation errors.

\subsubsection{7.2.2. Pulling a planar object to the right:}
We selected the deformable planar object to be a piece of silicon skin to validate our controller.
As shown in Figure~\ref{fig:planar_setup}(c,d), we set the desired deformation to be: pulling the silicon skin to the right.
The comparisons between the measured and the desired points of the object are given in  Figure~\ref{fig:planar_setup}(a,b), from which we can see that the object reached its desired deformation in general.
The CD between the resulting and desired points is: $d_{CD} = 1.97(mm)$.
In addition, the result curves in Figure~\ref{fig:linear_planar}(b) show the minimization of both the deformation feature errors and the manipulation errors.

\begin{figure}[t]
	\centering
	\begin{minipage}{\linewidth}
        \centering
    \includegraphics[width=0.85\linewidth]{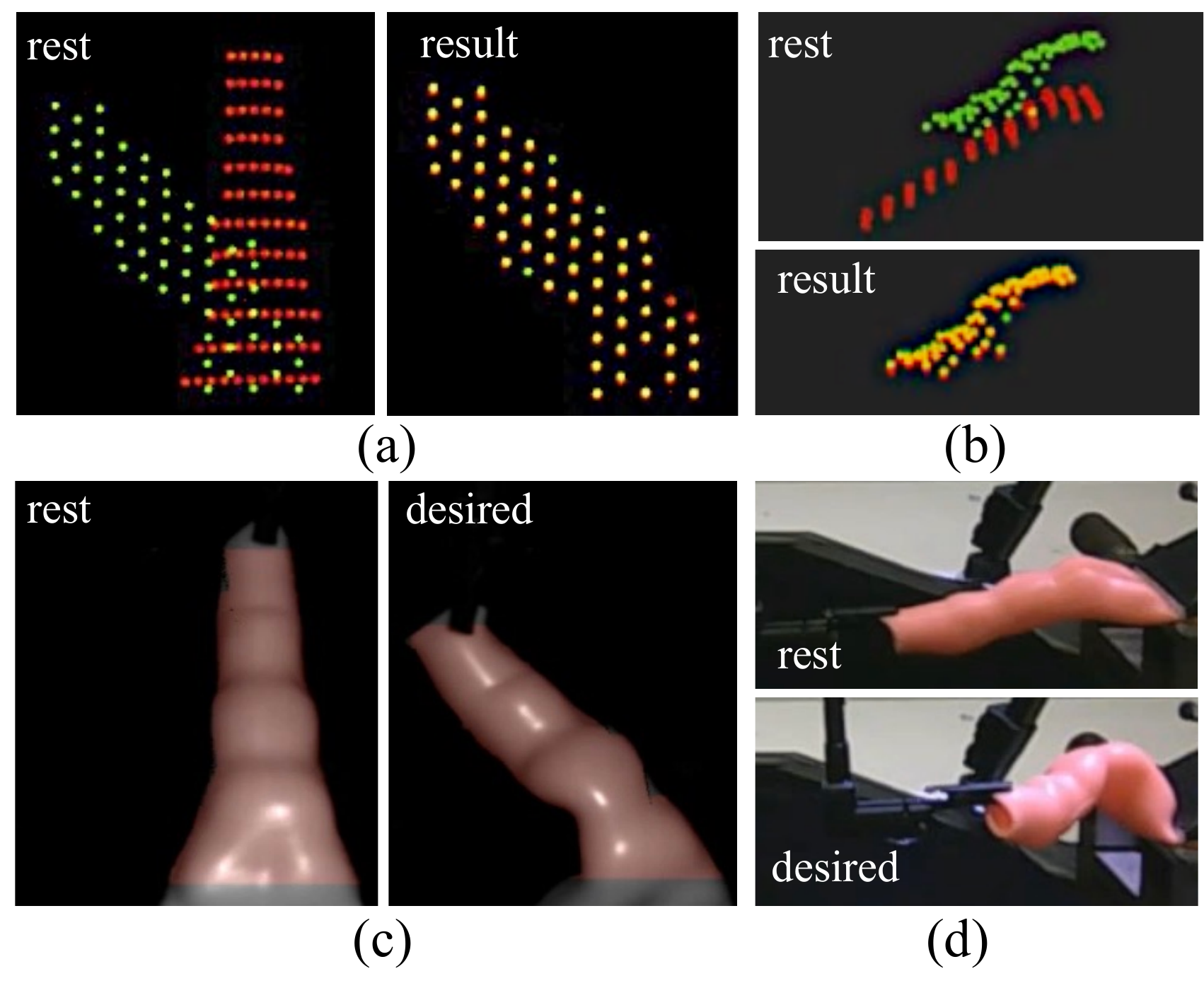}
    \vspace{-0.3cm}
    \caption{The tubular object experiment.
    (a/b) the front/side views of the points, where the red/green points are for the measured/desired configuration. (c/d) the left/third camera views.
    }
    \label{fig:tubular_setup}
	\end{minipage}
	\\
      \vspace{0.3cm}
	\begin{minipage}{\linewidth}
        \centering
    \includegraphics[width=0.85\linewidth]{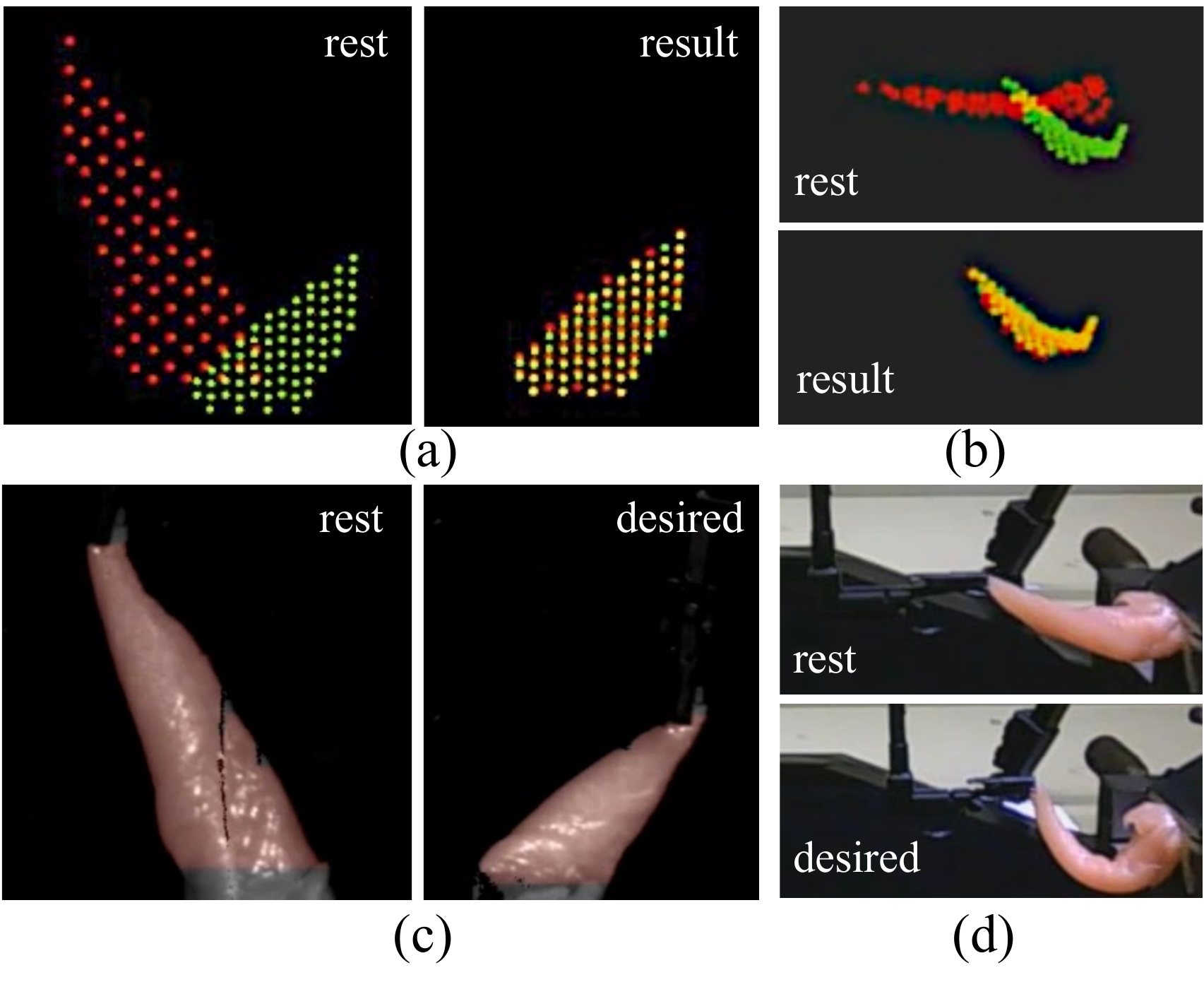}
    \vspace{-0.3cm}
    \caption{The volumetric tissue experiment.
    (a/b) the front/side views of the points, where the red/green points are for the measured/desired configuration. (c/d) the left/third camera views.
    }
    \label{fig:tissue_setup}
	\end{minipage}
        \\
        \vspace{0.3cm}
	\begin{minipage}{\linewidth}
     \centering
    \includegraphics[width=\linewidth]{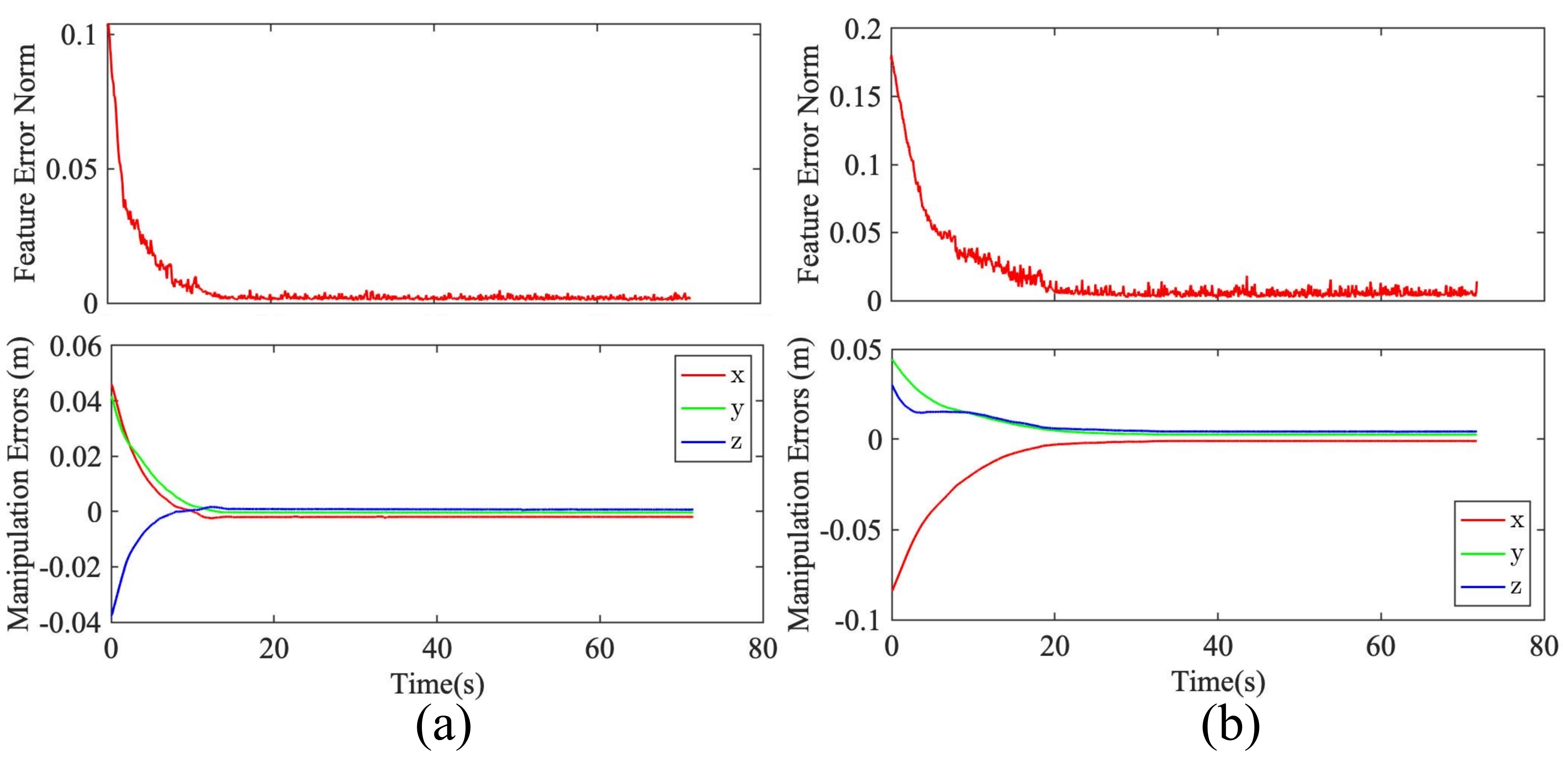}
    \vspace{-0.3cm}
    \caption{The result curves of the deformation feature errors and the manipulation errors of: (a) the tubular object experiment;
    (b) the volumetric tissue experiment. \\
    }
    \label{fig:tubular_tissue}
	\end{minipage}
	\vspace{-0.2cm}
\end{figure}

\subsubsection{7.2.3. Pushing a tubular object to the left:}
We selected the deformable tubular object to be a silicone colon model for validation.
As shown in Figure~\ref{fig:tubular_setup}(c,d), we set the desired deformation to be: pushing the colon model forward and to the left.
The comparisons between the measured and the desired points of the object are given in Figure~\ref{fig:tubular_setup}(a,b), from which we can see that the object reached its desired deformation in general.
The CD between the resulting and desired points is: $d_{CD} = 2.50(mm)$.
In addition, the result curves in Figure~\ref{fig:tubular_tissue}(a) show the minimization of both the deformation feature errors and the manipulation errors.

\subsubsection{7.2.4. Pushing a volumetric tissue down and to the right:}
We selected a volumetric tissue, a block of chicken tissue, to validate our controller.
As shown in Figure~\ref{fig:tissue_setup}(c,d), we set the desired deformation to be: pushing the tissue block down and to the right.
The comparisons between the measured and the desired points of the object are given in  Figure~\ref{fig:tissue_setup}(a,b), from which we can see that the object reached its desired deformation in general.
The CD between the resulting and desired points is: $d_{CD} = 2.29(mm)$.
In addition, the result curves in Figure~\ref{fig:tubular_tissue}(b) show the minimization of both the deformation feature errors and the manipulation errors.

\begin{figure}[t]
	\centering
	\begin{minipage}{\linewidth}
  \centering
  \includegraphics[width=0.85\linewidth]{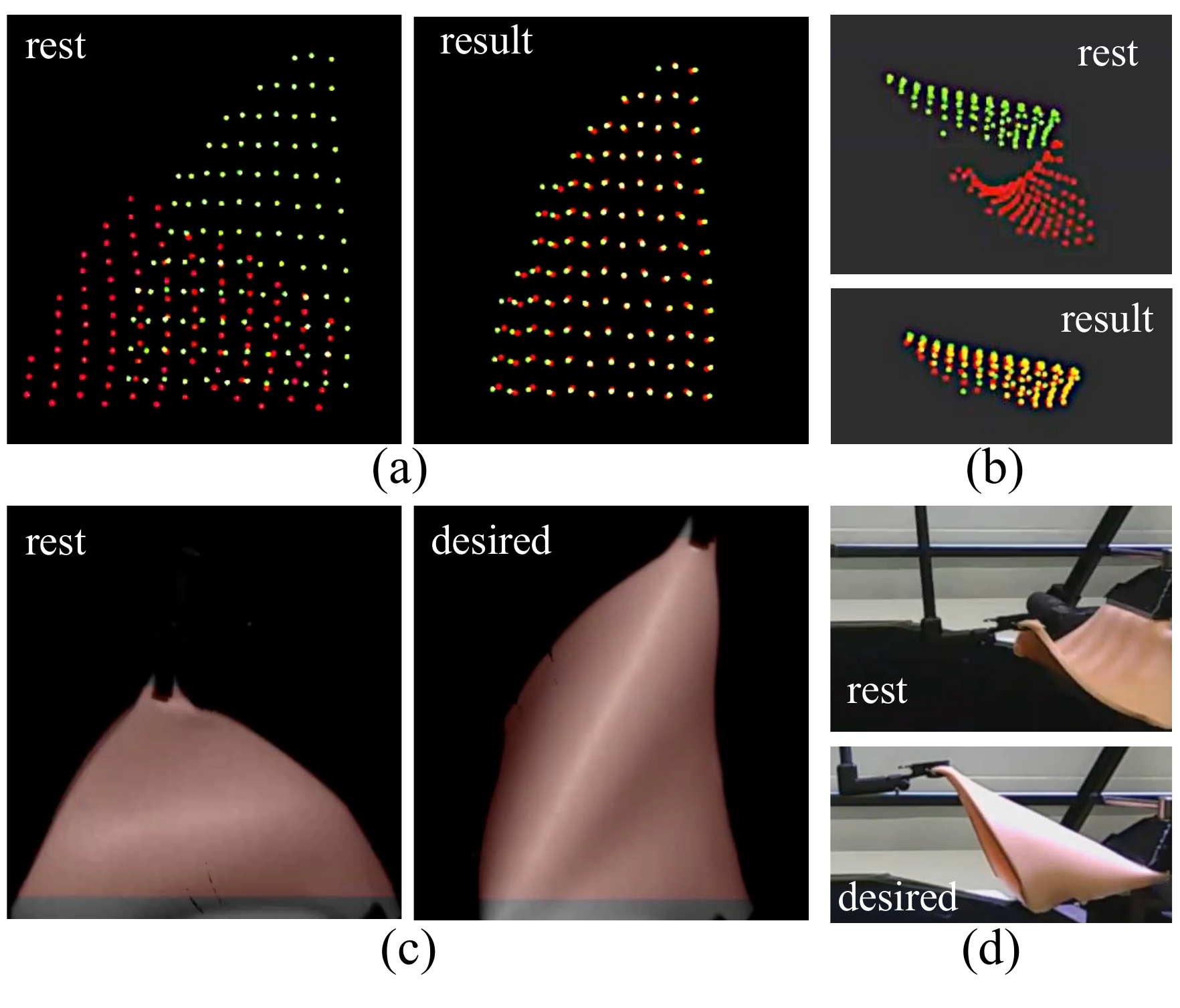}
  \vspace{-0.3cm}
    \caption{The silicon sheet experiment.
    (a/b) the front/side views of the points, where the red/green points are for the measured/desired configuration. (c/d) the left/third camera views.
    }
    \label{fig:graph_shape_setup}
	\end{minipage}
	\\
     \vspace{0.3cm}
	\begin{minipage}{\linewidth}
 \centering
    \includegraphics[width=\linewidth]{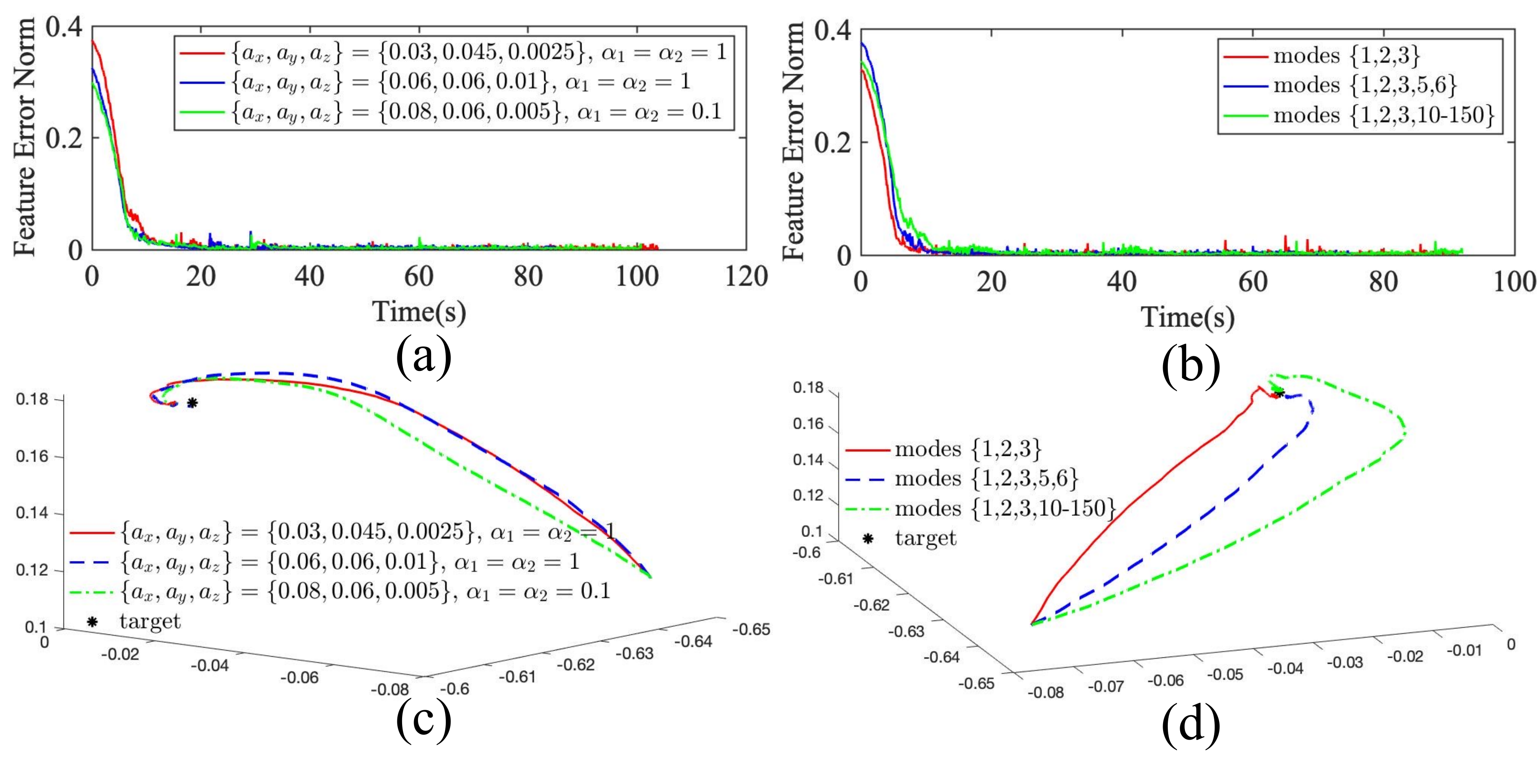}
    \vspace{-0.3cm}
    \caption{The result curves of the deformation feature errors and the manipulation trajectories for: (a,c) the experiments with different graph shapes;
    (b,d) the experiments with different modes.
    }
    \label{fig:graph_shape_sub_modes}
	\end{minipage}
 	\\
  \vspace{0.3cm}
	\begin{minipage}{\linewidth}
 \centering
    \includegraphics[width=\linewidth]{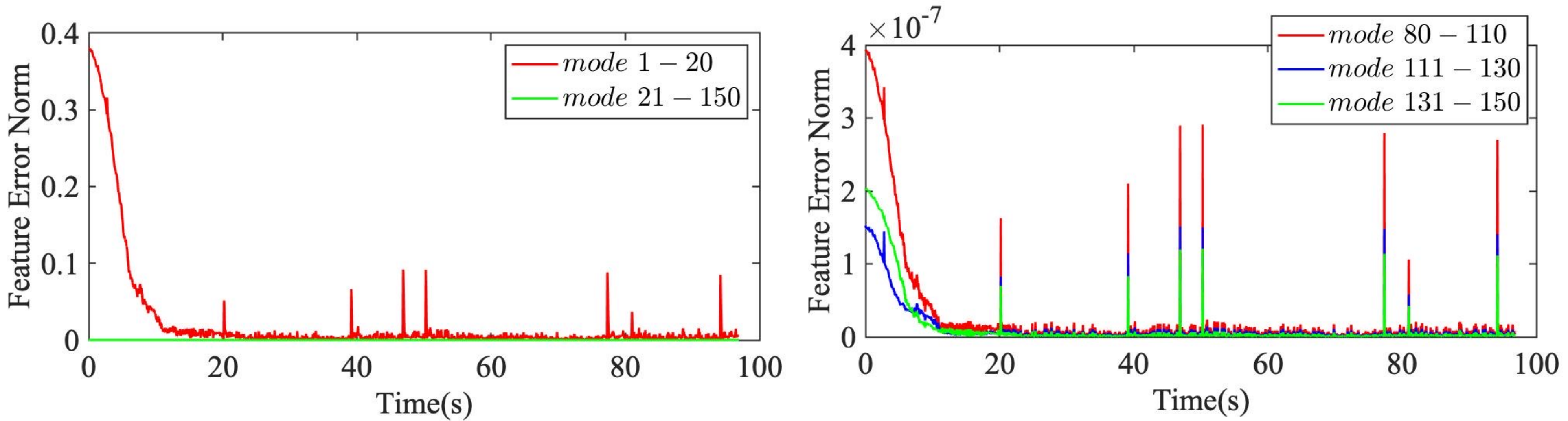}
    \vspace{-0.3cm}
    \caption{Feature error curves for the low-frequency modes used for control and the high-frequency modes that are not directly actuated.
    The bulges in the feature error curves are caused by the added measurement noises and occlusions.
    }
    \label{fig:high_modes}
	\end{minipage}
	\vspace{-0.3cm}
\end{figure}

\subsubsection{7.2.5. With different graph shapes:}
To show the influence of the graph shapes, we conducted a group of comparative experiments by setting different shape parameters to the benchmark graph used in this section.
As shown in Figure~\ref{fig:graph_shape_setup}(c,d), we selected a rectangular silicone sheet whose length, width, and height are about 0.16(m), 0.12(m), and 0.01(m) to be the object.
We set three cases:
the benchmark graph in the ellipsoidal shape with $\left \{ a_x, a_y,a_z\right \} = \left \{ 0.03, 0.045,0.0025\right \} (m)$ and $\alpha_1 = \alpha_2 = 1$;
a different-sized ellipsoidal graph with $\left \{ a_x, a_y,a_z\right \} = \left \{ 0.06, 0.06, 0.01\right \} (m)$ and $\alpha_1 = \alpha_2 = 1$;
a graph in the shape of a parallelepiped with $\left \{ a_x, a_y,a_z\right \} = \left \{ 0.08, 0.06, 0.005\right \} (m)$ and $\alpha_1 = \alpha_2 = 0.1$ (with the most similar shape to the object shape).

Results in Figure~\ref{fig:graph_shape_sub_modes}(a) 
show the minimization of the
deformation feature errors for all cases.
Changes in the graph shape only influence the deformation processes (illustrated by the manipulation trajectories in Figure~\ref{fig:graph_shape_sub_modes}(c)), and the manipulation target is always reached.
The more similar the shape of the modal graph is to the object, the better the manipulation trajectory.

\subsubsection{7.2.6. With different selections of modes:}
Experiments with different mode sets were also conducted
to further discuss the selection of modes.
We set the deformation task to be the same as Figure~\ref{fig:graph_shape_setup}(c,d).
First, we show the behaviors of the first 20 modes (that are embedded in our modal graph) and some high-frequency modes during manipulation with occasionally added measurement noises and occlusions.
The feature error curves in Figure~\ref{fig:high_modes}(a) show that:
1) the amplitudes of the high-frequency modes are much smaller than the low-frequency modes;
2) high-frequency modes are more sensitive to noises and occlusions.
Then, we show the performance of our method using different mode sets.
As our method requires using at least the first $3k$ modes, we compared three cases:
using modes $\left \{ 1,2,3 \right \} $;
using modes $\left \{ 1,2,3,5,6 \right \} $;
using modes $\left \{ 1,2,3,10-150 \right \} $.
Results in Figure~\ref{fig:graph_shape_sub_modes}(b) 
show the minimization of the
deformation feature errors for all cases.
Changes in the graph shape only influence the deformation processes (illustrated by the manipulation trajectories in Figure~\ref{fig:graph_shape_sub_modes}(d)), and the manipulation target is always reached.

Based on the above results, we discuss the selection of modes:
1) as high-frequency modes have little effect, we can only use a small number of low-frequency modes in our controller;
2) our method can work with a subset of the low-frequency modes as long as the first $3k$ modes are used.

\begin{figure}[!t]
	\centering
	\begin{minipage}{\linewidth}
  \centering
    \includegraphics[width=0.82\linewidth]{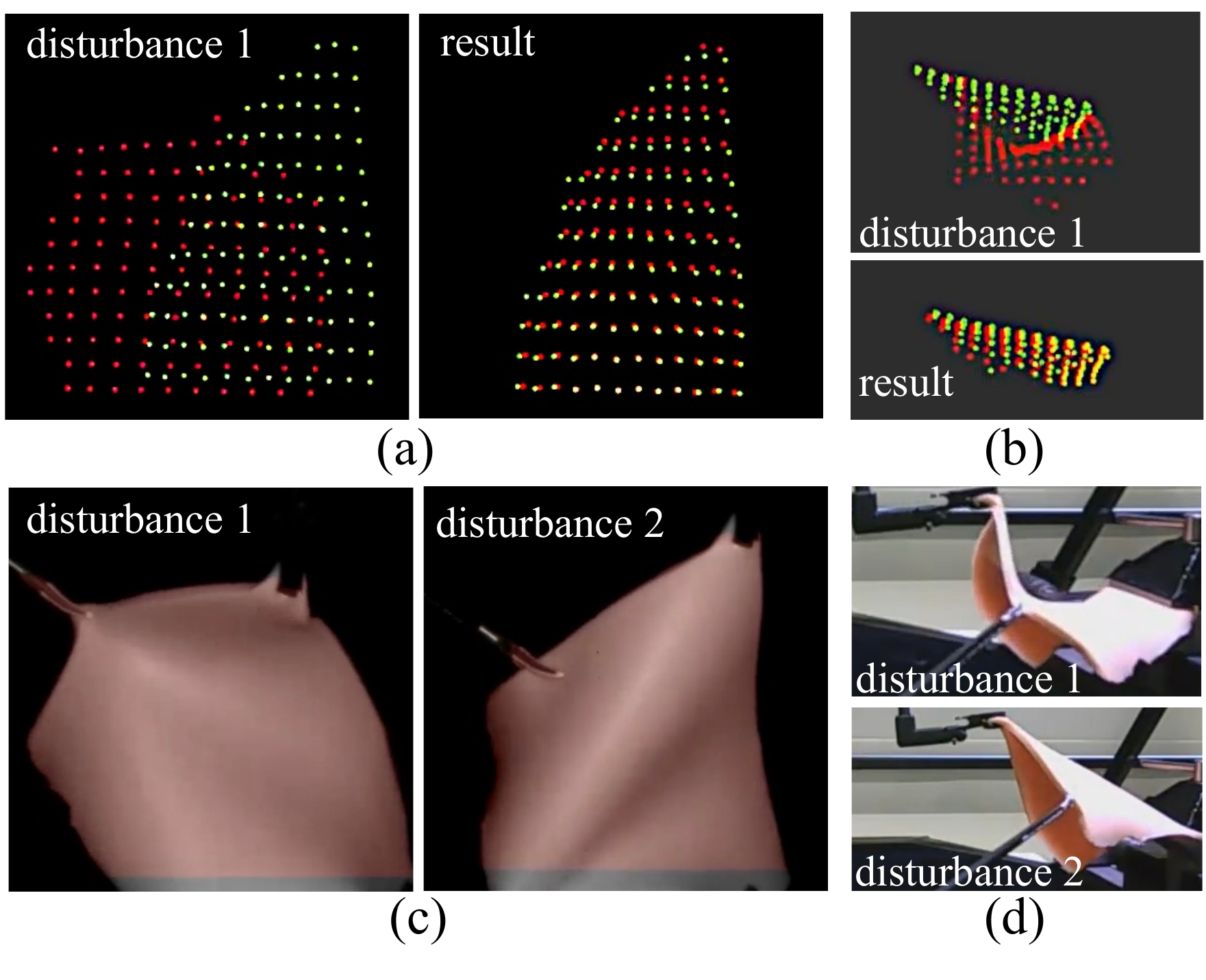}
      \vspace{-0.4cm}
    \caption{The experiments with external disturbances. (a/b) the front/side views of the points, where the red/green points are for the measured/desired configuration. (c/d) the left/third camera views.
    }
    \label{fig:disturbance_setup}
	\end{minipage}
	\\
     \vspace{0.3cm}
	\begin{minipage}{\linewidth}
 \centering
    \includegraphics[width=0.82\linewidth]{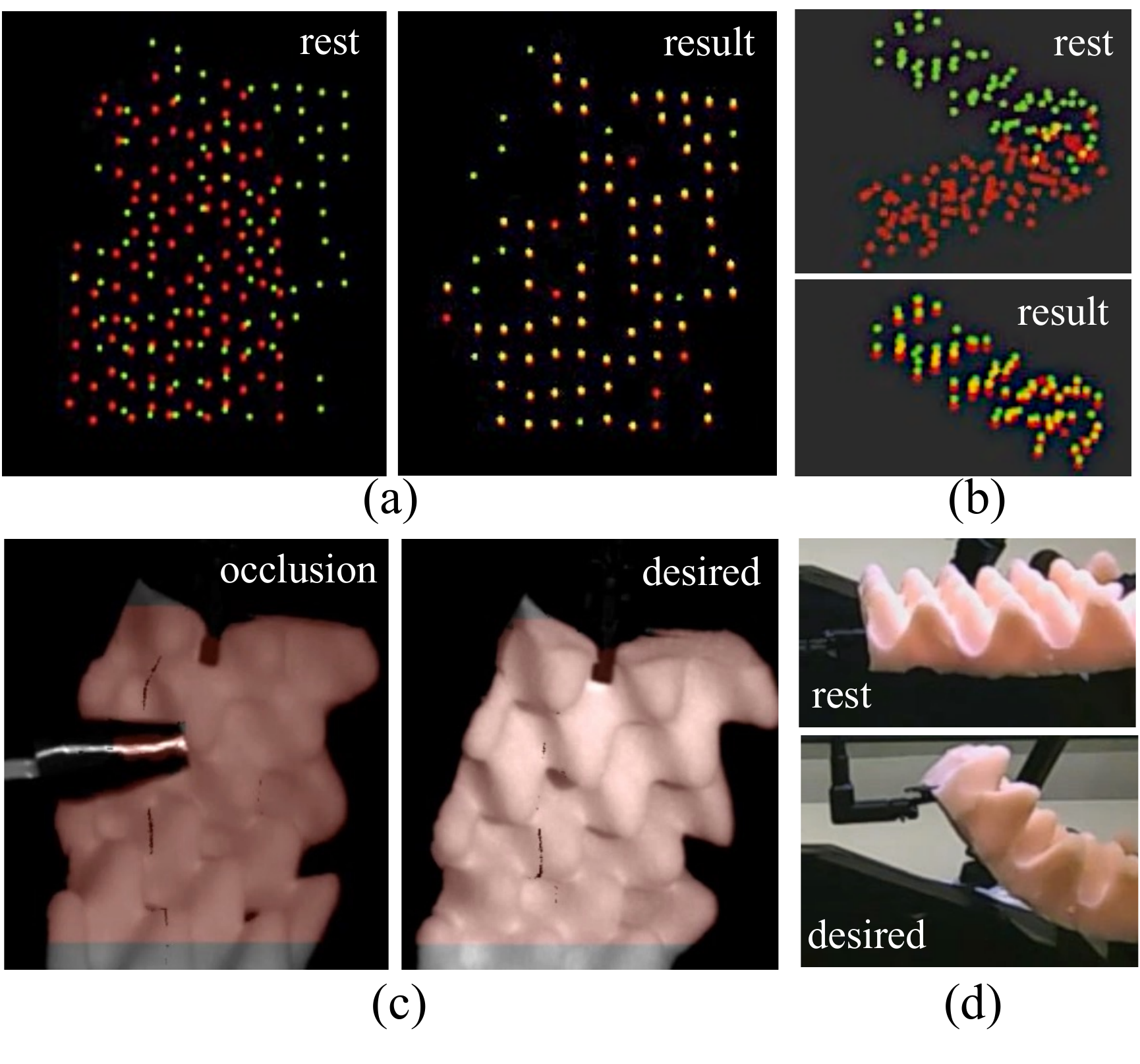}
    \vspace{-0.4cm}
    \caption{The experiments with the complex-shaped sponge block under occlusions. (a/b) the front/side views of the points, where the red/green points are for the measured/desired configuration. (c/d) the left/third camera views.
    }
    \label{fig:occlussions_setup}
	\end{minipage}
 	\\
  \vspace{0.3cm}
	\begin{minipage}{\linewidth}
 \centering
    \includegraphics[width=\linewidth]{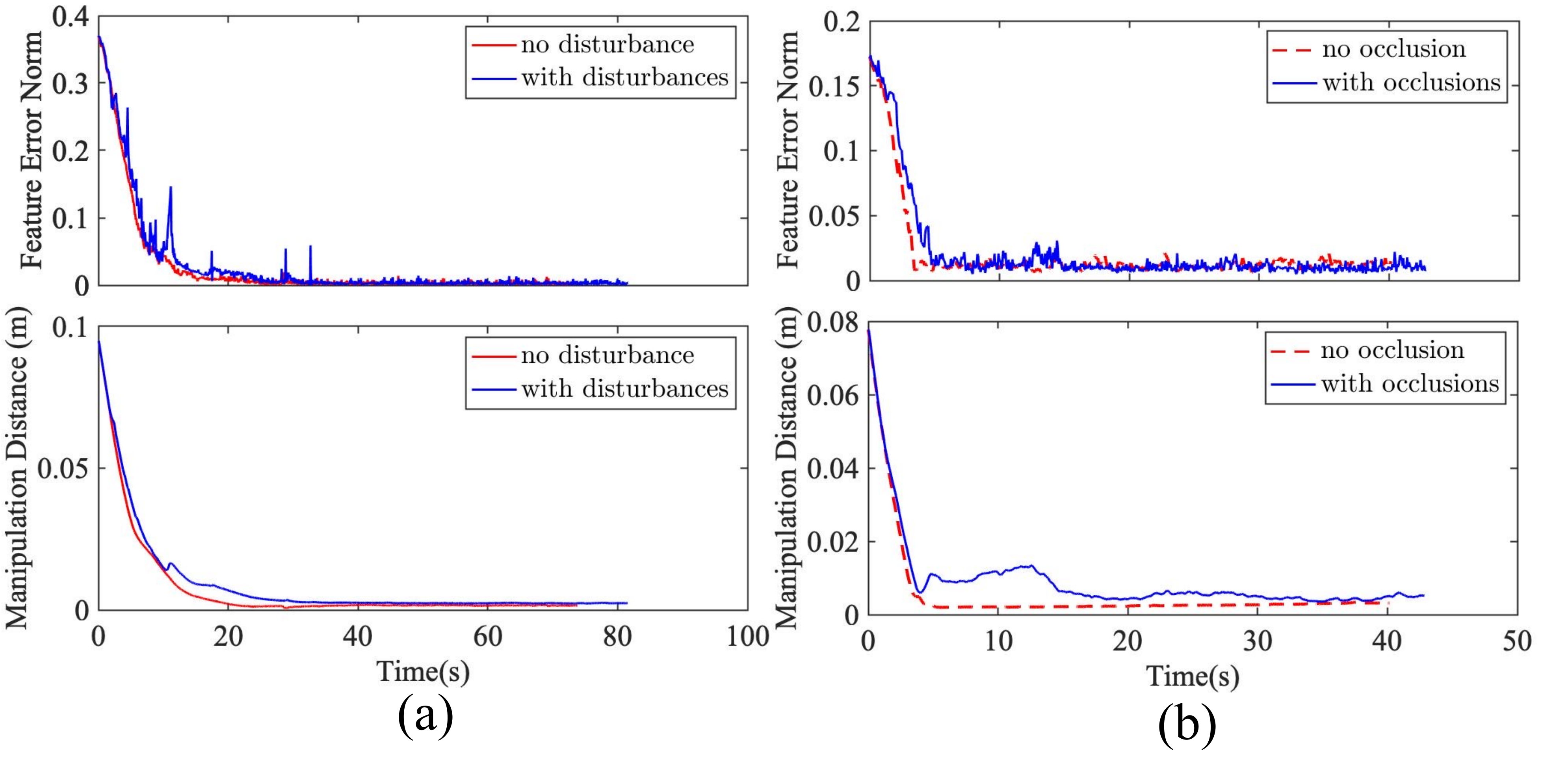}
    \vspace{-0.6cm}
    \caption{The result curves of the deformation feature errors and the manipulation distance
    ($\left \| \mathbf{e}_d(t) \right \|$)
    for: (a) the experiments with external disturbances.
    (b) the experiments with the complex-shaped sponge block under occlusions.
     The bulges in the feature error curves are caused by disturbances/occlusions and sensing noises.
    The turning points in the manipulation distance curves are caused by disturbances/occlusions.
    }
    \label{fig:disturbance_occlussion}
	\end{minipage}
	\vspace{-0.3cm}
\end{figure}

\subsection{7.3. Cases Under Different Settings}
\subsubsection{7.3.1. Under external disturbances:}
For the above silicon sheet experiment (Figure~\ref{fig:graph_shape_setup}(c,d)), 
we manually added external disturbances by pulling the object during the manipulation process (Figure~\ref{fig:disturbance_setup}(c,d)).
Note that the pulling operations not only disturbed the manipulation but also induced dynamic deformations that are not modeled in our controller (that is formulated under the quasi-static assumption).
The comparisons between the measured and the desired points of the object in
the experiments without and with the external disturbances are given in Figure~\ref{fig:graph_shape_setup}(a,b) and Figure~\ref{fig:disturbance_setup}(a,b), respectively.
We can see that the object reached its desired deformation in general for both cases.
The CD for the case without disturbances is $d_{CD} = 3.06(mm)$, and the CD for the case with disturbances is $d_{CD} = 3.81(mm)$.
We also compare their result curves in Figure~\ref{fig:disturbance_occlussion}(a).
The results show that:
although the external disturbances cause bumps in the curves of the feature errors and the manipulation distance, our control successfully achieves the minimization of these errors.
The input-to-state stability of our controller with the external disturbances is validated.

\subsubsection{7.3.2. Complex-shaped object with occlusions:}
To further validate our controller, as shown in Figure~\ref{fig:occlussions_setup}, we selected a sponge block with a complex shape.
In the experiments,
the visible parts of this complex-shaped object changed not only with robot manipulation, but also with the shadows produced by its bulges.
The comparisons between the measured and the desired points of the object
are given in Figure~\ref{fig:occlussions_setup}(a,b), from which we can see that the object reached its desired deformation in general.
The CD for the case without occlusions is $d_{CD} = 3.90(mm)$, and the CD for the case with occlusions is $d_{CD} = 4.07(mm)$.
In addition, we also manually and occasionally put external occlusions by placing a black card above the object (as shown in Figure~\ref{fig:occlussions_setup}(c)) during the deformation and after the convergence of the deformation feature errors to the steady state.
We compare the result curves of the experiments without and with the external occlusions in Figure~\ref{fig:disturbance_occlussion}(b).
The added occlusions cause bumps in the curves of the feature errors and the manipulation distance, but the minimization of these errors is successfully achieved.

\begin{figure}[t]
    \centering
    \includegraphics[width=0.825\linewidth]{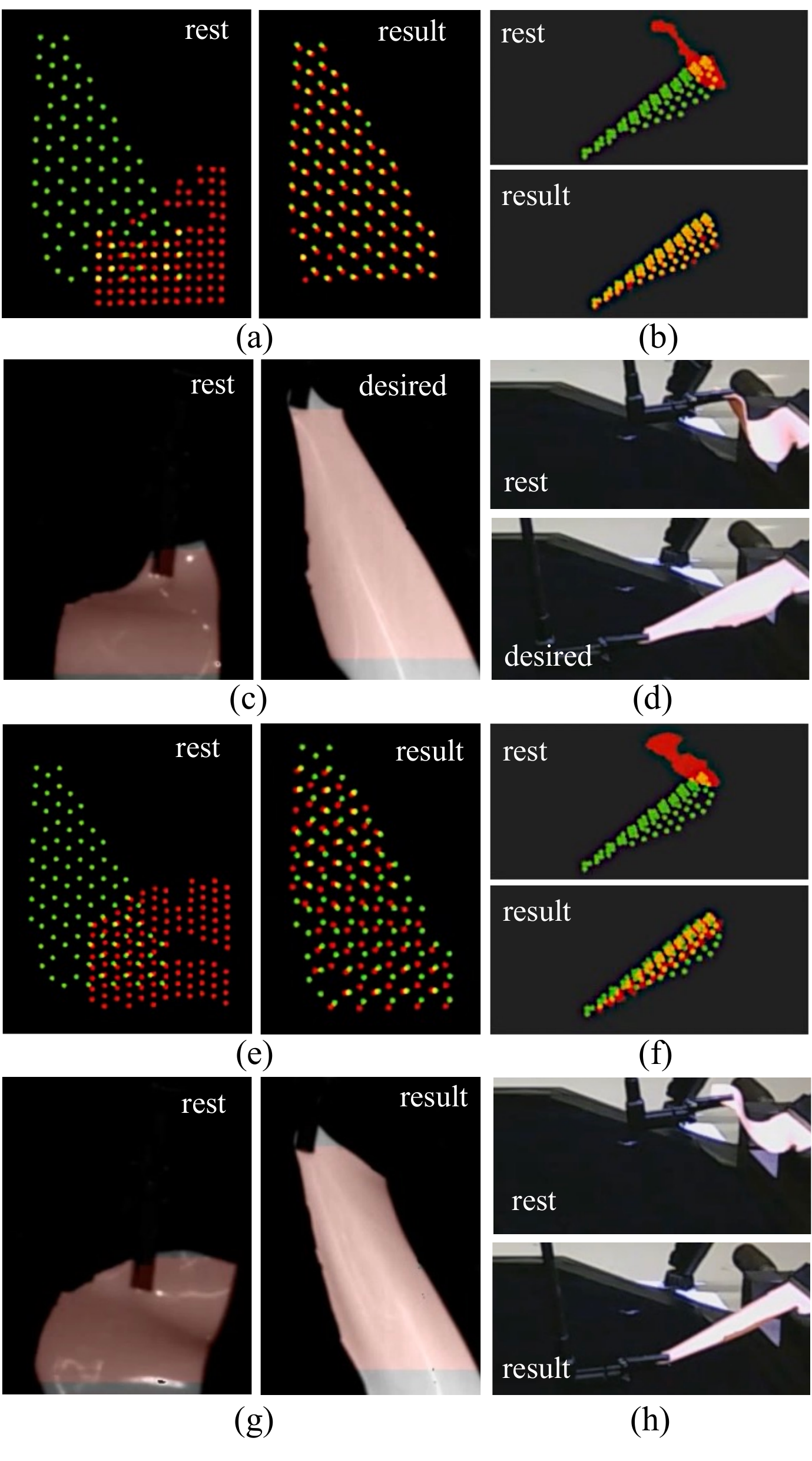}
    \caption{The experiments with large and unreachable desired deformations.
    For the experiment with large desired deformation:
    (a/b) the front/side views of the measured and desired points;
    (c/d) the left/third camera views.
    For the experiment with unreachable desired deformation:
    (e/f) the front/side views of the measured and desired points;
    (g/h) the left/third camera views of the object configurations.
    }
    \label{fig:large_setup}
\end{figure}

\begin{figure}[t]
    \centering
    \includegraphics[width=0.825\linewidth]{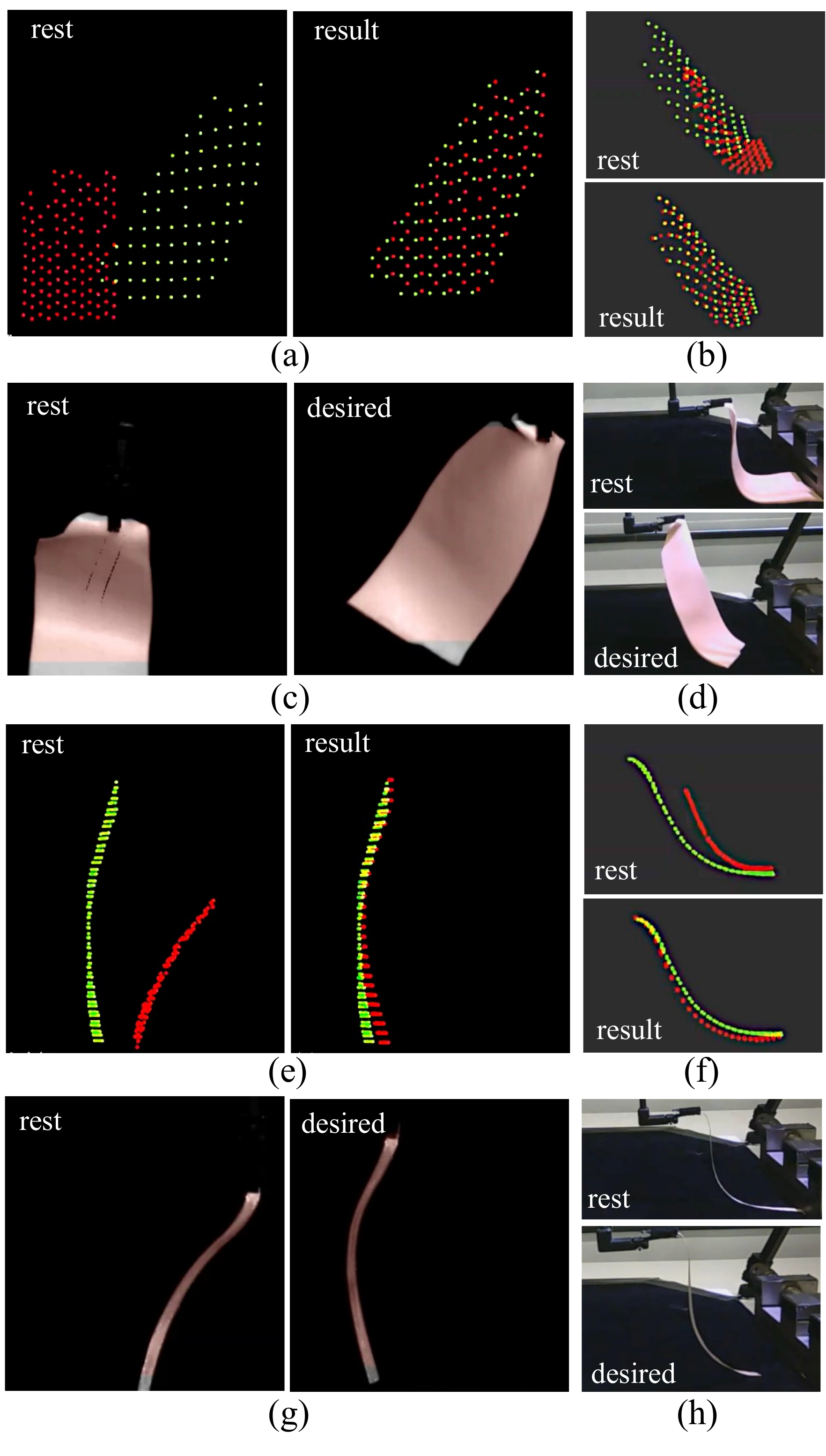}
    \caption{The experiments with non-fixed objects.
    For the case with the silicon sheet:
    (a/b) the front/side views of the measured and desired points;
    (c/d) the left/third camera views.
    For the case with the ribbon:
    (e/f) the front/side views of the measured and desired points;
    (g/h) the left/third camera views.
    }
    \label{fig:free_setup}
\end{figure}

\begin{figure}[t]
	\centering
    \includegraphics[width=\linewidth]{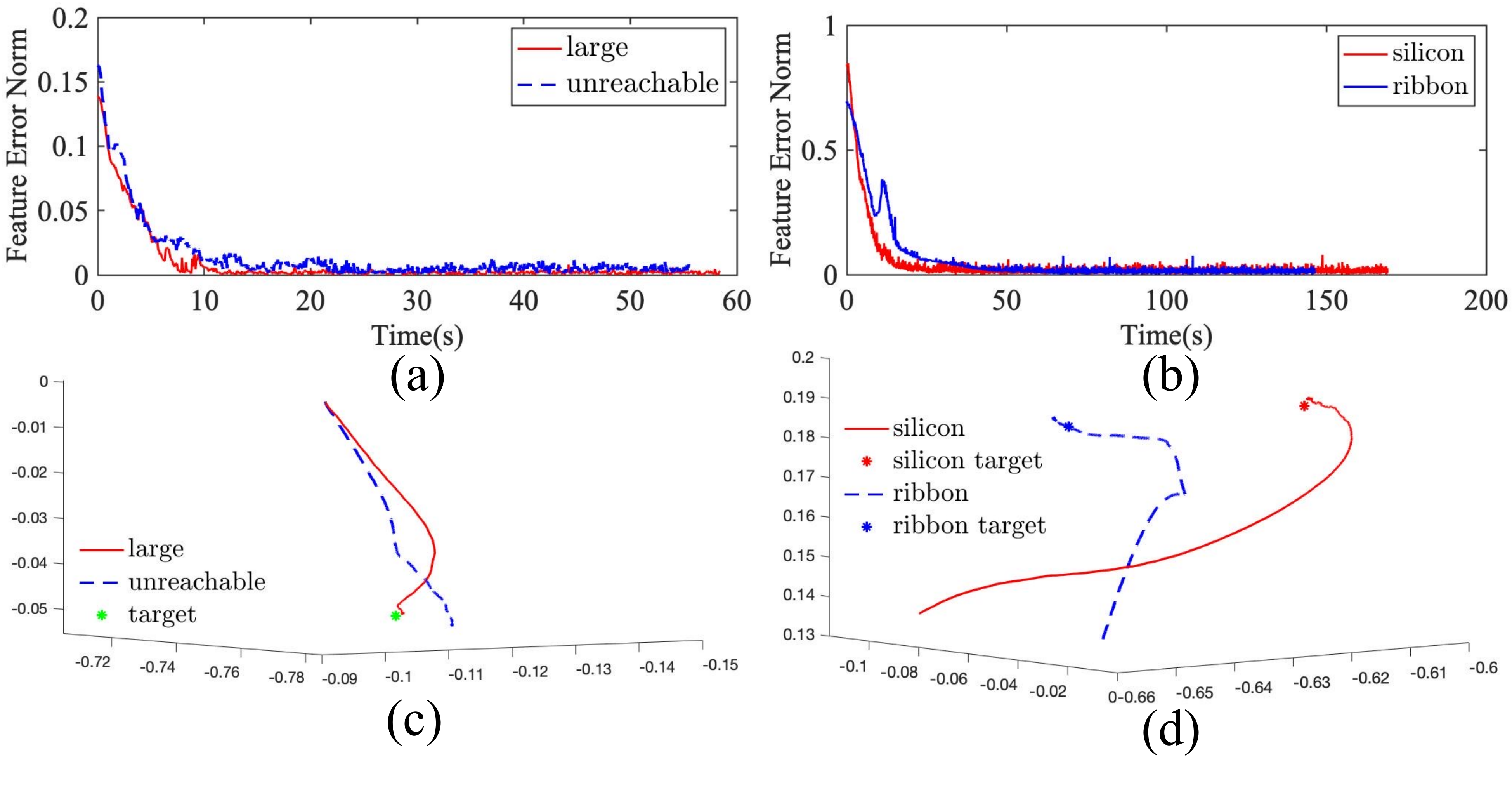}
    \caption{The result curves of the deformation feature errors and the manipulation trajectories for:
    (a,c) the experiments with large and unreachable desired deformations.
    The turning points in (a) are caused by the disturbances of the surface reflections during the manipulation;
    (b,d) the experiments with free-moving objects.
    }
    \label{fig:large_free}
\end{figure}

\subsubsection{7.3.3. Large deformation with unreachable target:}
We also validated our controller with relatively large deformation and an unreachable target.
As shown in Figure~\ref{fig:large_setup}(c,d), we selected a small and irregular-shaped silicon skin, and set a relatively large desired deformation by pulling it down to the right.
The comparisons between the measured and the desired points of the object
are given in Figure~\ref{fig:large_setup}(a,b), from which we can see that the measured deformation of the object (shown by the red points) reached its desired deformation (shown by the green points) in general.
The CD between the resulting and desired points is $d_{CD} = 2.23(mm)$.

Afterward, we set the robot back to the rest configuration but changed the manipulation point (shown in Figure~\ref{fig:large_setup}(g)).
In this way, the desired deformation in Figure~\ref{fig:large_setup}(c,d) became unreachable for the changed configuration.
We used the same controller to deform the object in this situation.
The object was deformed to the resulting state shown in Figure~\ref{fig:large_setup}(e,f).
We can see that the object was manipulated to a resulting state where the deformation feature errors are minimized (Figure~\ref{fig:large_free}(a,c)) and the measured points are in a similar shape to the desired points even though the desired target is unreachable.
The CD between the resulting and desired points is $d_{CD} = 7.04(mm)$.
On the other hand, as shown in Figure~\ref{fig:large_free}(c), the unreachable target does influence the manipulation trajectory and cause steady-state errors in the manipulation errors.

\subsubsection{7.3.4. With non-fixed objects:}
To further validate our controller with non-fixed objects, we remove the constraint that one end of the object is fixed in space.
We first conducted the experiment with a non-fixed silicon sheet Figure~\ref{fig:free_setup}(c,d).
We generated the desired deformation by manually moving the robot to drag the object up and to the right.
In this process, in addition to the object deformation caused by the manipulation, the contact between the object and the table also influenced the object shape.
Note that the deformation caused by this contact is not modeled in our modal-graph framework.
We set another case using a piece of thin ribbon (Figure~\ref{fig:free_setup}(e,f)).
Since the ribbon is very light, at the rest configuration, we (non-fixedly) tucked one end of the ribbon into the narrow gap between the vise and the table for less than $5$ mm.
As the ribbon is being manipulated, this end would be pulled out at some moment, further causing dynamic disturbances to our controller (that is formulated under the quasi-static assumption).

The comparisons between the measured and the desired points of the object
in the two cases are given in Figure~\ref{fig:free_setup}(a,b) and (e,f), respectively.
We can see that the silicon sheet reached its desired deformation in general.
The CD for the silicon case is $d_{CD} = 4.12(mm)$, and the CD for the ribbon stage is $d_{CD} = 7.73(mm)$.
As to the ribbon, the part close to the grasping point converges roughly to the desired target, but the part far from the grasping point only converges to a shape similar to the desired target.
In addition, results in Figure~\ref{fig:large_free}(b)
show the minimization of the
deformation feature errors for both cases.
The manipulation trajectories in Figure~\ref{fig:large_free}(d) show that the objects are manipulated to the resulting configurations that are close to their targets. 
The turning points of the ribbon case happened when the non-fixed end of the object was pulled out and fell on the table and, began to move with the manipulation.
The results show that the unmodeled effects of the non-fixed objects only cause small steady-state errors, validating the robustness of our controller.

\begin{figure}[t]
    \centering
    \includegraphics[width=0.825\linewidth]{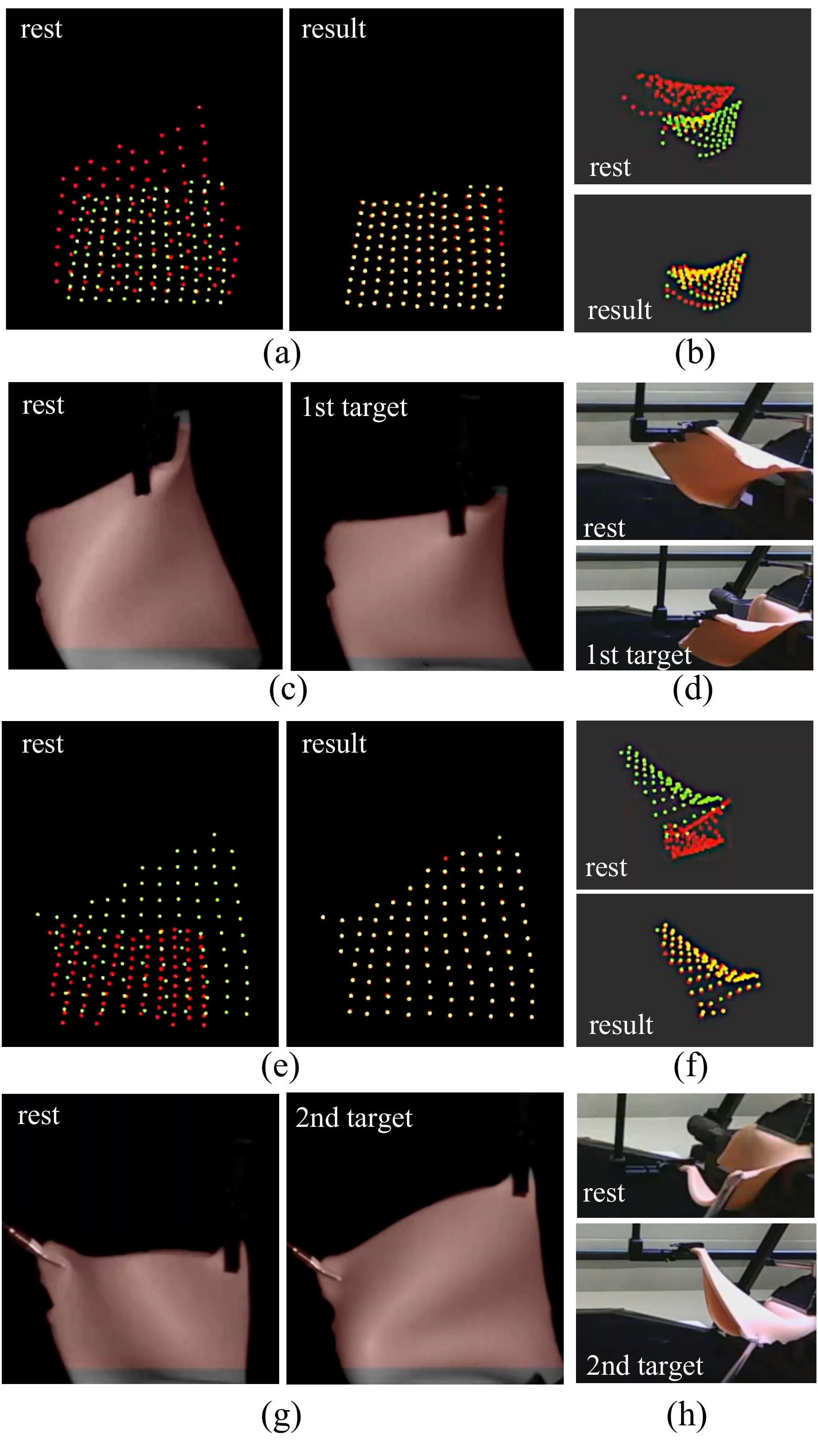}
    \caption{The two-stage experiments with changed manipulation points and boundary conditions.
    For the first stage:
    (a/b) the front/side views of the measured and desired points;
    (c/d) the left/third camera views.
    For the second stage:
    (e/f) the front/side views of the measured and desired points;
    (g/h) the left/third camera views.
    }
    \label{fig:mp_setup}
\end{figure}

\begin{figure}[!t]
	\centering
	\begin{minipage}{\linewidth}
  \centering
    \includegraphics[width=0.825\linewidth]{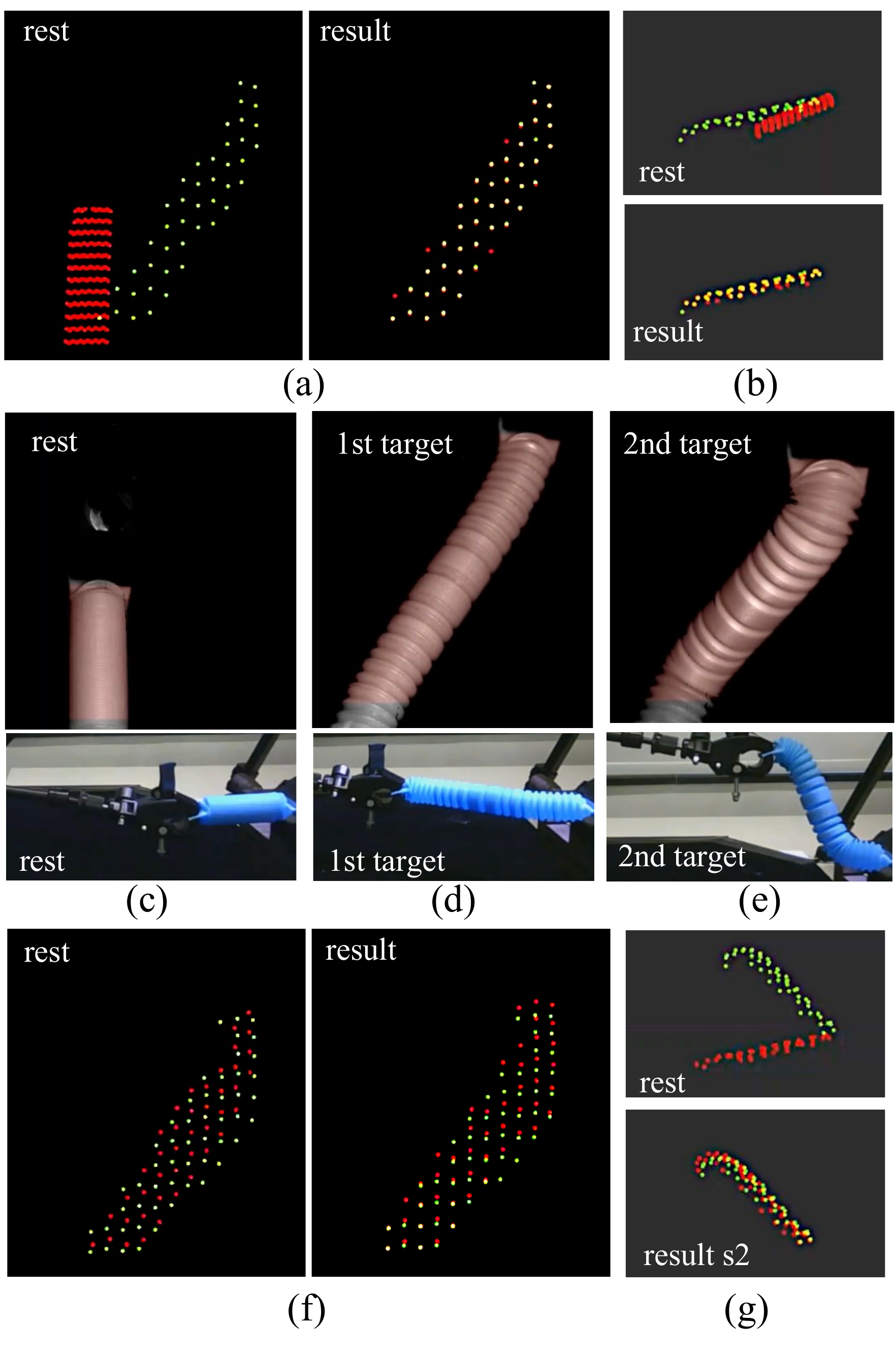}
    \vspace{-0.3cm}
    \caption{The experiments with elastic-plastic deformations.
    For the first stage:
    (a/b) the front/side views of the measured and desired points;
    (c/d) the left/third camera views.
    For the second stage:
    (d/e) the front/side views of the object configurations;
    (f/g) the left/third camera views.
    }
    \label{fig:plastic_setup}
	\end{minipage}
	\\
     \vspace{0.3cm}
	\begin{minipage}{\linewidth}
 \centering
   \includegraphics[width=\linewidth]{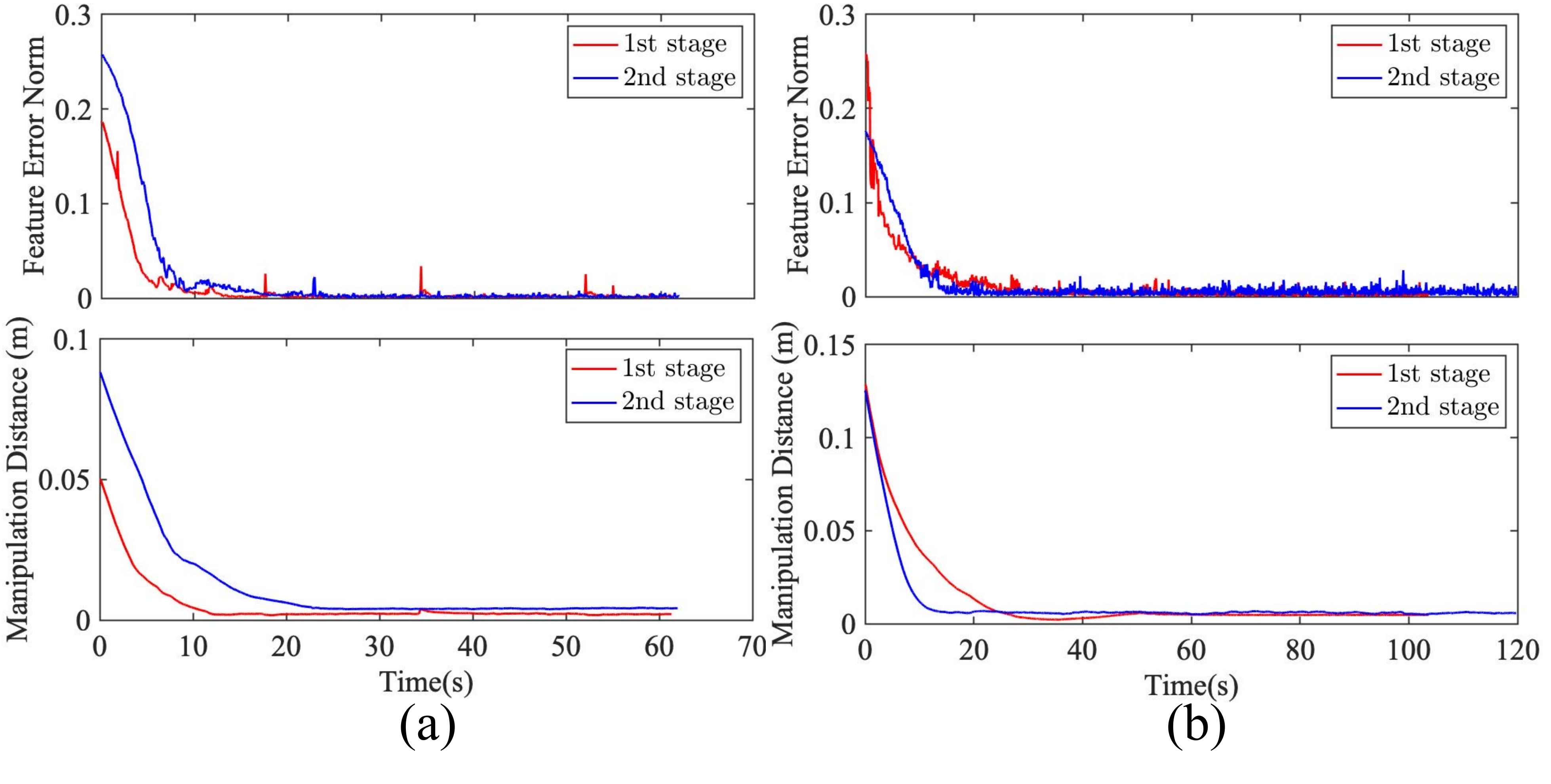}
   \vspace{-0.3cm}
    \caption{The result curves of the deformation feature errors and the manipulation distance  ($\left \| \mathbf{e}_d(t) \right \|$)
    (a) the experiments with changed manipulation points and boundary conditions.
    (b) the experiments with elastic-plastic deformations.
    }
    \label{fig:mp_plastic}
	\end{minipage}
	\vspace{-0.3cm}
\end{figure}

\subsubsection{7.3.5. With changes of the manipulation point and boundary conditions:}
We validated our controller using the silicon sheet in a two-stage task with changes in the manipulation point and boundary conditions.
In the first stage (Figure~\ref{fig:mp_setup}(c,d)), only one end of the object was fixed in space, and we selected a manipulation point to deform the object to the first target.
In the second stage (Figure~\ref{fig:mp_setup}(g,h)), we further fixed another end of the object, changed the manipulation point, and deformed the object to the second target.
In this way, both the manipulation points and the boundary conditions of the two stages are different.
However, we used the same modal graph and the same controller for these stages.
The comparisons between the measured and the desired points of the object in
the two stages are given in Figure~\ref{fig:mp_setup}(a,b) and (e,f), respectively.
We can see that the object reached its desired deformation in general for both stages.
The CD for the first stage is $d_{CD} = 3.52(mm)$, and the CD for the second stage is $d_{CD} = 2.69(mm)$.
In addition, the result curves in Figure~\ref{fig:mp_plastic}(a) show the minimization of the feature errors and the manipulation errors for both stages.

\subsubsection{7.3.6. With elastic-plastic deformations:}
We also validated our controller with elastic-plastic deformations.
As shown in Figure~\ref{fig:plastic_setup}, we selected a complaint plastic expansion bellows to be the object.
Note that the elastic-plastic deformation of the bellows is path dependent.
To generate a relatively large deformation, we need to conduct a two-stage task.
In the first-stage experiment, we stretched the bellows to the right (Figure~\ref{fig:plastic_setup} (c) to (d)).
In the second-stage experiment, we bent the bellows forward and upward (Figure~\ref{fig:plastic_setup} (d) to (e)).

The comparisons between the measured and the desired points of the object in the two stages are given in Figure~\ref{fig:plastic_setup}(a,b) and (f,g), respectively.
We can see that in the first stage, the object reached its desired deformation in general.
As for the second stage, the resulting shape of the object is close to the desired shape.
The CD for the first stage is $d_{CD} = 3.09(mm)$, and the CD for the second stage is $d_{CD} = 7.22(mm)$.
In addition, the result curves in Figure~\ref{fig:mp_plastic}(b) show the minimization of the feature errors and the manipulation errors for both stages.
The results validate that:
although our modal-graph framework does not model plastic deformations, the proposed adaptive robust controller can tackle these unmodeled effects.

\begin{figure}[!t]
	\centering
	\begin{minipage}{\linewidth}
  \centering
    \includegraphics[width=0.825\linewidth]{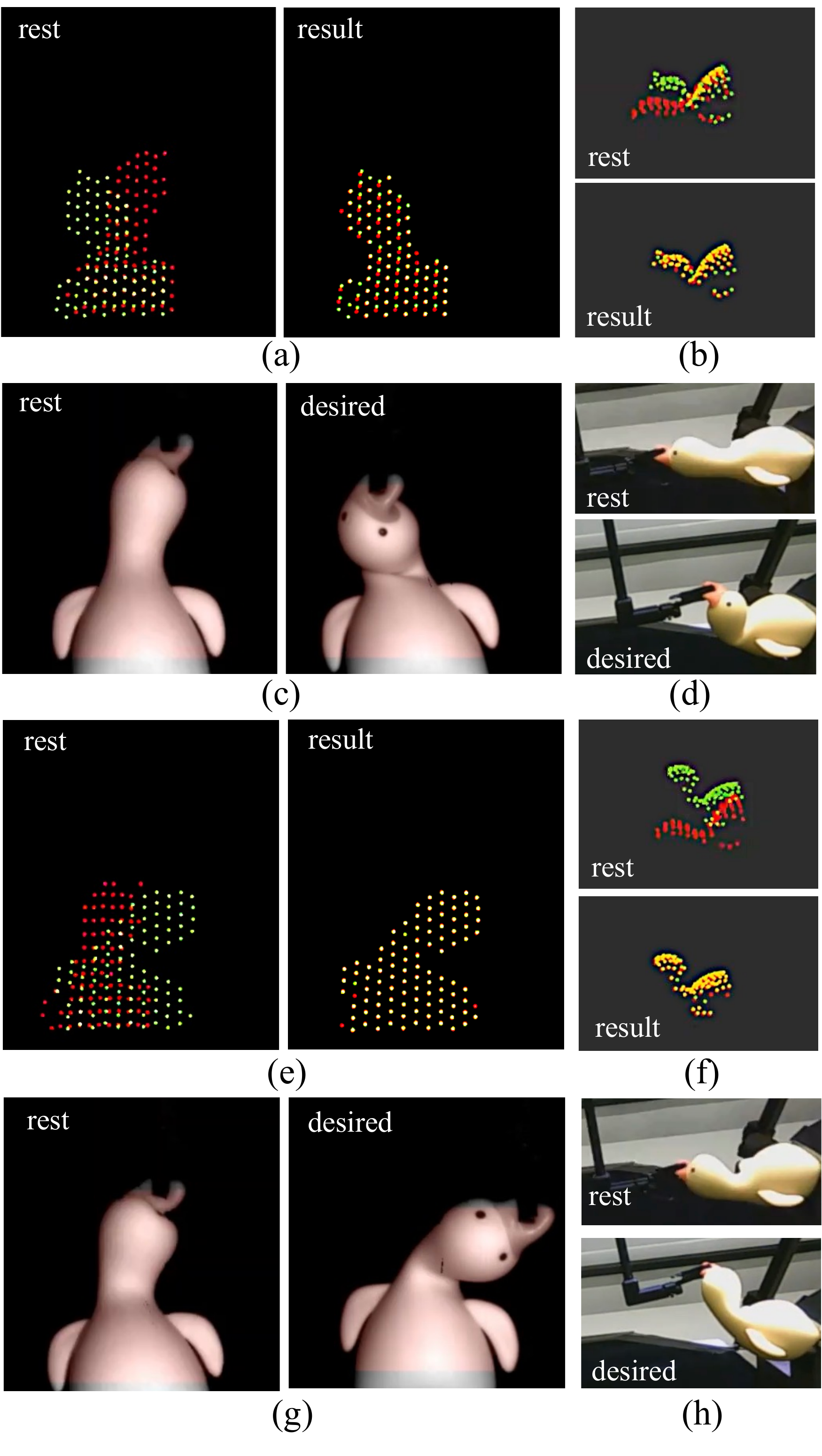}
    \vspace{-0.3cm}
    \caption{The experiments with both translational and rotational manipulations.
    For the case where the desired deformation is to the left:
    (a/b) the front/side views of the measured and desired points;
    (c/d) the left/third camera views.
    For the case where the desired deformation is to the right:
    (e/f) the front/side views of the measured and desired points;
    (g/h) the left/third camera views.
    }
    \label{fig:rotation_setup}
	\end{minipage}
	\\
     \vspace{0.3cm}
	\begin{minipage}{\linewidth}
 \centering
    \includegraphics[width=\linewidth]{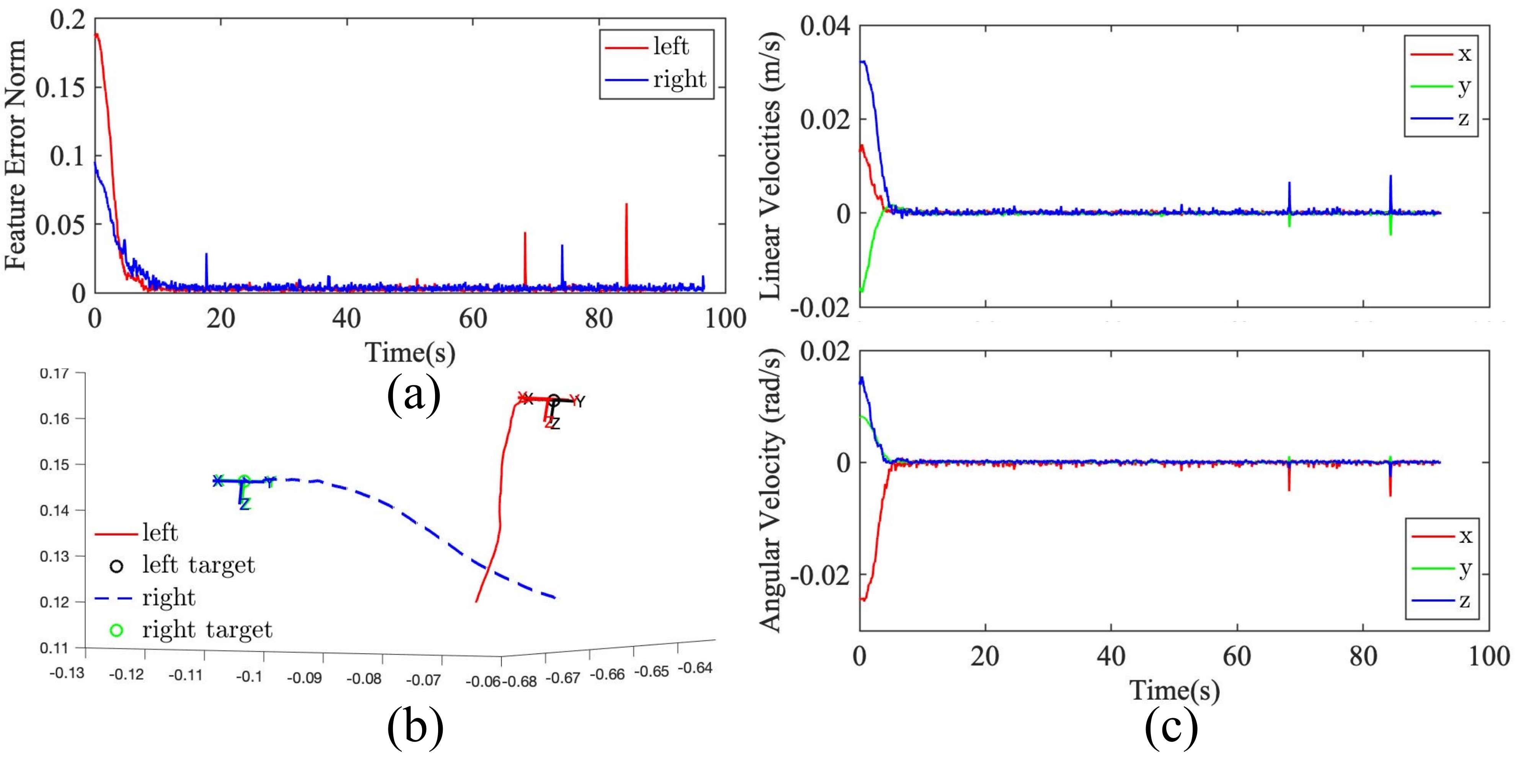}
    \vspace{-0.3cm}
    \caption{Results of the experiments with both translational and rotational manipulations. 
    (a) the result curves of deformation feature errors.
    (b) the manipulation trajectories of the two cases.
    (c) the linear and angular velocity commands of $\mathbf{r}$ for the "left" case;
    }
    \label{fig:rotation}
	\end{minipage}
	\vspace{-0.3cm}
\end{figure}

\subsubsection{7.3.7. With both translational and rotational manipulations:}
We also conducted experiments to show our modal graph framework can control both the translational and rotational motions of the robot manipulation through a simple extension of the feature inversion function.
At each control loop, we can extract the rotational component (locally around the manipulation point) from the changing modal features using the modal rotation technique in \cite{choi2005modal}:
\begin{equation}
\begin{aligned}
        \mathbf{w}_{\boldsymbol{\gamma}}(\mathbf{r},t) 
        & = \frac{1}{2} (\mathbf{\nabla} \times) \mathbf{\Psi}(\mathbf{r},\mathbf{n}_r\mathbf{\Phi}(\mathbf{n}_r) \mathbf{s}(t) \\ 
    & = \mathbf{\Psi}_{\mathbf{w}}(\mathbf{r},\mathbf{n}_r)\mathbf{\Phi}(\mathbf{n}_r)
    \Delta \mathbf{s}(t)
\end{aligned} 
\end{equation}
where $\mathbf{w}_{\boldsymbol{\gamma}}(o,t) \in \mathbb{R}^3$ is the rotation vector of the manipulation point's projection on the graph, and the product $\mathbf{\Psi}_{\mathbf{w}}(\mathbf{r},\mathbf{n}_r)\mathbf{\Phi}(\mathbf{n}_r) \in \mathbb{R}^{3 \times m}$ is the modal rotation matrix. 
Then, we extend the feature inversion function to be:
\begin{equation}
    \mathbf{D}(\mathbf{r},\Delta \mathbf{s}(t)) = 
        \begin{bmatrix}
    \Delta \mathbf{u}_{\boldsymbol{\gamma}}(\mathbf{r},t)   \\ \Delta \mathbf{w}_{\boldsymbol{\gamma}}(\mathbf{r},t)
\end{bmatrix} = \overline{\mathbf{\Psi}}(\mathbf{r},\mathbf{n}_r) \mathbf{\Phi}(\mathbf{n}_r)
    \Delta \mathbf{s}(t)
\end{equation}
where $\overline{\mathbf{\Psi}}(\mathbf{r},\mathbf{n}_r) = \begin{bmatrix}
\mathbf{\Psi}(\mathbf{r},\mathbf{n}_r),  & \mathbf{\Psi}_{\mathbf{w}}(\mathbf{r},\mathbf{n}_r)
\end{bmatrix}^T \in \mathbb{R}^{6 \times 3h}$ is the extended shape matrix of $\mathbf{r}$.
By replacing the shape matrix of $\mathbf{r}$ and the feature inversion function with their extended versions,
we can extend our adaptive robust laws to control 
the following linear and angular velocities:
\begin{equation}
\mathbf{v}(t) = \begin{bmatrix} \dot{\mathbf{u}}_{\boldsymbol{\gamma}}(\mathbf{r},t),   &   \boldsymbol{\omega}_{\boldsymbol{\gamma}}(\mathbf{r},t) \end{bmatrix}^T
\end{equation}
where $\dot{\mathbf{u}}_{\boldsymbol{\gamma}}(\mathbf{r},t) \in \mathbb{R}^{3k}$ and $\boldsymbol{\omega}_{\boldsymbol{\gamma}}(\mathbf{r},t) \in \mathbb{R}^{3k}$ denotes the linear and angular velocities for the projections of $\mathbf{r}$ on the graph.
They can be further transformed to the velocities of the manipulation points with:
\begin{equation}
    \begin{bmatrix} \dot{\mathbf{u}}_{\boldsymbol{\gamma}}(\mathbf{r},t)  \\  \boldsymbol{\omega}_{\boldsymbol{\gamma}}(\mathbf{r},t) \end{bmatrix} = \begin{bmatrix}
\mathbf{I}_{3 \times 3}  &    \begin{bmatrix}
\mathbf{u}_{\boldsymbol{\gamma}}(\mathbf{r},t_0)
\end{bmatrix}_{\times}^T                   \\
\mathbf{0}_{3 \times 3}  &  \mathbf{I}_{3 \times 3}
\end{bmatrix}
    \begin{bmatrix} \dot{\mathbf{x}}(\mathbf{r},t)  \\  \boldsymbol{\omega}(\mathbf{r},t) \end{bmatrix}
\end{equation}
where $\boldsymbol{\omega}(\mathbf{r},t) \in \mathbb{R}^{3k}$ denotes the angular velocities of $\mathbf{r}$.

For experimental validations, a duck-shaped silicone toy was selected to be the object.
We set a "left" case and a "right" case by deforming the object to the left (Figure~\ref{fig:rotation_setup}(c,d)) and right (Figure~\ref{fig:rotation_setup}(g,h)) respectively.
The desired deformations were generated with both translational and rotational motions.
The CD for the "left" case is $d_{CD} = 1.79(mm)$ and the "right" case is $d_{CD} = 2.64(mm)$.
The comparisons between the measured and the desired points of the object in the two cases
are given in Figure~\ref{fig:rotation_setup}(a,b) and (e,f), respectively.
We can see that the object reached its desired deformation in general for both cases.
The result curves in Figure~\ref{fig:rotation}(a) show the minimization of the feature errors.
To show the translational and rotational manipulating motions,
we plot the linear and angular velocity commands of $\mathbf{r}$ in Figure~\ref{fig:rotation}(c) for the "left" case.
In addition, the manipulation trajectories in Figure~\ref{fig:rotation}(b) show the manipulation point reaches its target pose in general for both cases.

\section{8. Discussions and Conclusions}
\subsection{8.1. Discussions}
We discuss the limitations of the proposed method.
First, our control laws were formulated under local and quasi-static assumptions, thus
we recommend using the method for local deformation control tasks with slow robot motion.
Second, the focus of this paper is on control but not on planning.
Although we have demonstrated the capabilities of our method to be applied in a two-stage task with changes of the manipulation point, further efforts can be made for the global planning of longer-horizon and the automatic selections of manipulation points. 
Third, our deformation features are computed using the linear modal analysis that cannot accurately model large rotational deformation due to the omission of the nonlinear term in the strain tensor.
In our experiments, we show that our adaptive robust controller can tackle this unmodeled effect.
For more complex tasks with large rotational deformation, computing features using nonlinear modal analysis needs to be further studied.
In addition, there are slight differences between the resulting points and the desired points in some experiments.
The potential reasons include the neglect of shape details in the modal truncation, sensing noises, disturbances, and unmodeled effects.
Overcoming these limitations is the focus of our future work.

We further compare our method with some state-of-the-art algorithms that develop learning-based methods for shape deformation control (\cite{thach2022learning}), rearranging of deformable objects (\cite{shi2022robocraft}), and shaping elasto-plastic objects (\cite{shen2022acid}) under point clouds.
Compared with these learning-based methods, our method is learning-free.
The advantages of our learning-free controller are discussed as follows:
1) our method does not require offline and/or online training. It can be easily adapted to new objects and deal with changing environmental conditions;
2) our method is designed with transparent and interpretable mathematical formulations, which allows us to evaluate its stability and robustness analytically. The proposed controller is rigorously proven to be input-to-state stable, which is critical to ensure reliable and safe object manipulation.
We conduct extensive experiments using different objects, under different settings, and with external and unmodeled disturbances.
The results demonstrate the stability and robustness of our method without requiring offline or/and online training, therefore implying the advantages of our learning-free controller compared with these learning-based works.
We also emphasize that our method is not incompatible with
the learning-based techniques.
It can be combined with the learning-based techniques to leverage the advantages of both learning-free and learning-based control.
For a complex DOM task, our deformation controller can be applied to a local stage to ensure stabilities and reliabilities, while the learning-based techniques can be used for more global task planning. 
This will be a direction of our future research.

\subsection{8.2. Conclusions}
In this paper, we proposed a novel modal-graph framework describing the low-frequency deformation structure of the manipulated object.
The modal graph enabled us to directly extract low-dimensional deformation features from raw point cloud measurements,
while preserving the spatial structure of the DOM system to inverse the feature changes into motions of the robot manipulation.
In addition,
based on the proposed framework, 
we designed a model-free controller for the shape servoing of deformable objects with unknown physical properties and undeformed geometries.
We also proved that the controller is ISS using the Lyapunov-based method.
We conducted a series of experiments to show that the framework can deal with linear, planar, tubular, and volumetric objects of different shapes and materials under different settings.
The results validate the stability and robustness of our method with disturbances and unmodeled effects and demonstrate the potential to be generalized to different tasks.
To be brief,
first,
our control framework is simple.
It does not require extra processing on the measured point cloud (such as registrations, refinements, and occlusion removal), nor offline learning or identifications of the object's deformation models.
Second,
the framework is general as it is not object-specific nor task-specific.
In addition, 
the framework can be extended for other tasks in DOM and combined with graph-based learning techniques. 

In the future, we will extend the proposed modal-graph framework to develop control and planning strategies for global deformation control, and for different manipulation tasks.

\begin{funding}
This work is supported in part by Shenzhen Portion of Shenzhen-Hong Kong Science and Technology Innovation Cooperation Zone under HZQB-KCZYB-20200089, in part of the HK RGC under T42-409/18-R and 14202918, in part by the Multi-Scale Medical Robotics Centre, InnoHK, and in part by the VC Fund 4930745 of the CUHK T Stone Robotics Institute.
\end{funding}

\bibliographystyle{SageH}
\bibliography{reference.bib}

\begin{thebibliography}{64}
\providecommand{\natexlab}[1]{#1}
\providecommand{\url}[1]{\texttt{#1}}
\providecommand{\urlprefix}{URL }
\expandafter\ifx\csname urlstyle\endcsname\relax
  \providecommand{\doi}[1]{DOI:\discretionary{}{}{}#1}\else
  \providecommand{\doi}{DOI:\discretionary{}{}{}\begingroup
  \urlstyle{rm}\Url}\fi

\bibitem[{Achlioptas et~al.(2018)Achlioptas, Diamanti, Mitliagkas and
  Guibas}]{achlioptas2018learning}
Achlioptas P, Diamanti O, Mitliagkas I and Guibas L (2018) Learning
  representations and generative models for 3d point clouds.
\newblock In: \emph{International conference on machine learning}. PMLR, pp.
  40--49.

\bibitem[{Allard et~al.(2007)Allard, Cotin, Faure, Bensoussan, Poyer, Duriez,
  Delingette and Grisoni}]{allard2007sofa}
Allard J, Cotin S, Faure F, Bensoussan PJ, Poyer F, Duriez C, Delingette H and
  Grisoni L (2007) Sofa-an open source framework for medical simulation.
\newblock In: \emph{MMVR 15-Medicine Meets Virtual Reality}, volume 125. IOP
  Press, pp. 13--18.

\bibitem[{An et~al.(2008)An, Kim and James}]{an2008optimizing}
An SS, Kim T and James DL (2008) Optimizing cubature for efficient integration
  of subspace deformations.
\newblock \emph{ACM transactions on graphics (TOG)} 27(5): 1--10.

\bibitem[{Aranda et~al.(2020)Aranda, Ramon, Mezouar, Bartoli and
  {\"O}zg{\"u}r}]{aranda2020monocular}
Aranda M, Ramon JAC, Mezouar Y, Bartoli A and {\"O}zg{\"u}r E (2020) Monocular
  visual shape tracking and servoing for isometrically deforming objects.
\newblock In: \emph{2020 IEEE/RSJ International Conference on Intelligent
  Robots and Systems (IROS)}. IEEE, pp. 7542--7549.

\bibitem[{Atluri and Zhu(2000)}]{atluri2000new}
Atluri S and Zhu T (2000) New concepts in meshless methods.
\newblock \emph{International journal for numerical methods in engineering}
  47(1-3): 537--556.

\bibitem[{Barbi{\v{c}} and James(2005)}]{barbivc2005real}
Barbi{\v{c}} J and James DL (2005) Real-time subspace integration for st.
  venant-kirchhoff deformable models.
\newblock \emph{ACM transactions on graphics (TOG)} 24(3): 982--990.

\bibitem[{Barr(1987)}]{barr1987global}
Barr AH (1987) Global and local deformations of solid primitives.
\newblock In: \emph{Readings in Computer Vision}. Elsevier, pp. 661--670.

\bibitem[{Belytschko et~al.(1994)Belytschko, Lu and Gu}]{belytschko1994element}
Belytschko T, Lu YY and Gu L (1994) Element-free galerkin methods.
\newblock \emph{International journal for numerical methods in engineering}
  37(2): 229--256.

\bibitem[{Berenson(2013)}]{berenson2013manipulation}
Berenson D (2013) Manipulation of deformable objects without modeling and
  simulating deformation.
\newblock In: \emph{2013 IEEE/RSJ International Conference on Intelligent
  Robots and Systems}. IEEE, pp. 4525--4532.

\bibitem[{Bookstein(1989)}]{bookstein1989principal}
Bookstein FL (1989) Principal warps: Thin-plate splines and the decomposition
  of deformations.
\newblock \emph{IEEE Transactions on pattern analysis and machine intelligence}
  11(6): 567--585.

\bibitem[{Botsch and Sorkine(2007)}]{botsch2007linear}
Botsch M and Sorkine O (2007) On linear variational surface deformation
  methods.
\newblock \emph{IEEE transactions on visualization and computer graphics}
  14(1): 213--230.

\bibitem[{Choi and Ko(2005)}]{choi2005modal}
Choi MG and Ko HS (2005) Modal warping: Real-time simulation of large
  rotational deformation and manipulation.
\newblock \emph{IEEE Transactions on Visualization and Computer Graphics}
  11(1): 91--101.

\bibitem[{Davatzikos et~al.(2003)Davatzikos, Tao and
  Shen}]{davatzikos2003hierarchical}
Davatzikos C, Tao X and Shen D (2003) Hierarchical active shape models, using
  the wavelet transform.
\newblock \emph{IEEE transactions on medical imaging} 22(3): 414--423.

\bibitem[{El~Ouatouati and Johnson(1999)}]{el1999new}
El~Ouatouati A and Johnson D (1999) A new approach for numerical modal analysis
  using the element-free method.
\newblock \emph{International Journal for Numerical Methods in Engineering}
  46(1): 1--27.

\bibitem[{Fayad et~al.(2010)Fayad, Agapito and Del~Bue}]{fayad2010piecewise}
Fayad J, Agapito L and Del~Bue A (2010) Piecewise quadratic reconstruction of
  non-rigid surfaces from monocular sequences.
\newblock In: \emph{European conference on computer vision}. Springer, pp.
  297--310.

\bibitem[{Fulton et~al.(2019)Fulton, Modi, Duvenaud, Levin and
  Jacobson}]{fulton2019latent}
Fulton L, Modi V, Duvenaud D, Levin DI and Jacobson A (2019) Latent-space
  dynamics for reduced deformable simulation.
\newblock In: \emph{Computer graphics forum}, volume~38. Wiley Online Library,
  pp. 379--391.

\bibitem[{Hu et~al.(2019)Hu, Han, Sun, Pan and Manocha}]{hu20193}
Hu Z, Han T, Sun P, Pan J and Manocha D (2019) 3-d deformable object
  manipulation using deep neural networks.
\newblock \emph{IEEE Robotics and Automation Letters} 4(4): 4255--4261.

\bibitem[{Hu et~al.(2018)Hu, Sun and Pan}]{hu2018three}
Hu Z, Sun P and Pan J (2018) Three-dimensional deformable object manipulation
  using fast online gaussian process regression.
\newblock \emph{IEEE Robotics and Automation Letters} 3(2): 979--986.

\bibitem[{Innmann et~al.(2016)Innmann, Zollh{\"o}fer, Nie{\ss}ner, Theobalt and
  Stamminger}]{innmann2016volumedeform}
Innmann M, Zollh{\"o}fer M, Nie{\ss}ner M, Theobalt C and Stamminger M (2016)
  Volumedeform: Real-time volumetric non-rigid reconstruction.
\newblock In: \emph{Computer Vision--ECCV 2016: 14th European Conference,
  Amsterdam, The Netherlands, October 11-14, 2016, Proceedings, Part VIII 14}.
  Springer, pp. 362--379.

\bibitem[{Jin et~al.(2019)Jin, Wang and Tomizuka}]{jin2019robust}
Jin S, Wang C and Tomizuka M (2019) Robust deformation model approximation for
  robotic cable manipulation.
\newblock In: \emph{2019 IEEE/RSJ International Conference on Intelligent
  Robots and Systems (IROS)}. IEEE, pp. 6586--6593.

\bibitem[{Kelemen and Gerig(1996)}]{kelemen1996segmentation}
Kelemen CB and Gerig G (1996) Segmentation of 2-d and 3-d objects from mri
  volume data using constrained elastic deformations of flexible fourier
  contour and surface models.
\newblock \emph{Medical image analysis} 1(1): 19--34.

\bibitem[{Lagneau et~al.(2020{\natexlab{a}})Lagneau, Krupa and
  Marchal}]{lagneau2020active}
Lagneau R, Krupa A and Marchal M (2020{\natexlab{a}}) Active deformation
  through visual servoing of soft objects.
\newblock In: \emph{2020 IEEE International Conference on Robotics and
  Automation (ICRA)}. IEEE, pp. 8978--8984.

\bibitem[{Lagneau et~al.(2020{\natexlab{b}})Lagneau, Krupa and
  Marchal}]{lagneau2020automatic}
Lagneau R, Krupa A and Marchal M (2020{\natexlab{b}}) Automatic shape control
  of deformable wires based on model-free visual servoing.
\newblock \emph{IEEE Robotics and Automation Letters} 5(4): 5252--5259.

\bibitem[{Li et~al.(2018)Li, Su and Liu}]{li2018vision}
Li X, Su X and Liu YH (2018) Vision-based robotic manipulation of flexible
  pcbs.
\newblock \emph{IEEE/ASME Transactions on Mechatronics} 23(6): 2739--2749.

\bibitem[{Lin et~al.(2022)Lin, Wang, Huang and Held}]{lin2022learning}
Lin X, Wang Y, Huang Z and Held D (2022) Learning visible connectivity dynamics
  for cloth smoothing.
\newblock In: \emph{Conference on Robot Learning}. PMLR, pp. 256--266.

\bibitem[{Lippi et~al.(2020)Lippi, Poklukar, Welle, Varava, Yin, Marino and
  Kragic}]{lippi2020latent}
Lippi M, Poklukar P, Welle MC, Varava A, Yin H, Marino A and Kragic D (2020)
  Latent space roadmap for visual action planning of deformable and rigid
  object manipulation.
\newblock In: \emph{2020 IEEE/RSJ International Conference on Intelligent
  Robots and Systems (IROS)}. IEEE, pp. 5619--5626.

\bibitem[{Liu et~al.(2020)Liu, Sheng, Yang, Shao and Hu}]{liu2020morphing}
Liu M, Sheng L, Yang S, Shao J and Hu SM (2020) Morphing and sampling network
  for dense point cloud completion.
\newblock In: \emph{Proceedings of the AAAI conference on artificial
  intelligence}, volume~34. pp. 11596--11603.

\bibitem[{Liu et~al.(2013)Liu, Bargteil, O'Brien and Kavan}]{liu2013fast}
Liu T, Bargteil AW, O'Brien JF and Kavan L (2013) Fast simulation of
  mass-spring systems.
\newblock \emph{ACM Transactions on Graphics (TOG)} 32(6): 1--7.

\bibitem[{Ma et~al.(2022)Ma, Zhu and Navarro-Alarcon}]{ma2022shape}
Ma W, Zhu J and Navarro-Alarcon D (2022) Shape control of elastic objects based
  on implicit sensorimotor models and data-driven geometric features.
\newblock In: \emph{International Conference on Intelligent Autonomous
  Systems}. Springer, pp. 518--531.

\bibitem[{McConachie et~al.(2020)McConachie, Dobson, Ruan and
  Berenson}]{mcconachie2020manipulating}
McConachie D, Dobson A, Ruan M and Berenson D (2020) Manipulating deformable
  objects by interleaving prediction, planning, and control.
\newblock \emph{The International Journal of Robotics Research} 39(8):
  957--982.

\bibitem[{Navarro-Alarcon and Liu(2018)}]{navarro2018fourier}
Navarro-Alarcon D and Liu YH (2018) Fourier-based shape servoing: a new
  feedback method to actively deform soft objects into desired 2-d image
  contours.
\newblock \emph{IEEE Transactions on Robotics} 34(1): 272--279.

\bibitem[{Navarro-Alarcon et~al.(2016)Navarro-Alarcon, Yip, Wang, Liu, Zhong,
  Zhang and Li}]{navarro2016automatic}
Navarro-Alarcon D, Yip HM, Wang Z, Liu YH, Zhong F, Zhang T and Li P (2016)
  Automatic 3-d manipulation of soft objects by robotic arms with an adaptive
  deformation model.
\newblock \emph{IEEE Transactions on Robotics} 32(2): 429--441.

\bibitem[{Nealen et~al.(2006)Nealen, M{\"u}ller, Keiser, Boxerman and
  Carlson}]{nealen2006physically}
Nealen A, M{\"u}ller M, Keiser R, Boxerman E and Carlson M (2006) Physically
  based deformable models in computer graphics.
\newblock In: \emph{Computer graphics forum}, volume~25. Wiley Online Library,
  pp. 809--836.

\bibitem[{Nguyen et~al.(2008)Nguyen, Rabczuk, Bordas and
  Duflot}]{nguyen2008meshless}
Nguyen VP, Rabczuk T, Bordas S and Duflot M (2008) Meshless methods: a review
  and computer implementation aspects.
\newblock \emph{Mathematics and computers in simulation} 79(3): 763--813.

\bibitem[{Paschalidou et~al.(2019)Paschalidou, Ulusoy and
  Geiger}]{paschalidou2019superquadrics}
Paschalidou D, Ulusoy AO and Geiger A (2019) Superquadrics revisited: Learning
  3d shape parsing beyond cuboids.
\newblock In: \emph{Proceedings of the IEEE/CVF Conference on Computer Vision
  and Pattern Recognition}. pp. 10344--10353.

\bibitem[{Pentland and Sclaroff(1991)}]{pentland1991closed}
Pentland A and Sclaroff S (1991) Closed-form solutions for physically based
  shape modeling and recognition.
\newblock \emph{IEEE Transactions on Pattern Analysis \& Machine Intelligence}
  (7): 715--729.

\bibitem[{Pentland and Williams(1989)}]{pentland1989good}
Pentland A and Williams J (1989) Good vibrations: Modal dynamics for graphics
  and animation.
\newblock In: \emph{Proceedings of the 16th annual conference on Computer
  graphics and interactive techniques}. pp. 215--222.

\bibitem[{Pentland(1987)}]{pentland1987perceptual}
Pentland AP (1987) Perceptual organization and the representation of natural
  form.
\newblock In: \emph{Readings in Computer Vision}. Elsevier, pp. 680--699.

\bibitem[{Pfaff et~al.(2020)Pfaff, Fortunato, Sanchez-Gonzalez and
  Battaglia}]{pfaff2020learning}
Pfaff T, Fortunato M, Sanchez-Gonzalez A and Battaglia PW (2020) Learning
  mesh-based simulation with graph networks.
\newblock \emph{arXiv preprint arXiv:2010.03409} .

\bibitem[{Shademan et~al.(2016)Shademan, Decker, Opfermann, Leonard, Krieger
  and Kim}]{shademan2016supervised}
Shademan A, Decker RS, Opfermann JD, Leonard S, Krieger A and Kim PC (2016)
  Supervised autonomous robotic soft tissue surgery.
\newblock \emph{Science translational medicine} 8(337): 337ra64--337ra64.

\bibitem[{Shen et~al.(2022)Shen, Jiang, Choy, J.~Guibas, Savarese, Anandkumar
  and Zhu}]{shen2022acid}
Shen B, Jiang Z, Choy C, J~Guibas L, Savarese S, Anandkumar A and Zhu Y (2022)
  Acid: Action-conditional implicit visual dynamics for deformable object
  manipulation.
\newblock \emph{Robotics: Science and Systems (RSS)} .

\bibitem[{Shepard(1968)}]{shepard1968two}
Shepard D (1968) A two-dimensional interpolation function for
  irregularly-spaced data.
\newblock In: \emph{Proceedings of the 1968 23rd ACM national conference}. pp.
  517--524.

\bibitem[{Shetab-Bushehri et~al.(2022)Shetab-Bushehri, Aranda, Mezouar and
  Ozgur}]{shetab2022rigid}
Shetab-Bushehri M, Aranda M, Mezouar Y and Ozgur E (2022) As-rigid-as-possible
  shape servoing.
\newblock \emph{IEEE Robotics and Automation Letters} .

\bibitem[{Shi et~al.(2022)Shi, Xu, Huang, Li and Wu}]{shi2022robocraft}
Shi H, Xu H, Huang Z, Li Y and Wu J (2022) Robocraft: Learning to see,
  simulate, and shape elasto-plastic objects with graph networks.
\newblock \emph{arXiv preprint arXiv:2205.02909} .

\bibitem[{Shin et~al.(2019)Shin, Ferguson, Pedram, Ma, Dutson and
  Rosen}]{shin2019autonomous}
Shin C, Ferguson PW, Pedram SA, Ma J, Dutson EP and Rosen J (2019) Autonomous
  tissue manipulation via surgical robot using learning based model predictive
  control.
\newblock In: \emph{2019 International Conference on Robotics and Automation
  (ICRA)}. IEEE, pp. 3875--3881.

\bibitem[{Shiu and Ahmad(1989)}]{shiu1989calibration}
Shiu Y and Ahmad S (1989) Calibration of wrist-mounted robotic sensors by
  solving homogeneous transform equations of the form ax= xb.
\newblock \emph{IEEE Transactions on Robotics and Automation} 5(1): 16--29.

\bibitem[{Slavcheva et~al.(2017)Slavcheva, Baust, Cremers and
  Ilic}]{slavcheva2017killingfusion}
Slavcheva M, Baust M, Cremers D and Ilic S (2017) Killingfusion: Non-rigid 3d
  reconstruction without correspondences.
\newblock In: \emph{Proceedings of the IEEE Conference on Computer Vision and
  Pattern Recognition}. pp. 1386--1395.

\bibitem[{Solina and Bajcsy(1990)}]{solina1990recovery}
Solina F and Bajcsy R (1990) Recovery of parametric models from range images:
  The case for superquadrics with global deformations.
\newblock \emph{IEEE transactions on pattern analysis and machine intelligence}
  12(2): 131--147.

\bibitem[{Sontag(2008)}]{sontag2008input}
Sontag ED (2008) Input to state stability: Basic concepts and results.
\newblock In: \emph{Nonlinear and optimal control theory}. Springer, pp.
  163--220.

\bibitem[{Sui et~al.(2019)Sui, He, Lyu, Wang and Liu}]{sui20193d}
Sui C, He K, Lyu C, Wang Z and Liu YH (2019) 3d surface reconstruction using a
  two-step stereo matching method assisted with five projected patterns.
\newblock In: \emph{2019 International Conference on Robotics and Automation
  (ICRA)}. IEEE, pp. 6080--6086.

\bibitem[{Sundaresan et~al.(2022)Sundaresan, Antonova and
  Bohg}]{sundaresan2022diffcloud}
Sundaresan P, Antonova R and Bohg J (2022) Diffcloud: Real-to-sim from point
  clouds with differentiable simulation and rendering of deformable objects.
\newblock \emph{arXiv preprint arXiv:2204.03139} .

\bibitem[{Tan et~al.(2020)Tan, Pan, Gao and Manocha}]{tan2020realtime}
Tan Q, Pan Z, Gao L and Manocha D (2020) Realtime simulation of thin-shell
  deformable materials using cnn-based mesh embedding.
\newblock \emph{IEEE Robotics and Automation Letters} 5(2): 2325--2332.

\bibitem[{Tang and Tomizuka(2018)}]{tang2018track}
Tang T and Tomizuka M (2018) Track deformable objects from point clouds with
  structure preserved registration.
\newblock \emph{The International Journal of Robotics Research} :
  0278364919841431.

\bibitem[{Tang and Tomizuka(2022)}]{tang2022track}
Tang T and Tomizuka M (2022) Track deformable objects from point clouds with
  structure preserved registration.
\newblock \emph{The International Journal of Robotics Research} 41(6):
  599--614.

\bibitem[{Terzopoulos et~al.(1987)Terzopoulos, Platt, Barr and
  Fleischer}]{terzopoulos1987elastically}
Terzopoulos D, Platt J, Barr A and Fleischer K (1987) Elastically deformable
  models.
\newblock In: \emph{Proceedings of the 14th annual conference on Computer
  graphics and interactive techniques}. pp. 205--214.

\bibitem[{Thach et~al.(2022)Thach, Cho, Kuntz and Hermans}]{thach2022learning}
Thach B, Cho BY, Kuntz A and Hermans T (2022) learning visual shape control of
  novel 3d deformable objects from partial-view point clouds.
\newblock In: \emph{2022 International Conference on Robotics and Automation
  (ICRA)}. IEEE, pp. 8274--8281.

\bibitem[{Tsurumine et~al.(2019)Tsurumine, Cui, Uchibe and
  Matsubara}]{tsurumine2019deep}
Tsurumine Y, Cui Y, Uchibe E and Matsubara T (2019) Deep reinforcement learning
  with smooth policy update: Application to robotic cloth manipulation.
\newblock \emph{Robotics and Autonomous Systems} 112: 72--83.

\bibitem[{Wang et~al.(2022)Wang, Zhang, Zhang, Wu, Zhu, Jin, Tang and
  Tomizuka}]{wang2022offline}
Wang C, Zhang Y, Zhang X, Wu Z, Zhu X, Jin S, Tang T and Tomizuka M (2022)
  Offline-online learning of deformation model for cable manipulation with
  graph neural networks.
\newblock \emph{IEEE Robotics and Automation Letters} 7(2): 5544--5551.

\bibitem[{Wu et~al.(2021)Wu, Pan, Zhang, Wang, Liu and Lin}]{wu2021balanced}
Wu T, Pan L, Zhang J, Wang T, Liu Z and Lin D (2021) Balanced chamfer distance
  as a comprehensive metric for point cloud completion.
\newblock \emph{Advances in Neural Information Processing Systems} 34:
  29088--29100.

\bibitem[{Yang et~al.(2023)Yang, Lu, Chen, Zhong and Liu}]{10122176}
Yang B, Lu B, Chen W, Zhong F and Liu YH (2023) Model-free 3-d shape control of
  deformable objects using novel features based on modal analysis.
\newblock \emph{IEEE Transactions on Robotics.} 39(4): 3134--3153.

\bibitem[{Yu et~al.(2022)Yu, Lv, Zhong, Song and Li}]{yu2022global}
Yu M, Lv K, Zhong H, Song S and Li X (2022) Global model learning for large
  deformation control of elastic deformable linear objects: An efficient and
  adaptive approach.
\newblock \emph{IEEE Transactions on Robotics} 39(1): 417--436.

\bibitem[{Zhou et~al.(2021)Zhou, Zhu, Huo and
  Navarro-Alarcon}]{zhou2021lasesom}
Zhou P, Zhu J, Huo S and Navarro-Alarcon D (2021) Lasesom: A latent and
  semantic representation framework for soft object manipulation.
\newblock \emph{IEEE Robotics and Automation Letters} 6(3): 5381--5388.

\bibitem[{Zienkiewicz et~al.(2005)Zienkiewicz, Taylor and
  Zhu}]{zienkiewicz2005finite}
Zienkiewicz OC, Taylor RL and Zhu JZ (2005) \emph{The finite element method:
  its basis and fundamentals}.
\newblock Elsevier.

\bibitem[{Zollh{\"o}fer et~al.(2014)Zollh{\"o}fer, Nie{\ss}ner, Izadi, Rehmann,
  Zach, Fisher, Wu, Fitzgibbon, Loop, Theobalt et~al.}]{zollhofer2014real}
Zollh{\"o}fer M, Nie{\ss}ner M, Izadi S, Rehmann C, Zach C, Fisher M, Wu C,
  Fitzgibbon A, Loop C, Theobalt C et~al. (2014) Real-time non-rigid
  reconstruction using an rgb-d camera.
\newblock \emph{ACM Transactions on Graphics (ToG)} 33(4): 1--12.

\end{thebibliography}

\end{document}